\documentclass[11pt]{article}

\usepackage{times}

\usepackage[
paper=letterpaper,
left=1in,
right=1in,
top=1in,
bottom=1in
]{geometry}

\usepackage{amsmath,amssymb,amsthm}

\usepackage{graphicx}
\usepackage{subcaption}

\usepackage{subcaption}

\usepackage{hyperref}
\usepackage{doi}

\usepackage{bm}
\usepackage{bbm}
\usepackage{mathrsfs}
\usepackage{enumitem}
\usepackage{xcolor}
\usepackage[authoryear,round]{natbib} 

\newtheorem{theo}{Theorem}

\newtheorem{lem}[theo]{Lemma}

\newtheorem{coro}[theo]{Corollary}
\newtheorem{remark}[theo]{Remark}
\newtheorem{const}[theo]{Construction}
\newcommand{\norm}[1]{\left\lVert #1 \right\rVert}

\newcommand{\widesim}[2][1.1]{
	\mathrel{\overset{#2}{\scalebox{#1}[1]{$\sim$}}}
}
\newcommand{\suppf}[1]{\mathrm{supp}\,#1}

\title{Minimax learning rates for estimating binary classifiers under margin conditions}

\author{Jonathan Garc\'ia \footnote{Faculty of Mathematics, University of Vienna, Austria. \\ \texttt{jonathan.garcia.rebellon@univie.ac.at }} \and 
	Philipp Petersen  \footnote{Faculty of Mathematics and Research Network Data Science @ Uni Vienna, University of Vienna, Austria. \\ \texttt{philipp.petersen@univie.ac.at}}}
\date{}

\begin{document}

\maketitle

\begin{abstract}
	\normalsize
	We study classification problems using binary estimators where the decision boundary is described by horizon functions and where the data distribution satisfies a geometric margin condition. A key novelty of our work is the derivation of lower bounds for the worst-case learning rates over broad classes of functions, under a geometric margin condition---a setting that is almost universally satisfied in practice, but remains theoretically challenging. Moreover, 
    we work in the noiseless setting, where lower bounds are particularly hard to establish. Our general results cover, in particular, classification problems with decision boundaries belonging to several classes of functions: for Barron-regular functions, Hölder-continuous functions, and convex-Lipschitz functions with strong margins, we identify optimal rates close to the fast learning rates of $\mathcal{O}(n^{-1})$ for $n \in \mathbb{N}$ samples.
\end{abstract}

\medskip



\medskip



\section{Introduction}

\begin{figure}[htb]
	\centering
	\includegraphics[width = \textwidth]{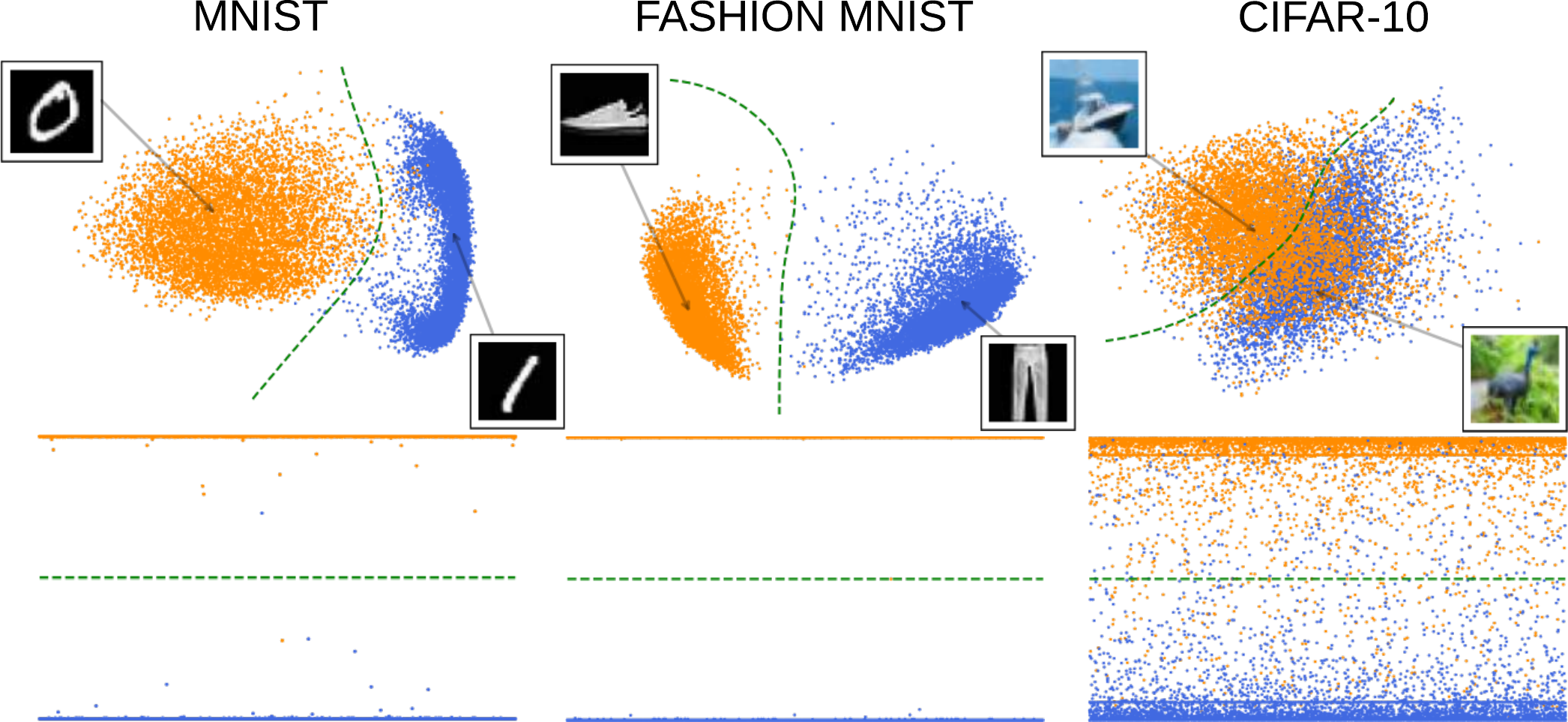}
	\caption{
			Geometric margin in common classification problems. The top row shows a two-dimensional embedding on the first two principal components and a decision boundary identified by a support vector machine. Clearly MNIST \cite{MNISTref} and Fashion MNIST  \cite{FashionMNISTref} exhibit a strong margin between some classes. For the CIFAR-10 data \cite{CIFAR10ref} the margin is not visible in the two-dimensional embedding. In the second row, we show the class probabilities predicted by a support vector classifier, which again shows an extremely strong margin for MNIST and Fashion MNIST, but also reveals that the CIFAR-10 data exhibits a margin, albeit a weaker one. Which lower bounds on learning can be found in the presence of such various types of margins will be demonstrated in our main results (Section~\ref{mainresult}).	
	}
	\label{graph0}
\end{figure}

How well can we solve classification problems with complex decision boundaries in deep learning?  A lot of emphasis has been put on the noise and the decision boundary in the problem. However, in practice, data sets may have very strong margins (formally defined in \ref{marginc} below) between the classes, which makes learning much simpler (see e.g. Figure~\ref{graph0}).  
The presence of a margin seriously complicates the identification of lower bounds on learning success. This is intuitively clear, since in the extreme case, where certain regions between the classes almost surely do not contain any data points, many decision boundaries are valid. In this work, we overcome these issues and present lower bounds on learning under margin conditions.

\subsection{Statistical framework for binary classification}\label{sectfmk}
We consider $ n \in \mathbb{N}$ samples $ \bm{S}_{n}:=\left((\bm{x}_{i},y_{i})\right)_{i=1}^{n} $, where $ \bm{x}_{i}\in \mathcal{X}:=[0,1]^{d} $ with $ d\in\mathbb{N}_{\geq 2} $, are input vectors; and $ y_{i}\in \{0,1\} $ are class labels. We assume the random variables $ (\bm{x}_{i},y_{i}) $ to be independent and identically distributed (iid) according to an unknown joint probability measure $ \bm{\mu} $ on $\Lambda:=\mathcal{X}\times \{0,1\}$, and we denote this by $ (\bm{x}_{i},y_{i})\widesim{\text{iid}} \bm{\mu} $. We call $ \mu $ the marginal probability measure on $ \mathcal{X} $ induced by $ \bm{\mu} $, and assume that $\mu$ admits a density $ f:\mathcal{X}\to [0,\infty) $ with respect to the Lebesgue measure $ \lambda $, such that $ f\in L^{\infty}(\lambda) $. Moreover, we fix a reference measure on $\Lambda$ of the form 
\begin{equation}
	\bm{\lambda}:=\lambda\times\eta, ~\text{ where }~ \eta~ \text{ is a probability measure on }~\{ 0,1 \}~\text{ with }~\eta(\{0\}),\eta(\{1\})>0. \label{bmlambda}
\end{equation}
With this convention, $\bm{\mu}$ admits a density with respect to $\bm{\lambda}$. Consequently, the distribution of the iid sample $ \bm{S}_{n} $ on $\Lambda^{n}$ is $ \bm{\mu}^{\otimes n} $ and admits a density with respect to the product reference measure $ \bm{\lambda}^{\otimes n} $.

A binary classifier can be defined as an indicator function $ h:=\mathbbm{1}_{\Omega} $, where $ \Omega\subset\mathcal{X} $ is a decision region, such that in a noiseless setting---assumed throughout this paper---$ h(\bm{x}_{i})=y_{i} $, for all $ i=1,\ldots,n $.

\begin{remark}\label{noisere} 
	The noiseless assumption introduced above is an important assumption if we want to resolve the precise role of the regularity of the decision boundary and the margin conditions. 
	Indeed, even the presence of low noise could yield vastly different lower bounds, because the learning problem then requires resolving the noise and this complication can mask the role of the decision boundary and the margin. 
	An extended discussion of this is given in \cite[Section 1.1, Point 1]{petersen2021optimal}. 
	For a quick argument, we highlight, e.g., \cite{noise1}, where it was obtained that the optimal learning rate, to learn a function $f\in C^{k}([0,1]^{d})$ with $\norm{f}_{C^{k}}\leq 1$ and noise defined as a parameter $ \varepsilon \widesim{iid} N(0,\sigma^{2})$ for $\sigma>0$, is of the order of $O(n^{-k/(2k+d)})$,
	and decays slower than $n^{-1/2}$. 
	On the other hand, in \cite{noiseless1}, where the same problem is considered without noise, learning rates of the order of $\mathcal{O}(n^{-k/d})$ were obtained, in some cases faster than $n^{-1}$.
\end{remark}

An estimator $\hat{h}_{n}$ is a classifier constructed from the observed sample $ \bm{S}_{n} $, intended to approximate the true classifier $h$ over $\mathcal{X}$. Such estimators are produced by learning algorithms, which are measurable mappings
\begin{equation}
	A\in \mathcal{A}_{n}(\mathcal{G}):=\left\{A:\Lambda^{n}\to \mathcal{G} \mid  A \text{ is measurable} \right\},\quad\text{where}\quad \mathcal{G} \subseteq L^2(\lambda).\label{defestim}
\end{equation} 
Here, the algorithm $A$ specifies the procedure that maps the observed sample to the estimator.

\subsection{Conditions}

We now specify the conditions on the classifiers and distributions that allow us to formulate our minimax problem.

\begin{enumerate}[label=C\arabic*)]
	\item\label{learseth} The decision regions can be described by horizon functions (see  \cite{petersen2021optimal}):
	
	We consider a class of continuous functions $	\mathscr{C}\subset C([0,1]^{d-1};[0,1]) $ and define, for each $ b\in \mathscr{C} $, the associated horizon function as
	\begin{align}
		h_{b}:\mathcal{X}&\to \{0,1\},\nonumber\\
		\bm{x}=(x_{1},\ldots,x_{d})&\mapsto \mathbbm{1}_{b(\bm{x}^{(d)})\leq x_{d}},\label{bcont}
	\end{align}
	where $ \bm{x}^{(d)}:= (x_{1},\ldots,x_{d-1})$. We denote by $ H_{\mathscr{C}}:=\left\{h_{b}\mid b\in \mathscr{C}\right\} $  the set of horizon functions associated to $ \mathscr{C} $. We assume that
	\begin{equation}
		\Omega:=\Omega_{h}=\left\{\bm{x}\in \mathcal{X}\mid h(\bm{x})=1 \right\} \quad\text{where}\quad h\in H_{\mathscr{C}}.\label{defhbO}
	\end{equation}
	\begin{remark}\label{remhorc}
		~		
		\begin{enumerate}[label=(\alph*),ref=(\alph*)]
			
			\item\label{item1rhorc} For each $ h\in H_{\mathscr{C}} $, there exists a unique $ b\in \mathscr{C} $ such that $ h=h_{b} $, and conversely, each $ b\in \mathscr{C} $ is uniquely determined by its associated $ h_{b} $. This follows from the bijectivity of the map $ b \mapsto h_{b}  $. 
			
			\item Each horizon function $h\in H_{\mathscr{C}}$, associated with a function 
			$b\in\mathscr{C}$, defines a decision region $\Omega_h$ with its corresponding
			boundary
			\[
			\partial\Omega_{h}=\left\{\bm{x}\in \mathcal{X}\mid b(\bm{x}^{(d)})=x_{d} \right\}.
			\] 
		\end{enumerate}
	\end{remark}

	\item\label{conticon} Regular boundary (Hölder continuity): Let $ \alpha\in(0,1] $ be fixed. For all $ b\in \mathscr{C} $, there exists a constant $ K_{b}>0 $, such that 
	\begin{equation}
		|b(\bm{z})-b(\bm{w})|\leq K_{b}\norm{\bm{z}-\bm{w}}_{2}^{\alpha}\quad\text{for all}\quad \bm{z},\bm{w}\in [0,1]^{d-1}.\label{holdineq}
	\end{equation}

	\item\label{marginc} Geometric margin near the decision boundary: Let $ (h,\mu) $ be a pair consisting of a classifier $ h\in H_{\mathscr{C}}  $ and the marginal distribution $ \mu $ of the input vectors $\{\bm{x}_{i}\}_{i=1}^{n}$ in the sample $ \bm{S}_{n} $ introduced in Section~\ref{sectfmk}. We say that $ (h,\mu) $
 	satisfies the margin condition with exponent $ \gamma>0 $ if there exists $C>0$ such that, for all $ \varepsilon>0 $, 
	\begin{equation}
		\mu(B_{\varepsilon}^{h})\leq C\varepsilon^{\gamma}\quad \text{where}\quad B_{\varepsilon}^{h}:= \left\{\bm{x}\in\mathcal{X}\mid \mathrm{dist}\left(\bm{x},\partial\Omega_{h}\right)\leq \varepsilon\right\}
		\label{eq-marginc}
	\end{equation}
	is the neighborhood of radius $ \varepsilon>0 $ around $ \partial\Omega_{h} $ with respect to the Euclidean distance 
	\[
		\mathrm{dist}\left(\bm{x},\partial\Omega_{h}\right):=\inf_{\bm{y}\in \partial\Omega_{h}}\norm{\bm{x}-\bm{y}}_{2}.
	\] 
	We call $ \gamma $ the margin exponent (see \cite{SVM, fastc}).
	
	\begin{remark}
		The measure $ \mu $ associated with $ h $ is in general not unique: different data-generating distributions may share the same decision boundary $ h $, and conversely a given marginal distribution $ \mu $ may be compatible with multiple admissible boundaries in  $ H_{\mathscr{C}} $.
	\end{remark}
	Let $ \mathcal{L}_{\lambda} $ be the class of all probability measures on $\mathcal{X}$ that admit an essentially bounded density with respect to the Lebesgue measure $\lambda$, i.e.,  
	\begin{equation}
		\mathcal{L}_{\lambda}\!:=\!\left\{\mu\mid \text{ there exists } f:\mathcal{X}\to[0,\infty)\text{ such that } \! \int_{\mathcal{X}}fd\lambda\!=\!1,~ d\mu\!=\! f d\lambda \text{ and } f\in L^{\infty}(\lambda) \right\}.\label{measwrtleb}
	\end{equation}

	We denote by $ \mathcal{P}_{\mathscr{C}}(\mathcal{M}) $  the set of all pairs $ (h,\mu) $ that satisfy the margin condition, with $ h\in H_{\mathscr{C}} $ and $ \mu \in \mathcal{M}\subseteq \mathcal{L}_{\lambda}$. That is,
	\begin{equation}
		\mathcal{P}_{\mathscr{C}}(\mathcal{M}):=\left\{(h,\mu)\in H_{\mathscr{C}} \times\mathcal{M} \mid (h,\mu) ~\text{ satisfies }~ \eqref{eq-marginc}\right\}\quad\text{where}\quad\mathcal{M}\subseteq\mathcal{L}_{\lambda}.\label{sethmmargins}
	\end{equation} 
	
	\begin{remark}  
		 In particular, the margin condition is satisfied by $ (h,\mu) $ if the density $f$ of $\mu$ with respect to $\lambda$ satisfies
			\begin{equation*}
				f(\bm{x})  \lesssim \begin{cases}
					\min\left\{\frac{\varepsilon^{\gamma}}{\lambda(B_{\varepsilon}^{h})},1\right\} & \text{  if  }\; \bm{x}\in B_{\varepsilon}^{h}\\
					1 & \text{otherwise,}
				\end{cases}
			\end{equation*}
			for almost every $\bm{x}\in \mathcal{X}$ and for all $\varepsilon>0$. It is also satisfied when
			\[
			f(\bm{x})\lesssim \mathrm{dist}^{\gamma}\left(\bm{x},\partial\Omega_{h}\right)\qquad\text{for almost every}\qquad \bm{x}\in \mathcal{X}.
			\] 
		
	\end{remark}
\end{enumerate} 

Under the above conditions, we establish lower bounds for the minimax error associated with estimating binary classifiers whose decision regions satisfy \ref{learseth} on a function class $\mathscr{C}$ with regularity \ref{conticon}, and where the classifier–distribution pair satisfies the margin condition \ref{marginc}; i.e., we lower bound the following $\inf$-$\sup$ expression
\begin{equation}
	\mathcal{I}_{n}(\mathscr{C}):=\inf_{A\in\mathcal{A}_{n}(L^{2}(\lambda))}\, \sup_{(h,\mu)\in \mathcal{P}_{\mathscr{C}}(\mathcal{L}_{\lambda})} \mathbb{E}_{\{\bm{x}_{i}\}_{i=1}^{n} \widesim{iid} \mu}\norm{A(\bm{S}_{n})-h}_{L^{2}(\mu)}^{2},\label{eqinfsup}
\end{equation}
where $\bm{S}_{n}:=\left((\bm{x}_{i},y_{i})\right)_{i=1}^{n} $ denotes the sample, $ \mathcal{L}_{\lambda} $ is defined in \eqref{measwrtleb} and $ \mathcal{P}_{\mathscr{C}} $ is as in \eqref{sethmmargins}.

\begin{remark}
	~
	\begin{itemize}
		\item Throughout this work, we restrict to marginals $\mu$ with density $f=d\mu/d\lambda \in L^\infty(\lambda)$. Hence, for every $A(\bm{S}_{n})\in L^2(\lambda)$, the expression $ \norm{A(\bm{S}_{n})-h}_{L^{2}(\mu)}^{2} $ is well-defined.
		
		\item For simplicity, $ \{\bm{x}_{i}\}_{i=1}^{n} \widesim{iid} \mu $ means that $ \bm{x}_{i} \widesim{\text{iid}} \mu $ for all $ i\in \{1,\ldots,n\} $. In \eqref{eqinfsup}, we use the notation
		\begin{equation}
			\mathbb{E}_{\{\bm{x}_{i}\}_{i=1}^{n} \widesim{iid} \mu} :=\mathbb{E}_{\bm{S}_{n}\widesim{~~}\bm{\mu}^{\otimes n}},\label{bmmutens}
		\end{equation}
		where the marginal of $ \bm{\mu} $ on $\mathcal{X}$ is $ \mu $,  and $ (\bm{x}_{1},\ldots,\bm{x}_{n})\widesim{~~}\mu^{\otimes n} $. We adopt this notation since the marginal distribution on $\mathcal{X}$ is mainly relevant in our analysis.
		
	\end{itemize}
\end{remark}

\subsection{Function classes}\label{somespaces}

The classes of functions considered in this work are the following.

\begin{itemize}
	\item\textbf{H\"older continuous functions.} Let $\mathcal{H}_\alpha $ denote the class of functions satisfying the Hölder continuity condition~\eqref{holdineq}.
	
	\item \textbf{Barron regular functions.} Various definitions of Barron function classes can be found in the literature, differing slightly in their formulation. Essentially, these are functions with a bounded first-order Fourier moment, which we make explicit below (see \cite{Barron1994, NNbarronclass, petersen2021optimal}).
	
	A function $ f:[0,1]^{d-1}\to \mathbb{R} $ is said to be of Barron class with constant $ C>0 $, if there are $ c\in [-C,C] $ and a measurable function $ F:\mathbb{R}^{d-1}\to \mathbb{C} $ satisfying
	\begin{equation}
		f(\bm{z})=c+\int_{\mathbb{R}^{d-1}} (e^{i\bm{z}\cdot\bm{\xi}}-1) F(\bm{\xi})d\bm{\xi}\quad\text{and}\quad \int_{\mathbb{R}^{d-1}}\norm{\bm{\xi}}_{1}|F(\bm{\xi})|d\bm{\xi} \leq C\label{barroncondition0}
	\end{equation}
	for all $ \bm{z}\in[0,1]^{d-1} $, where $ \bm{\xi}:=(\xi_{1},\ldots,\xi_{d-1}) $ and $ \norm{\bm{\xi}}_{1}:=\sum_{j=1}^{d-1}|\xi_{j}| $. The set of all Barron functions with constant $C$ is known as the Barron space and is denoted by $\mathcal{B}_{C}$.
	
	\begin{remark}\label{Rlipbarron}
		Every Barron function $f \in \mathcal{B}_C$ is Lipschitz on $[0,1]^{d-1}$, since	
		\begin{align*}
			|f(\bm{z}) - f(\bm{w})|&=\left|\int_{\mathbb{R}^{d-1}} \left(e^{i\bm{z}\cdot\bm{\xi}}-e^{i\bm{w}\cdot\bm{\xi}}\right) F(\bm{\xi})d\bm{\xi}\right|\\
			&\leq \int_{\mathbb{R}^{d-1}} \left|e^{i\bm{z}\cdot\bm{\xi}}-e^{i\bm{w}\cdot\bm{\xi}}\right||F(\bm{\xi})|d\bm{\xi}\\
			&\leq L_{0}  \int_{\mathbb{R}^{d-1}} \left|(\bm{z}-\bm{w})\cdot\bm{\xi}\right||F(\bm{\xi})|d\bm{\xi}\\
			&\leq L\norm{\bm{z}-\bm{w}}_{2}\quad \text{for some constants}\quad L_{0},L>0,
		\end{align*}
		and all $\bm{z},\bm{w}\in [0,1]^{d-1}  $,   i.e., $f$ satisfies the condition \eqref{holdineq} with exponent $\alpha=1$.
	\end{remark}

	\item \textbf{Convex-Lipschitz functions.} We denote by $\mathcal{C}:=\mathcal{C}([0,1]^{d-1};[0,1])$ the class of all convex functions on $[0,1]^{d-1}$ taking values in  $ [0,1] $, which are uniformly Lipschitz as in \eqref{holdineq} with $\alpha=1$ (see \cite{Convexf}).

\end{itemize}

\subsection{Previous works and our contribution}\label{prev-contr}

Some related work on binary classifiers under the margin condition \ref{marginc} can be found in: \cite[Section 8]{SVM}, for support vector machines where learning rates were sometimes as fast as  $ n^{-1} $, being $ n $ the number of data points; \cite{fastc}, based on neural networks with hinge loss that achieve fast convergence rates when $ d\lesssim \gamma $, particularly as fast as $ n^{-(q+1)/(q+2)} $ when the margin exponent $\gamma \to \infty$, where $q$ is a noise parameter; and \cite{barronmarginup}, where it is found that using ReLU neural networks, the strong margin conditions imply fast learning bounds that are close to  $n^{-1}(1+\log n)$.  Furthermore, in \cite{barronmarginup, fastc}, a regularity condition is assumed on the decision boundary, where $\partial\Omega$ can be described by classes of functions $ \mathscr{C}  $ satisfying condition \ref{learseth} on the elements of a covering for the set $\Omega$. In \cite{fastc}, the class of Hölder continuous functions is considered, and in \cite{barronmarginup}, the Barron class. The last two works mentioned above only provide upper bounds for the learning rate when binary classifiers are approximated by neural networks under the margin condition. Our contribution now, by finding minimax lower bounds for learning rates on binary estimators, is to confirm that these learning rates are indeed optimal. It is important to highlight that the margin plays a main role in the fast learning rates obtained in each function space, when $\gamma$ is sufficiently large, the curse of dimensionality is overcome.

In  \cite{YangBarron}, it was shown that through an entropy notion on density spaces, it is possible to determine lower bounds for the learning rate of binary estimators.  Then, in \cite{petersen2021optimal}, a relation between distances in a specific set of densities and the norms $L^2(\lambda)$ in $H_{\mathscr{C}}$ and $L^1([0,1]^{d-1})$ in $\mathscr{C}$ was demonstrated, thus adapting the main results of  \cite{YangBarron} to function spaces $\mathscr{C}$ as in \ref{learseth}. However, the margin condition was not assumed in either \cite{petersen2021optimal} or \cite{YangBarron}.

In this paper, we establish a lower bound for the minimax expression $\mathcal{I}_{n}(\mathscr{C})$ in \eqref{eqinfsup} through Construction~\ref{mainconst}, which yields a finite subfamily $\mathscr{C}_{\Theta} \subset \mathscr{C}$ indexed by $\Theta = \{0,1\}^{m}$, the set of binary vectors of length $m$. More precisely, given an arbitrary algorithm $A \in \mathcal{A}_{n}(L^{2}(\lambda))$, we first show that the supremum in \eqref{eqinfsup} can be lower bounded by its restriction to $\mathscr{C}_{\Theta}$ and to a carefully constructed family of densities $\mathcal{F}_{\Theta}$, chosen so as to satisfy Condition~\ref{marginc} together with some additional properties. Next, by introducing a suitable projection of the estimator $A(\bm{S}_{n})$ onto the discrete set $\Theta$, we further reduce the problem to lower bounding an expression involving the Hamming distance between the projected estimator and an arbitrary element of $\Theta$. Then, after this reduction, we apply Assouad's lemma \cite[Theorem~2.12]{tsybankovnonpa} to obtain Theorem~\ref{maintheo}.

Finally, we apply Theorem~\ref{maintheo} to the three classes of functions introduced in Section~\ref{somespaces}, namely the class of Hölder continuous functions, the Barron class of functions, and the class of convex-Lipschitz functions, thereby obtaining Corollary~\ref{maincoro}. In particular, we get the following. 
 
\begin{itemize}[leftmargin=*]
	\item For the Barron class,  $\mathscr{C}=\mathcal{B}_{C}$, the lower bound in \eqref{BarrCor} is given by
	\[
	\mathcal{I}_{n}(\mathcal{B}_{C})  \gtrsim \,n^{-\frac{ \gamma}{\gamma+\left(\frac{2(d-1)}{d+1}\right)}}  \quad \text{for all}\quad  \gamma\geq 1\quad  \text{and} \quad n\in \mathbb{N},
	\]
	where $ 2(d-1)/(d+1)\to 2 $ when $ d\to\infty $. On the other hand, in \cite{barronmarginup}, upper bounds for the learning rate were established in the form 
	\[
	\mathcal{I}_{n}(\mathcal{B}_{C}) \lesssim n^{-\frac{\gamma}{\gamma+2}}(1+\log n)\quad\text{for all}\quad \gamma>0\quad  \text{and} \quad n\in \mathbb{N}.
	\]
	Therefore, our lower bound matches the upper bound in \cite{barronmarginup} up to logarithmic factors in the high-dimensional regime for $\gamma\geq 1$. Indeed, as $d\to\infty$, the exponent of $ n $ in both bounds converges to  $ -\gamma/(\gamma+2) $, showing that the learning rate obtained in \cite{barronmarginup} is asymptotically optimal as the dimension tends to infinity. 
	
	\item For the H\"older continuous class, $\mathscr{C}=\mathcal{H}_{\alpha}$, we obtain in \eqref{HolCor}, that 
	\[
		\mathcal{I}_{n}(\mathcal{H}_\alpha)\gtrsim n^{-\frac{\gamma}{\gamma+(d-1)}}  \quad \text{for all} \quad \alpha\in(0,1],\quad \gamma\geq\alpha \quad \text{and}\quad n\in \mathbb{N}.
	\]
	In contrast, under the noiseless assumption (see Remark \ref{noisere}), \cite[Theorem~$3.4$]{fastc} provides the upper bound
	\[
	\mathcal{I}_{n}(\mathcal{H}_{\alpha}) \lesssim\left(\frac{\log^{3}n}{n}\right)^{\frac{\alpha\gamma}{\alpha\gamma+(d-1)}} \quad \text{for all} \quad \alpha\in(0,1],\quad \gamma\geq1 \quad \text{and}\quad n\in \mathbb{N}.
	\]
	Comparing these two bounds, we see that for values of $\alpha$ close to $1$, the exponents of $n$ in the lower and upper bounds become similar, indicating that the corresponding learning rates are nearly the same, up to logarithmic factors. Moreover, for Lipschitz functions ($\alpha=1$), the learning rate is asymptotically optimal. However, for small values of $\alpha$, the lower and upper bounds differ substantially.

	\item For the convex-Lipschitz class, $ \mathscr{C}=\mathcal{C} $, for which in \eqref{ConvLCor}, we lower bound by
	\[
		\mathcal{I}_{n}(\mathcal{C})\gtrsim \,n^{-\frac{\gamma}{\gamma+(d-1)/2}} \quad \text{for all}\quad  \gamma\geq 1\quad  \text{and} \quad n\in \mathbb{N}.
	\]
	Compared with the lower bound for general Lipschitz functions in the previous item (when $\alpha=1$), the exponent for $ n $ in the present bound is $-\gamma/(\gamma+(d-1)/2)$ rather than $-\gamma/(\gamma+(d-1))$. This suggests that the additional convexity assumption could lead to  a faster optimal learning rate.

\end{itemize}

\section{Main results}\label{mainresult}

We begin by presenting a general result that provides a lower bound for the minimax expression \eqref{eqinfsup}. Here we make the assumption that a finite subfamily of the function class $ \mathscr{C} $ can be constructed by adding independent, localized perturbations of small amplitude to a fixed baseline function over a partition of the domain. We first describe the construction, and then state our main theorem with the corresponding lower bound.

\begin{const}\label{mainconst}
	Assume the existence of an even integer $ M\geq 2 $; a baseline function $ b_{0}\in \mathscr{C} $; a Hölder continuous function $ \varphi:\mathbb{R}^{d-1}\to \mathbb{R} $ with exponent $\alpha\in (0,1] $ and constant $ K_{\varphi}>0 $ with respect to the Euclidean norm (as in \eqref{holdineq}). Further assume that $\varphi$ is Hölder continuous at $ \bm{0} $ with exponent $\alpha  $ and constant $ C_{\varphi}>0 $ with respect to the $ \ell_{\infty} $-norm, i.e.,
	\begin{equation}
		|\varphi(\bm{z})-\varphi(\bm{0})|\leq C_{\varphi}\norm{\bm{z}}_{\infty}^{\alpha}\quad\text{for all}\quad \bm{z} \in \mathbb{R}^{d-1};\label{varphihold0}
	\end{equation}
	and a constant  $ C_{\mathscr{C}}>0 $.
	\begin{itemize}
		\item\textbf{Partition of the domain.} Let $ \Gamma_{M}:=\left\{1,2,\ldots,M\right\}^{d-1} $ be the set of all $ (d-1) $-dimensional vectors with integer entries between $1$ and $M$. Let $ s:=M/2$, $ m:=s^{d-1} $ and consider the set of vectors in $\Gamma_{M}$ with odd integer entries as  $ \Gamma:=\left\{1,3,\ldots,2s-1\right\}^{d-1} $. Since $|\Gamma| = m$, we write $ \Gamma=\left\{\bm{v}_{1},\ldots,\bm{v}_{m}\right\} $, where $\bm{v}_{1},\ldots,\bm{v}_{m}$ denote an arbitrary enumeration of the elements of $\Gamma$.
		
		We define $ P:=\{Q_{1},\ldots,Q_{m}\} $ as a family of sets forming a partition of  $ [0,1]^{d-1} $ up to boundaries\footnote{Meaning that any intersection between two sets in this family occurs only on their boundaries.}, with
		\begin{align}
			Q_{j}&:=\left\{\bm{z}\in [0,1]^{d-1} \mid \norm{\bm{z}-\bm{v}_{j}/M}_{\infty}\leq 1/M\right\}\nonumber\\
			&\,=\prod_{k=1}^{d-1}\left[\frac{v_{jk}-1}{M},\frac{v_{jk}+1}{M}\right]\quad\text{where}\quad\bm{v}_{j}=(v_{j1},v_{j2},\ldots,v_{j(d-1)}),\label{defQj}
		\end{align}
		for all $ j=1,\ldots,m $.
		
		\item \textbf{Localized perturbations. }  Let $ \varphi $ satisfy 
		\begin{equation}
			\suppf{\varphi}\subset (-1,1)^{d-1},\quad \varphi(\bm{0})=1,\quad\text{and}\quad 0\leq \varphi(\bm{z})\leq 1 \quad\text{for all}\quad \bm{z}\in \mathbb{R}^{d-1}.\label{suppphijdef}
		\end{equation}
		We define the local perturbation function $ \varphi_{j}:[0,1]^{d-1}\to [0,1] $  by
		\begin{equation}
			\varphi_{j}(\bm{z}):= \varphi\left(M(\bm{z}-\bm{v}_{j}/M)\right),\quad \text{for each}\quad j=1,\ldots,m.\label{eqvarphij}
		\end{equation}
		
		\item \textbf{Finite subfamily of $\mathscr{C}$.} Let $  \Theta:=\{0,1\}^{m} $ denote the set of binary vectors of length $ m $. For all $ \bm{\theta}\in \Theta $, define 
		\begin{equation}
			b_{\bm{\theta}}:=b_{0}+\bm{\theta}\cdot\bm{\varphi}  \quad \text{with}\quad \bm{\varphi}:= C_{\mathscr{C}}(\varphi_{1},\ldots,\varphi_{m}), \label{condtildeC}
		\end{equation} 
		and the family 
		\begin{equation}
			\mathscr{C}_{\Theta}:=\left\{ b_{\bm{\theta}} \mid  \bm{\theta}\in \Theta \right\}.\label{Cthetafam}
		\end{equation}		
		Assume that   
		\begin{equation}
		  C_{\mathscr{C}}\leq M^{-\alpha}/4 \quad\text{and}\quad	b_{0}(\bm{z})\in [C_{\mathscr{C}},1-3C_{\mathscr{C}}] \quad\text{for all}\quad \bm{z}\in[0,1]^{d-1}.\label{Rinside01}
		\end{equation}		
		
	\end{itemize} 
\end{const}

\begin{figure}[htbp]
 	\centering 
 	
 	\quad\includegraphics[width=0.48\linewidth]{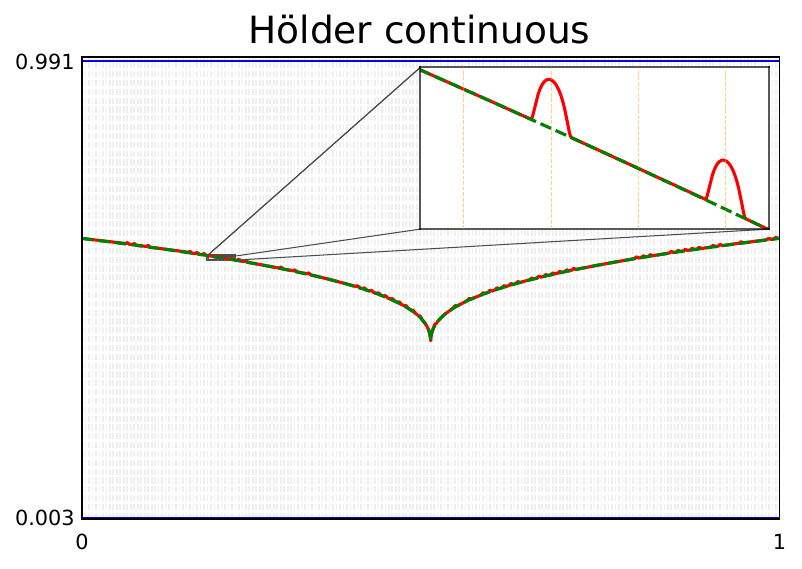}
 	\includegraphics[width=0.48\linewidth]{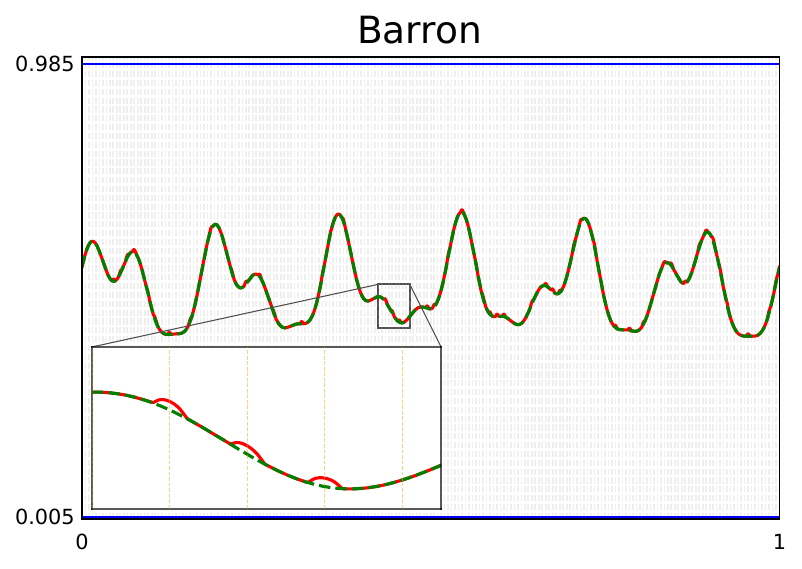}\\
 	
 	\includegraphics[width=0.55\linewidth]{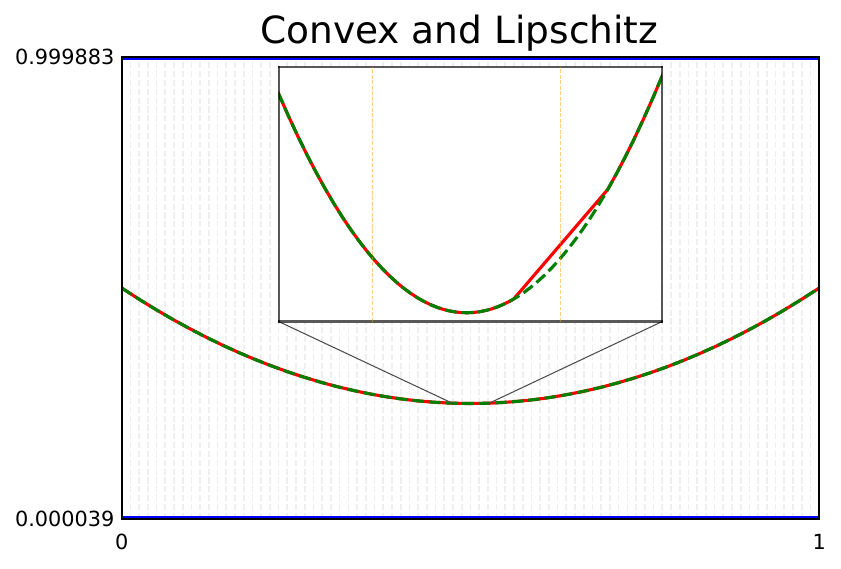}
 	
 	\caption{Examples of functions in $ \mathscr{C}_{\Theta} $ for the three function classes $ \mathscr{C} $ introduced in Section~\ref{somespaces}, with $ d=2 $ and specific choices of $ b_{0} $ and $ \varphi $ in each case. The green curves represent the baseline function $b_0$, while the red curves show a function $b_{\bm{\theta}}$ obtained by adding localized perturbations as in Construction~\ref{mainconst}. The horizontal blue lines indicate the bounds $C_{\mathscr{C}}$ and $1-3C_{\mathscr{C}}$, as required in \eqref{Rinside01}. The background grid in each plot corresponds to the partition associated with $M$ in \eqref{defQj}. Each panel also includes a magnified view to better visualize the local effect of the perturbations on the baseline function. In particular, for the H\"older continuous case, the parameters are chosen according to Subsection~\ref{subsecHolder}: $\varphi$ is defined by \eqref{vphiCorHol}, with $ \psi $ given by \eqref{partiHpsi}; the baseline function is $  b_{0}(z):=0.38+0.3|z-0.5|^{\alpha} $ with $ \alpha=0.4 $, which satisfies \eqref{Rinside01} for $ C_{\mathscr{C}} $ as in \eqref{choCcH}. For the Barron class, the parameters are chosen following Subsection~\ref{subsecbarron}: we take $ b_{0}(z)=0.5+0.15\left(0.6\sin(12\pi z)+0.3\cos(22\pi z)+0.2\sin(34\pi z)\right) $, define $ \varphi $ as in \eqref{barrvarphi} with $ \psi $ given by \eqref{partiHpsi}, and choose $ C_{\mathscr{C}} $ as in \eqref{CCfirsBar}. Lastly, for the Convex-Lipschitz case, the parameters are chosen according to Subsection~\ref{subsecconvexlips}: $ b_{0}(z)=1/4+|z-1/2|^{2} $ as in \eqref{b0convd}, $ \varphi(z)=\max\{1-4z^{2},0\} $ and $ C_{\mathscr{C}} $ as in \eqref{chconvarp}. For visualization purposes, we take $ M=200 $ in the H\"older and Barron cases, and $ M=80 $ in the Convex-Lipschitz case.} 	
 	\label{ExamElBCH}
 \end{figure}

Note that the objects used to define $ b_{\bm{\theta}} $ in \eqref{condtildeC} are required to satisfy \eqref{defQj}, \eqref{suppphijdef} and \eqref{eqvarphij}, and that the constant $ C_{\mathscr{C}} $ in \eqref{condtildeC} will be chosen in order to ensure that $ b_{\bm{\theta}} $  belongs to the class  $ \mathscr{C} $ for all  $ \bm{\theta}\in \Theta $; see Corollary \ref{maincoro} and Figure~\ref{ExamElBCH} for explicit choices. In addition, \eqref{Rinside01} is imposed for convenience in the proof of the following theorem.

\begin{theo}\label{maintheo}
	Assume conditions \ref{learseth},  \ref{conticon} with parameter $ \alpha\in (0,1] $, and \ref{marginc} with margin exponent $ \gamma\geq\alpha $. Let $ n\in \mathbb{N} $ be arbitrary, $ \widetilde{\gamma}:=\gamma/\alpha $, and consider the minimax expression $ \mathcal{I}_{n}(\mathscr{C}) $ in \eqref{eqinfsup}. If Construction \ref{mainconst} holds,
	\begin{equation}
		\mathscr{C}_{\Theta} \subseteq \mathscr{C}  \quad\text{and}\quad	2^{\widetilde{\gamma}+6}nC_{\mathscr{C}}^{\widetilde{\gamma}}M^{-(d-1)}\leq 1.\label{eqmaint0}
	\end{equation}  
	Then,  
	\begin{equation}
		\mathcal{I}_{n}(\mathscr{C})\geq \left(\frac{r^{d-1}}{8^{\widetilde{\gamma}+2}}\right) C_{\mathscr{C}}^{\widetilde{\gamma}},\label{eqmaint1}
	\end{equation} 
	where $ r:= \min\{1,(2C_{\varphi})^{-1/\alpha}\} $, and $ C_{\varphi} $ is as in \eqref{varphihold0}.
\end{theo} 


\begin{remark}
   Although $C_{\mathscr{C}}$ and $M$ appear as abstract parameters in Theorem~\ref{maintheo}, condition \eqref{eqmaint0} implicitly makes them depend on $n$.
	In applications of Theorem~\ref{maintheo}, the constant $C_{\mathscr{C}}$ is typically chosen depending on the parameter $M$ in Construction~\ref{mainconst}; see, for instance, the proof of Corollary~\ref{maincoro} in Section~\ref{seccoroproff}. Since $M$ is later chosen depending on $n$ in order to satisfy \eqref{eqmaint0}, the resulting constant $C_{\mathscr{C}}$ will in general depend on $n$ through $M$.
\end{remark}


To prove Theorem \ref{maintheo} in Section~\ref{mainteopro}, we use the subfamily $ \mathscr{C}_{\Theta} \subset \mathscr{C} $, obtained in Construction~\ref{mainconst}, in order to simplify the problem of lower bounding the minimax expression in \eqref{eqinfsup} to that of lower bounding a similar expression \eqref{eqimpsup00} (for an arbitrary algorithm $A \in \mathcal{A}_{n}(L^{2}(\lambda))$) restricted to the subfamily $ \mathscr{C}_{\Theta} $ and to a particular class of densities $ \mathcal{F}_{\Theta} $ defined in \eqref{defdenst}, which satisfy the margin condition and the properties presented in Lemmas~\ref{premainlemma}~and~\ref{lemauxtoass}. Then, under the projection~\eqref{projonT} of the estimators $ A(\bm{S}_{n}) $ onto the finite set $ \Theta $, we further lower bound \eqref{eqimpsup00}, thereby obtaining the lower bound \eqref{eqimpsup}, which depends on the Hamming distance between the projection of the estimator and the elements of $ \Theta $. Finally, we apply in \eqref{concluAss} a classical result from minimax arguments, namely Assouad's Lemma \ref{AssouadLemma}, and arrive at the lower bound \eqref{eqmaint1}.

The previous theorem provides a general lower bound under Construction~\ref{mainconst}. We now show how this construction applies to the function classes introduced in Section~\ref{somespaces}, leading to the following corollary.

\begin{coro}\label{maincoro}
	Under the assumptions of Theorem~\ref{maintheo}, the following lower bounds hold.
	\begin{enumerate}[label=\roman*)]
		\item\label{holditem} For the Hölder continuous class  $ \mathscr{C}:=\mathcal{H}_\alpha $,
		\begin{equation}
			\mathcal{I}_{n}(\mathcal{H}_\alpha)\geq  2^{-\left(\frac{5(d-1)}{\alpha}+13\widetilde{\gamma}+6\right)} n^{-\frac{\gamma}{\gamma+(d-1)}} \quad \text{for all} \quad \alpha\in(0,1],\quad \gamma\geq\alpha \quad \text{and}\quad n\in \mathbb{N}.\label{HolCor}
		\end{equation} 
		
		\item For the Barron class $  \mathscr{C}:=\mathcal{B}_{C} $,
		\begin{equation}
		\mathcal{I}_{n}(\mathcal{B}_{C})  \geq c_{d,C,\gamma} \,n^{-\frac{ \gamma}{\gamma+\left(\frac{2(d-1)}{d+1}\right)}}  \quad \text{for all} \quad  \gamma\geq1 \quad \text{and}\quad n\in \mathbb{N},	\label{BarrCor}
		\end{equation} 
		where $c_{d,C,\gamma}$ is a constant depending only on $ d $, $ C $ and $\gamma$.
		
		\item For the convex-Lipschitz class $  \mathscr{C}:=\mathcal{C}  $,
		\begin{equation}
			\mathcal{I}_{n}(\mathcal{C})\geq c_{d,\gamma} \,n^{-\frac{\gamma}{\gamma+(d-1)/2}}
			\quad \text{for all} \quad  \gamma\geq1 \quad \text{and}\quad n\in \mathbb{N},\label{ConvLCor}
		\end{equation}
		where $ c_{d,\gamma} $ is a constant that depends only on $ d $ and $ \gamma $. 
		
	\end{enumerate}
\end{coro}
 
 The proof of Corollary~\ref{maincoro} in Section~\ref{seccoroproff} consists of applying Theorem~\ref{maintheo} to each of the function classes considered in Section~\ref{somespaces}. In every case, we verify that Construction~\ref{mainconst} holds by choosing an appropriate baseline function $b_{0}$, a suitable perturbation $\varphi$, and a constant $C_{\mathscr{C}}$ such that the resulting family $\mathscr{C}_{\Theta}$ is contained in the class under consideration. We then choose the scale parameter $M$ so that condition~\eqref{eqmaint0} is satisfied. Once these hypotheses are checked, Theorem~\ref{maintheo} gives the desired lower bound through \eqref{eqmaint1}, for the Hölder, Barron, and convex-Lipschitz classes.  
\begin{remark}
	The main results, Theorem~\ref{maintheo} and Corollary~\ref{maincoro}, identify minimax lower bounds for the learning rates of binary classifiers from noiseless samples under a geometric margin condition. More precisely, Theorem~\ref{maintheo} provides a general lower bound under Construction~\ref{mainconst}, while Corollary~\ref{maincoro} applies this result to the three function classes introduced in Section~\ref{somespaces}. There are, however, some limitations of these results that should be emphasized:
	\begin{enumerate}
		\item The established rates are understood in the minimax sense, where the supremum is taken over all pairs $(h,\mu)$ satisfying the margin condition. In practice, additional favorable assumptions on the data-generating distribution may hold, and the observed learning rates may therefore be faster.
		
		\item As discussed in Remark~\ref{noisere}, the analysis is restricted to the noiseless setting. This assumption is essential in order to clarify the role of the regularity of the decision boundary and the margin condition in the lower bounds, since in the presence of noise it may be unclear whether the difficulty comes from the noise or from the geometry of the classifier (see \cite[Section 1.1, Point 1]{petersen2021optimal}). Still, noisy data appears very often in applications, and that situation is not covered by our main results.
	\end{enumerate}
\end{remark}

\begin{remark}
	The proof of Theorem~\ref{maintheo} is inspired by the use of Assouad's lemma in minimax lower bound arguments; see for instance \cite[Section~2.7.2]{tsybankovnonpa} or \cite[Proof of Theorem~3]{tsysmoothdis}. For Corollary~\ref{maincoro}, the constructions in the particular function classes are motivated by the arguments used to derive lower bounds for Kolmogorov entropy in those spaces; see for instance \cite{Clements1963}  or  \cite{1993_Tikhomirov} for the Hölder case, \cite{petersen2021optimal} for the Barron case, and \cite{Convexf} for the convex-Lipschitz case.
\end{remark}
\appendix

\section{Auxiliary results}

This section collects technical results that are used in the proof of the main results. 

\begin{figure}[htbp]
		\centering

		\includegraphics[width=0.6\linewidth]{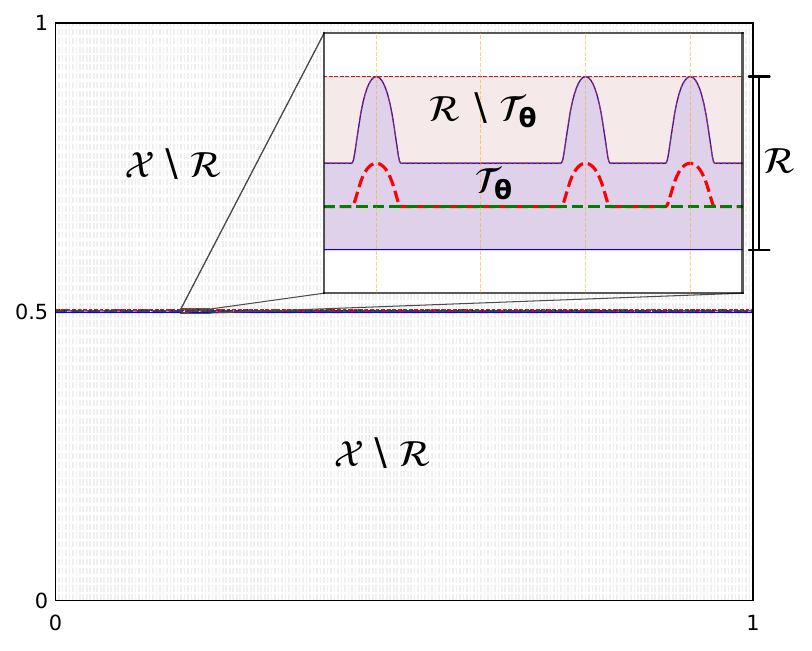} 
		
		\caption{Illustration of the regions used to define $ f_{\bm{\theta}} $ in the case $ b_{0}:=1/2 $. The figure shows the sets $ \mathcal{X}\setminus\mathcal{R} $, $ \mathcal{T}_{\bm{\theta}} $, $ \mathcal{R}\setminus\mathcal{T}_{\bm{\theta}} $, and $ \mathcal{R} $. The background grid corresponds to the partition associated with $M$ in \eqref{defQj}, and the inset provides a magnified view of the localized perturbations. For visualization purposes, we take $ M=200 $ and choose $\varphi $ as a bump function, although any function satisfying the conditions of Construction~\ref{mainconst} could be used.}
		\label{Fregionsft} 
	\end{figure}

 Under the assumption of Construction \ref{mainconst} and using the class $\mathscr{C}_{\Theta}  $ in \eqref{Cthetafam}, we define an associated family of densities as follows. For each $\bm{\theta}\in\Theta$ and $ b_{\bm{\theta}}\in \mathscr{C}_{\Theta} $, let the tube $ \mathcal{T}_{\bm{\theta}} $ around the boundary $\partial\Omega_{\bm{\theta}}$ be given by
	\begin{equation}
		\mathcal{T}_{\bm{\theta}}:=\left\{\bm{x}\in \mathcal{X}\mid |x_d-b_{\bm{\theta}}(\bm{x}^{(d)})|\leq C_{\mathscr{C}}+\bm{\theta}\cdot\bm{\varphi}(\bm{x}^{(d)}) \right\}.\label{deftubethet}
	\end{equation}
	Define the region
	\begin{equation}
		\mathcal{R}:=\left\{\bm{x}\in \mathcal{X}\mid x_{d}\in[b_{0}(\bm{x}^{(d)})-C_{\mathscr{C}},b_{0}(\bm{x}^{(d)})+3C_{\mathscr{C}}]\right\}\label{defregR}
	\end{equation}
	and the density function $ f_{\bm{\theta}}:\mathcal{X}\to [0,\infty)$ with respect to Lebesgue, as\footnote{In the case $\widetilde{\gamma}=1$, we use the convention $0^0:=1$.}
		
	\begin{equation}
		f_{\bm{\theta}}(\bm{x}):=\frac{1}{2}|x_d-b_{\bm{\theta}}(\bm{x}^{(d)})|^{\widetilde{\gamma}-1} \mathbbm{1}_{\mathcal{R}\cap\mathcal{T}_{\bm{\theta}}}(\bm{x})+ \frac{1}{2}(C_{\mathscr{C}}-\bm{\theta}\cdot\bm{\varphi}(\bm{x}^{(d)}))^{\widetilde{\gamma}-1}\mathbbm{1}_{\mathcal{R}\setminus \mathcal{T}_{\bm{\theta}}}(\bm{x})+C_{\bm{\theta}}\mathbbm{1}_{\mathcal{X}\setminus\mathcal{R}}(\bm{x}),\label{defdenst}
	\end{equation}
	where $ \widetilde{\gamma}\geq1 $ is a suitable constant, $ \bm{\varphi} $ and $ C_{\mathscr{C}} $ are as in \eqref{condtildeC}, and 
	$ C_{\bm{\theta}} $ is a normalizing constant depending only on $ \bm{\theta} $, i.e., such that $ \int_{\mathcal{X}} f_{\bm{\theta}} d\lambda =1 $. See Figure~\ref{Fregionsft} for an example of the regions defining $ f_{\bm{\theta}} $ in a particular case.	
	We denote the set of all these densities as $ \mathcal{F}_{\Theta}=\left\{ f_{\bm{\theta}} \mid  \bm{\theta}\in \Theta \right\} $. Each density $ f_{\bm{\theta}} $ induces a probability measure $ \mu_{\bm{\theta}} $, and we denote $ \mathcal{M}_{\Theta}=\left\{\mu_{\bm{\theta}} \mid  \bm{\theta}\in \Theta \right\} $ as the collection of all such measures. We write $  h_{\bm{\theta}}=h_{b_{\bm{\theta}}}  $ for all $ h_{b_{\bm{\theta}}}\in H_{\mathscr{C}_{\Theta}} $.
	
	We introduce the Hamming distance on $\Theta $, defined as
	\begin{equation}
		\rho_{Ham}(\bm{\theta},\bm{\theta}')=\#J(\bm{\theta},\bm{\theta}') \quad\text{where}\quad J(\bm{\theta},\bm{\theta}'):=\left\{j\in\{1,\ldots,m\}\mid\theta_{j}\neq \theta^{\prime}_{j}\right\}.\label{indexsetham}
	\end{equation}

    In the following lemma, we establish additional properties of the families $ \mathscr{C}_{\Theta} $, $ \mathcal{F}_{\Theta} $ and $ \mathcal{M}_{\Theta} $ that are needed in the proof of the lower bound \eqref{eqmaint1}. In particular, they provide control of the geometry of the perturbed decision boundaries, ensure that the constructed classifier-distribution pairs satisfy the margin condition in \ref{marginc}, and yield the estimates required later for the application of Assouad’s Lemma \ref{AssouadLemma}.
    
	\begin{lem}\label{premainlemma}
		Let $ \mathscr{C} $ satisfy \ref{conticon} with $ \alpha\in (0,1] $. The following statements hold.
		\begin{enumerate}[label=\arabic*)]
			\item\label{itemdishol}  For all $ b\in \mathscr{C} $ and all $\bm{x}\in \mathcal{X}$, 
			\begin{equation}
				\left(\frac{1}{\widetilde{K_{b}}}|x_{d}-b(\bm{x}^{(d)})|\right)^{\frac{1}{\alpha}}\leq \mathrm{dist}\left(\bm{x},\partial\Omega_{h_{b}}\right) \leq |x_{d}-b(\bm{x}^{(d)})|,\label{boundistofb}
			\end{equation}
			where 
			\begin{equation}
				1<\widetilde{K_{b}}:=\widetilde{K_{b}}(\alpha)=\begin{cases}
					\sqrt{1+K_{b}^{2}}	& \text{ if }\alpha=1 \\
					\max\left\{2^{\alpha},2K_{b}\right\}	& \text{ if }\alpha<1,\label{boundistofb1}
				\end{cases}
			\end{equation}
			and $ K_{b}>0 $ is a constant satisfying \eqref{holdineq}. 
			
			\item\label{itemdishol2}   For all $ j,k\in\{1,\ldots,m\} $ with $ j\neq k $, it holds (see \eqref{eqvarphij}) that 
			\begin{equation}
					\suppf{\varphi_{j}}\cap \suppf{\varphi_{k}}=\emptyset.\label{aesuppsemp}
			\end{equation}
			Furthermore, for each $ \bm{z}\in [0,1]^{d-1} $, there exists $ j \in \{1,\ldots,m\} $ such that
			\begin{equation}
				\bm{\varphi}(\bm{z})=C_{\mathscr{C}}\varphi_{j}(\bm{z}) \bm{e}_{j},\label{bCaetophi}
			\end{equation}
			where $ \bm{e}_{j} $ denotes the vector with $1$ in its $j$-th entry and $0$ in all others.  
			
			\item\label{itemdishol21} Let  $ K_{b_{0}}>0 $ be the constant satisfying \eqref{holdineq} for $ b_{0}\in \mathscr{C} $, and let $K_{\varphi} $ be the Hölder continuity constant of $ \varphi $. Then, \eqref{holdineq} is satisfied for all $ b\in \mathscr{C}_{\Theta} $ with constant 
			\begin{equation}
				K_{b}=K_{\Theta}:=K_{b_{0}} + K_{\varphi}/2.\label{KTheta}
			\end{equation}
			That is, $ \mathscr{C}_{\Theta} $ satisfies \ref{conticon} uniformly with exponent $ \alpha $ and constant $ K_{\Theta} $.

			\item\label{contenaeTR} For all $ \bm{\theta}\in \Theta $,  $ \mathcal{T}_{\bm{\theta}}\subseteq \mathcal{R} $. Moreover,
			\begin{equation}
				\mathcal{R}\setminus \mathcal{T}_{\bm{\theta}}=\left\{\bm{x}\in \mathcal{X}\mid   b_{0}(\bm{x}^{(d)})+  C_{\mathscr{C}}+2\bm{\theta}\cdot\bm{\varphi}(\bm{x}^{(d)})< x_{d} \leq b_{0}(\bm{x}^{(d)})+3C_{\mathscr{C}}\right\}.\label{idRminusT}
			\end{equation}
			
			\item\label{itemdishol3}   For each $ \bm{\theta}\in \Theta $, 
			\begin{equation}
				C_{\bm{\theta}}=\frac{1-\int_{\mathcal{R}} f_{\bm{\theta}}(\bm{x}) d\bm{x}}{1-4C_{\mathscr{C}}}\quad\text{ and }\quad\frac{1-2C_{\mathscr{C}}}{1-4C_{\mathscr{C}}}\leq C_{\bm{\theta}} \leq \frac{1}{1-4C_{\mathscr{C}}} ,\label{Cthetaeq}
			\end{equation}
			where $ C_{\bm{\theta}} $ is defined in \eqref{defdenst}.

			\item\label{itemdisho4}  If  $ \widetilde{\gamma}:=\gamma/\alpha $ is chosen in \eqref{defdenst} with $ \gamma\geq \alpha $, then every pair $ (h_{\bm{\theta}},\mu_{\bm{\theta}})\in H_{\mathscr{C}_{\Theta}}\times\mathcal{M}_{\Theta} $ satisfies the margin condition~\ref{marginc} with margin exponent $\gamma$.
			
			\item\label{itemdisho5}  For all $ \bm{\theta},\bm{\theta}'\in \Theta $ with $ \rho_{Ham}(\bm{\theta},\bm{\theta}')=1 $, we have that 
			\begin{equation} 
				\int_{\mathcal{X}} \left| f_{\bm{\theta}} - f_{\bm{\theta}'}\right| d\lambda\leq2^{\widetilde{\gamma}+3}C_{\mathscr{C}}^{\widetilde{\gamma}}M^{-(d-1)}\label{boundpHam1}
			\end{equation}
			and 
			\begin{equation}
				\int_{\left\{\bm{x}\in \mathcal{X}\mid h_{\bm{\theta}}(\bm{x})\neq h_{\bm{\theta}'}(\bm{x})\right\}}  (f_{\bm{\theta}}   + f_{\bm{\theta}'})   d\lambda\leq 2^{\widetilde{\gamma}+1}C_{\mathscr{C}}^{\widetilde{\gamma}}M^{-(d-1)}.\label{boundpHam11}
			\end{equation}	 	
			Here,  $ \rho_{Ham} $ is the Hamming distance \eqref{indexsetham}. 
		\end{enumerate}		
		
	\end{lem}
	
	\begin{proof}
		We prove each case separately.\\
				
		\noindent\ref{itemdishol} Let $ K_{b}>0 $ be a constant satisfying \eqref{holdineq}. For all $ \bm{x'}\in \partial\Omega_{h_{b}}=\left\{\bm{x}\in \mathcal{X}\mid b(\bm{x}^{(d)})=x_{d} \right\}  $ (see Remark~\ref{remhorc}), and all $ \bm{x} \in\mathcal{X} $, we know that 
		\begin{equation}
			\norm{\bm{x}-\bm{x'}}_{2}^{2}=\norm{\bm{x}^{(d)}-\bm{x'}^{(d)}}_{2}^{2}+|x_{d}-b({\bm{x'}}^{(d)})|^{2}.\label{prfnewbeq00}
		\end{equation}
		We set $  t:=\norm{\bm{x}^{(d)}-\bm{x'}^{(d)}}_{2}  $ and $ w:=|x_{d}-b({\bm{x}}^{(d)})| $, and consider two possibilities:
		\begin{itemize}
			\item If $ w-K_{b}t^{\alpha}\leq 0 $, we get $t^{2}\geq (w/K_{b})^{2/\alpha}$, and 
			\begin{equation}
				\norm{\bm{x}-\bm{x'}}_{2}^{2}=t^{2}+|x_{d}-b({\bm{x'}}^{(d)})|^{2}\geq(w/K_{b})^{2/\alpha}.\label{prfnewbeq0}
			\end{equation}
			\item If $ w-K_{b}t^{\alpha}\geq 0 $, condition \ref{conticon} implies 
			\begin{align*}
				w=|x_{d}-b({\bm{x}}^{(d)})|&\leq |x_{d}-b({\bm{x'}}^{(d)})| + |b({\bm{x}}^{(d)})-b({\bm{x'}}^{(d)})|\\
				&\leq |x_{d}-b({\bm{x'}}^{(d)})| + K_{b}\norm{{\bm{x}}^{(d)}-{\bm{x'}}^{(d)}}_{2}^{\alpha}\\
				&= |x_{d}-b({\bm{x'}}^{(d)})| + K_{b}t^{\alpha},
			\end{align*}
			therefore
			\begin{equation}
				\norm{\bm{x}-\bm{x'}}_{2}^{2}=t^{2}+|x_{d}-b({\bm{x'}}^{(d)})|^{2}\geq t^{2} + \left(w-K_{b}t^{\alpha}\right)^{2}.\label{miniexp}
			\end{equation}
			So, when $ \alpha=1 $, it follows that
			\begin{align}
				\norm{\bm{x}-\bm{x'}}_{2}^{2}&\geq(1+K_{b}^{2})t^{2} -2wK_{b}t + w^{2}\nonumber\\
				&\geq \frac{4w^{2}(1+K_{b}^{2})-(2wK_{b})^{2}}{4(1+K_{b}^{2})}\nonumber\\
				&=\frac{w^{2}}{1+K_{b}^{2}}.\label{prfnewbeq1}
			\end{align}
			Otherwise, $ \alpha<1 $ and we consider a threshold $ t_{0}:=(w/(2K_{b}))^{1/\alpha} $  to minimize the right-hand side of \eqref{miniexp} on $ t $. By cases:
			\begin{itemize}
				
				\item If  $ t\leq t_{0} $, we have that $ w-K_{b}t^{\alpha}\geq w-K_{b}t_{0}^{\alpha}=w/2>0 $, therefore 
				\begin{equation}
					\norm{\bm{x}-\bm{x'}}_{2}^{2}\geq t^{2} + \left(w-K_{b}t_{0}^{\alpha}\right)^{2}=t^{2}+(w/2)^{2}\geq (w/2)^{2}.\label{prfnewbeq2}
				\end{equation}
				
				\item If $ t>t_{0} $, we obtain
				\begin{equation}
					\norm{\bm{x}-\bm{x'}}_{2}^{2}\geq t^{2} + \left(w-K_{b}t^{\alpha}\right)^{2} \geq t^{2}>(w/(2K_{b}))^{2/\alpha}.\label{prfnewbeq3}
				\end{equation}
				
			\end{itemize}
		\end{itemize}
		Thus, \eqref{prfnewbeq0}, \eqref{prfnewbeq1}, \eqref{prfnewbeq2} and \eqref{prfnewbeq3}, imply 
		\[
		\norm{\bm{x}-\bm{x'}}_{2}\geq \left(\frac{1}{\widetilde{K_{b}}}|x_{d}-b({\bm{x}}^{(d)})|\right)^{\frac{1}{\alpha}},
		\]
		where $ \widetilde{K_{b}} $ is as in \eqref{boundistofb1}. Then, by the above inequality and identity \eqref{prfnewbeq00}, we arrive at
		\[
		\left(\frac{1}{\widetilde{K_{b}}}|x_{d}-b({\bm{x}}^{(d)})|\right)^{\frac{1}{\alpha}}\leq \inf_{\bm{x'}\in \partial\Omega_{h_{b}}}\norm{\bm{x}-\bm{x'}}_{2}\leq  |x_{d}-b({\bm{x}}^{(d)})|.
		\]
		Here, the upper bound was obtained using that $ (x_{1},\ldots,x_{d-1},b(\bm{x}^{(d)}))\in \partial\Omega_{h_{b}} $. \\


		\noindent\ref{itemdishol2} From \eqref{suppphijdef}, we know that $ \suppf{\varphi}\subset (-1,1)^{d-1}=\left\{\bm{z}\in \mathbb{R}^{d-1} \mid \norm{\bm{z}}_{\infty}< 1 \right\} $. Hence, by \eqref{defQj} and \eqref{eqvarphij},
		\begin{align*}
			\suppf{{\varphi}_{j}}&=\left\{\bm{z}\in [0,1]^{d-1} \mid M(\bm{z}-\bm{v}_{j}/M)\in\suppf{\varphi} \right\}\nonumber\\
			&\subset \left\{\bm{z}\in [0,1]^{d-1} \mid \norm{M(\bm{z}-\bm{v}_{j}/M)}_{\infty}<1 \right\}\nonumber\\
			&=\operatorname{int}Q_{j}\quad\text{for all}\quad j\in\{1,\ldots,m\},
		\end{align*}
		where $ \operatorname{int}Q_{j} $ denotes the interior of $ Q_{j} $. Therefore,
		\[
		\suppf{\varphi_{j}}\cap \suppf{\varphi_{k}}\subseteq \operatorname{int}Q_{j}\cap \operatorname{int}Q_{k}=\emptyset~\text{ for all }~ j,k\in\{1,\ldots,m\} ~\text{ with }~j\neq k,
		\]
		since $P$ is a partition of $[0,1]^{d-1}$ up to boundaries (see \eqref{defQj}). Thus, \eqref{aesuppsemp} holds, and \eqref{bCaetophi} follows from the definition of $ \bm{\varphi} $ in \eqref{condtildeC}  together with \eqref{aesuppsemp}. \\
		
		
		\noindent\ref{itemdishol21} Let $ \bm{\theta}\in \Theta $ and $ \bm{z},\bm{w}\in [0,1]^{d-1} $. By \eqref{eqvarphij} and since $ \varphi $ is Hölder continuous with exponent $\alpha $ and constant $ K_{\varphi} $, we obtain
		\begin{align*}
			|\varphi_{j}(\bm{z})-\varphi_{j}(\bm{w})|&=|\varphi\left(M(\bm{z}-\bm{v}_{j}/M)\right)-\varphi\left(M(\bm{w}-\bm{v}_{j}/M)\right)|\\
			&\leq K_{\varphi} M^{\alpha}\norm{\bm{z}-\bm{w}}_{2}^{\alpha}.
		\end{align*}
		Furthermore, \eqref{aesuppsemp} and \eqref{bCaetophi} imply that there exist $ j_{\bm{z}},j_{\bm{w}}\in \{1,\ldots,m\} $, such that  
		\[
			\bm{\varphi}(\bm{z})=C_{\mathscr{C}}\varphi_{j_{\bm{z}}}(\bm{z})\bm{e}_{j_{\bm{z}}}, \quad \bm{\varphi}(\bm{w})=C_{\mathscr{C}}\varphi_{j_{\bm{w}}}(\bm{w})\bm{e}_{j_{\bm{w}}},
		\]
		and $ \varphi_{j_{\bm{z}}}(\bm{w})=\varphi_{j_{\bm{w}}}(\bm{z})=0 $ whenever $ j_{\bm{z}}\neq j_{\bm{w}} $.  Thus,
		\begin{align}
			 C_{\mathscr{C}}^{-1}|\bm{\theta}\cdot(\bm{\varphi}(\bm{z})-\bm{\varphi}(\bm{w}))|
			&= |\bm{\theta}\cdot(\varphi_{j_{\bm{z}}}(\bm{z}) \bm{e}_{j_{\bm{z}}}-\varphi_{j_{\bm{w}}}(\bm{w}) \bm{e}_{j_{\bm{w}}})|\nonumber\\
			&\leq \begin{cases}
				|\varphi_{j}(\bm{z}) -\varphi_{j}(\bm{w})| & \text{if }~j_{\bm{w}}=j_{\bm{z}}=j\\
				|\varphi_{j_{\bm{z}}}(\bm{z}) -\varphi_{j_{\bm{z}}}(\bm{w})|+|\varphi_{j_{\bm{w}}}(\bm{z}) -\varphi_{j_{\bm{w}}}(\bm{w})| & \text{if }~j_{\bm{w}}\neq j_{\bm{z}}
			\end{cases}\nonumber\\
			&\leq 2K_{\varphi} M^{\alpha}\norm{\bm{z}-\bm{w}}_{2}^{\alpha}.\label{holbmphi}
		\end{align}
		We know that $ K_{b_{0}}>0 $ is the constant satisfying \eqref{holdineq} for $ b_{0}\in \mathscr{C} $, so using \eqref{condtildeC},  \eqref{holbmphi}, we get 
		\begin{align*}
			|b_{\bm{\theta}}(\bm{z})-b_{\bm{\theta}}(\bm{w})|&=|b_{0}(\bm{z})+\bm{\theta}\cdot\bm{\varphi}(\bm{z})-(b_{0}(\bm{w})+\bm{\theta}\cdot\bm{\varphi}(\bm{w}))|\\
			&\leq |b_{0}(\bm{z}) -b_{0}(\bm{w})|+|\bm{\theta}\cdot\bm{\varphi}(\bm{z})-\bm{\theta}\cdot\bm{\varphi}(\bm{w})|\\
			&\leq K_{b_{0}}\norm{\bm{z}-\bm{w}}_{2}^{\alpha} + |\bm{\theta}\cdot(\bm{\varphi}(\bm{z})-\bm{\varphi}(\bm{w}))|\\ 
			&\leq \left(K_{b_{0}} + 2K_{\varphi} C_{\mathscr{C}}M^{\alpha}\right)\norm{\bm{z}-\bm{w}}_{2}^{\alpha}.
		\end{align*}
		We assumed in \eqref{Rinside01} that $C_{\mathscr{C}}\leq M^{-\alpha}/4  $, then
		\[
			|b_{\bm{\theta}}(\bm{z})-b_{\bm{\theta}}(\bm{w})|\leq  \left(K_{b_{0}} + K_{\varphi}/2\right)\norm{\bm{z}-\bm{w}}_{2}^{\alpha}.
		\]
		Since $ \bm{\theta} $ is arbitrary,  we conclude that  $ \mathscr{C}_{\Theta} $ satisfies \ref{conticon} uniformly with $ \alpha $ and $ K_{\Theta} $ as in \eqref{KTheta}.\\

		
		\noindent\ref{contenaeTR} We rewrite \eqref{deftubethet} and \eqref{defregR} as
		\begin{align*}
			\mathcal{T}_{\bm{\theta}}&:=\left\{\bm{x}\in \mathcal{X}\mid b_{0}(\bm{x}^{(d)})- C_{\mathscr{C}}\leq x_d\leq b_{0}(\bm{x}^{(d)})+  C_{\mathscr{C}}+2\bm{\theta}\cdot\bm{\varphi}(\bm{x}^{(d)}) \right\}\quad\text{and}\\
			\mathcal{R}&:=\left\{\bm{x}\in \mathcal{X}\mid b_{0}(\bm{x}^{(d)})-C_{\mathscr{C}}\leq x_{d} \leq b_{0}(\bm{x}^{(d)})+3C_{\mathscr{C}}\right\}.
		\end{align*}
		Then, by  \eqref{eqvarphij} and \eqref{bCaetophi}, we get 
		\begin{equation}
			\mathcal{T}_{\bm{\theta}} =\left\{\bm{x}\in \mathcal{X}\mid b_{0}(\bm{x}^{(d)})- C_{\mathscr{C}}\leq x_d\leq b_{0}(\bm{x}^{(d)})+  C_{\mathscr{C}}+2C_{\mathscr{C}}\varphi_{j}(\bm{x}^{(d)})(\bm{\theta}\cdot \bm{e}_{j}),~\text{ for some }~ j\right\},\label{phijarg}
		\end{equation}
		and since $ \varphi_{j}(\bm{x}^{(d)})(\bm{\theta}\cdot \bm{e}_{j})\leq 1 $, we obtain
		\[
			\mathcal{T}_{\bm{\theta}}\subseteq \mathcal{R}\quad\text{for all}\quad \bm{\theta}\in\Theta.
		\]
		Moreover, from the above it is clear that \eqref{idRminusT} holds. \\

		
		\noindent\ref{itemdishol3} The constant $ C_{\bm{\theta}} $ was defined in \eqref{defdenst} such that $ \int_{\mathcal{X}} f_{\bm{\theta}} d\lambda =1 $, for all $ \bm{\theta}\in\Theta $. Therefore,
		\[
		1=\int_{\mathcal{X}} f_{\bm{\theta}}(\bm{x}) d\bm{x}
		=\int_{\mathcal{R}} f_{\bm{\theta}}(\bm{x}) d\bm{x}+\int_{\mathcal{X}\setminus\mathcal{R}} C_{\bm{\theta}} d\bm{x}
		=\int_{\mathcal{R}} f_{\bm{\theta}}(\bm{x}) d\bm{x}+C_{\bm{\theta}}(1-\lambda(\mathcal{R})).
		\]
		By \eqref{Rinside01} and \eqref{defregR}, we obtain $ \lambda(\mathcal{R})=4C_{\mathscr{C}} $ and
		\[
		C_{\bm{\theta}}=\frac{1-\int_{\mathcal{R}} f_{\bm{\theta}}(\bm{x}) d\bm{x}}{1-4C_{\mathscr{C}}}.
		\]
		Furthermore, we use \eqref{suppphijdef}, \eqref{eqvarphij}, \eqref{condtildeC}, \eqref{Rinside01} and \eqref{bCaetophi} to see that
		\begin{align}
			b_{\bm{\theta}}(\bm{z})&=b_{0}(\bm{z})+C_{\mathscr{C}}\varphi_{j}(\bm{z})(\bm{\theta}\cdot \bm{e}_{j})\in [0,1]\quad\text{and}\label{btCphij}\\
			(C_{\mathscr{C}}-\bm{\theta}\cdot\bm{\varphi}(\bm{z}))^{\widetilde{\gamma}-1}&=C_{\mathscr{C}}^{\widetilde{\gamma}-1}(1-\varphi_{j}(\bm{z})(\bm{\theta}\cdot \bm{e}_{j}))^{\widetilde{\gamma}-1}\in[0,1],\label{btCphij2}
		\end{align}
		for each $ \bm{z}\in [0,1]^{d-1} $, for some $ j=j(\bm{z})\in \{1,\ldots,m\} $ and for all $\widetilde{\gamma}\geq 1$. 
		Then, \eqref{defdenst} implies 
		\begin{align*}
			0\leq f_{\bm{\theta}}\mathbbm{1}_{\mathcal{R}}(\bm{x})&=\frac{1}{2}\left(|x_d-b_{\bm{\theta}}(\bm{x}^{(d)})|^{\widetilde{\gamma}-1} \mathbbm{1}_{\mathcal{R}\cap\mathcal{T}_{\bm{\theta}}}(\bm{x})+ (C_{\mathscr{C}}-\bm{\theta}\cdot\bm{\varphi}(\bm{x}^{(d)}))^{\widetilde{\gamma}-1}\mathbbm{1}_{\mathcal{R}\setminus \mathcal{T}_{\bm{\theta}}}(\bm{x})\right)\nonumber\\
			&\leq \frac{1}{2}\left( \mathbbm{1}_{\mathcal{R}\cap\mathcal{T}_{\bm{\theta}}}(\bm{x})+  \mathbbm{1}_{\mathcal{R}\setminus \mathcal{T}_{\bm{\theta}}}(\bm{x})\right)\nonumber\\
			&=\frac{1}{2}\mathbbm{1}_{\mathcal{R}}(\bm{x}),
		\end{align*}
		and 
		\[
		1-2C_{\mathscr{C}}=1-\lambda(\mathcal{R})/2\leq 1-\int_{\mathcal{R}} f_{\bm{\theta}}(\bm{x}) d\bm{x}\leq 1.
		\]
		Hence,  
		\[
			\frac{1-2C_{\mathscr{C}}}{1-4C_{\mathscr{C}}}\leq \frac{1-\int_{\mathcal{R}} f_{\bm{\theta}}(\bm{x}) d\bm{x}}{1-4C_{\mathscr{C}}}\leq \frac{1}{1-4C_{\mathscr{C}}}
		\]
		and \eqref{Cthetaeq} holds.\\
		


		\noindent\ref{itemdisho4} Let $ \widetilde{\gamma}:=\gamma/\alpha $ in \eqref{defdenst} with $ \gamma\geq \alpha $, and let $ \bm{\theta}\in \Theta $ be arbitrary. By items  \ref{itemdishol} and \ref{itemdishol21} above, we get 
		\[
			\left(\frac{1}{\widetilde{K_{\Theta}}}|x_{d}-b_{\bm{\theta}}(\bm{x}^{(d)})|\right)^{\frac{1}{\alpha}}\leq \mathrm{dist}\left(\bm{x},\partial\Omega_{h_{\bm{\theta}}}\right).
		\] 
		Therefore, with the notation in condition \ref{marginc}, we obtain
		\begin{align}
			B_{\varepsilon}^{h_{\bm{\theta}}}&=\left\{\bm{x}\in\mathcal{X}\mid \mathrm{dist}\left(\bm{x},\partial\Omega_{h_{\bm{\theta}}}\right)\leq \varepsilon\right\}\nonumber\\
			&\subseteq \left\{\bm{x}\in\mathcal{X}\mid|x_{d}-b_{\bm{\theta}}(\bm{x}^{(d)})|\leq \widetilde{K_{\Theta}}\varepsilon^{\alpha} \right\},\quad\text{for all}\quad \varepsilon>0.\label{contthetait5}
		\end{align}		
		By the previous item \ref{contenaeTR},  $ \mathcal{T}_{\bm{\theta}}\subseteq \mathcal{R} $, and therefore 
		\begin{equation}
			\mathcal{R}\cap\mathcal{T}_{\bm{\theta}}=\mathcal{T}_{\bm{\theta}}.\label{RcapTae}
		\end{equation}			
		So, we consider the following cases.
		\begin{itemize}
			\item If $ \widetilde{K_{\Theta}}\varepsilon^{\alpha}\leq C_{\mathscr{C}} $,	then \eqref{deftubethet}, \eqref{RcapTae} and \eqref{contthetait5}  imply $ B_{\varepsilon}^{h_{\bm{\theta}}}\subseteq \mathcal{T}_{\bm{\theta}}=\mathcal{R}\cap\mathcal{T}_{\bm{\theta}} $. Therefore,
			\begin{align}
				\mu_{\bm{\theta}}(B_{\varepsilon}^{h_{\bm{\theta}}})&= \mu_{\bm{\theta}}((\mathcal{R}\cap\mathcal{T}_{\bm{\theta}})\cap B_{\varepsilon}^{h_{\bm{\theta}}})\nonumber\\
				&=\int_{B_{\varepsilon}^{h_{\bm{\theta}}}} f_{\bm{\theta}} \mathbbm{1}_{\mathcal{R}\cap\mathcal{T}_{\bm{\theta}}} d\lambda\nonumber\\
				&\leq \frac{1}{2}\int_{|x_{d}-b_{\bm{\theta}}(\bm{x}^{(d)})|\leq \widetilde{K_{\Theta}}\varepsilon^{\alpha}} |x_d-b_{\bm{\theta}}(\bm{x}^{(d)})|^{\widetilde{\gamma}-1} d\bm{x}\nonumber\\
				&\leq \frac{1}{2}(\widetilde{K_{\Theta}}\varepsilon^{\alpha})^{\widetilde{\gamma}-1} \int_{[0,1]^{d-1}}\int_{|x_{d}-b_{\bm{\theta}}(\bm{x}^{(d)})|\leq \widetilde{K_{\Theta}}\varepsilon^{\alpha}} 1  dx_d d\bm{x}^{(d)}\nonumber\\
				&\leq (\widetilde{K_{\Theta}}\varepsilon^{\alpha})^{\widetilde{\gamma}}\nonumber\\
				&= \widetilde{K_{\Theta}}^{\widetilde{\gamma}} \varepsilon^{\gamma}.\label{cmargp}
			\end{align}
			The inequality above shows that condition \ref{marginc} is satisfied with constant $ \widetilde{K_{\Theta}}^{\widetilde{\gamma}} $ and margin exponent $ \gamma $.
			
			\item If $ \widetilde{K_{\Theta}}\varepsilon^{\alpha}> C_{\mathscr{C}} $. We have, similarly to \eqref{cmargp}, that
			\begin{align*}
				\mu_{\bm{\theta}}(B_{\varepsilon}^{h_{\bm{\theta}}})&=\int_{B_{\varepsilon}^{h_{\bm{\theta}}}} f_{\bm{\theta}} \mathbbm{1}_{\mathcal{R}\cap\mathcal{T}_{\bm{\theta}}} d\lambda+\int_{B_{\varepsilon}^{h_{\bm{\theta}}}} f_{\bm{\theta}} \mathbbm{1}_{\mathcal{X}\setminus(\mathcal{R}\cap\mathcal{T}_{\bm{\theta}})} d\lambda\\
				&\leq \widetilde{K_{\Theta}}^{\widetilde{\gamma}} \varepsilon^{\gamma} +\int_{\mathcal{X}\setminus(\mathcal{R}\cap\mathcal{T}_{\bm{\theta}})} f_{\bm{\theta}}  d\lambda.
			\end{align*}
			By \eqref{defdenst},
			\[
			\int_{\mathcal{X}\setminus(\mathcal{R}\cap\mathcal{T}_{\bm{\theta}})} f_{\bm{\theta}}  d\lambda=  \int_{\mathcal{X}} \left(\frac{1}{2}(C_{\mathscr{C}}-\bm{\theta}\cdot\bm{\varphi}(\bm{x}^{(d)}))^{\widetilde{\gamma}-1}\mathbbm{1}_{\mathcal{R}\setminus \mathcal{T}_{\bm{\theta}}}(\bm{x})+C_{\bm{\theta}}\mathbbm{1}_{\mathcal{X}\setminus\mathcal{R}}(\bm{x})\right) d\bm{x}.
			\]
			Using \eqref{Cthetaeq},  we obtain  
			\begin{align*}
				\mu_{\bm{\theta}}(B_{\varepsilon}^{h_{\bm{\theta}}})&\leq \widetilde{K_{\Theta}}^{\widetilde{\gamma}} \varepsilon^{\gamma}+\int_{\mathcal{X}} \left(\frac{1}{2}(C_{\mathscr{C}}-\bm{\theta}\cdot\bm{\varphi}(\bm{x}^{(d)}))^{\widetilde{\gamma}-1}\mathbbm{1}_{\mathcal{R}\setminus \mathcal{T}_{\bm{\theta}}}(\bm{x})+C_{\bm{\theta}}\mathbbm{1}_{\mathcal{X}\setminus\mathcal{R}}(\bm{x})\right) d\bm{x}\\
				&\leq \widetilde{K_{\Theta}}^{\widetilde{\gamma}} \varepsilon^{\gamma}+\frac{1}{2}C_{\mathscr{C}}^{\widetilde{\gamma}-1}\int_{\mathcal{R}\setminus \mathcal{T}_{\bm{\theta}}} 1 d\bm{x}+(1-4C_{\mathscr{C}})^{-1}\int_{\mathcal{X}\setminus\mathcal{R}} 1 d\bm{x}\\
				&\leq \widetilde{K_{\Theta}}^{\widetilde{\gamma}} \varepsilon^{\gamma}+\frac{1}{2}C_{\mathscr{C}}^{\widetilde{\gamma}-1}+(1-4C_{\mathscr{C}})^{-1}.
			\end{align*}
			So, we use $  \varepsilon^{\gamma}> (C_{\mathscr{C}}/\widetilde{K_{\Theta}})^{\gamma/\alpha} $ in the above inequality and set
			\[
			C_{0}:=\left(\widetilde{K_{\Theta}}^{\widetilde{\gamma}} +\left(\frac{1}{2}C_{\mathscr{C}}^{\widetilde{\gamma}-1}+(1-4C_{\mathscr{C}})^{-1}\right)(C_{\mathscr{C}}/\widetilde{K_{\Theta}})^{-\widetilde{\gamma}}\right),
			\]
			to obtain  
			\[
			\mu_{\bm{\theta}}(B_{\varepsilon}^{h_{\bm{\theta}}})\leq C_{0}\varepsilon^{\gamma}.
			\]
			Thus, we conclude that \ref{marginc} is fulfilled with constant $ C_{0} $ and margin exponent $ \gamma $.
		\end{itemize} 
		In both cases above, for all $ (h_{\bm{\theta}},\mu_{\bm{\theta}})\in H_{\mathscr{C}_{\Theta}}\times\mathcal{M}_{\Theta} $, the margin condition holds with the same margin exponent $ \gamma $ and in particular for every constant $ C\geq C_{0} $.\\
		

		\noindent\ref{itemdisho5} For all $ \bm{\theta},\bm{\theta}'\in \Theta $ with  $ \rho_{Ham}(\bm{\theta},\bm{\theta}')=1 $, the index set $ J:=J(\bm{\theta},\bm{\theta}') $ defined in \eqref{indexsetham} contains exactly one element, i.e., $ \# J=1 $. Moreover, there exists a unique $ j^{*}\in \{1,\ldots,m\} $, such that $ J=\{j^{*}\} $. We assume without loss of generality that $ \bm{\theta}-\bm{\theta}'=\bm{e}_{j^{*}} $. Then, by \eqref{eqvarphij} and \eqref{condtildeC}, we get 
		\begin{align*}
			\bm{\theta}\cdot\bm{\varphi}(\bm{z})-\bm{\theta}'\cdot\bm{\varphi}(\bm{z})&=(\bm{\theta}-\bm{\theta}')\cdot\bm{\varphi}(\bm{z})\\
			&=\bm{e}_{j^{*}}\cdot\bm{\varphi}(\bm{z})\\
			&=C_{\mathscr{C}}\left(\bm{e}_{j^{*}}\cdot(\varphi_{1}(\bm{z}),\ldots,\varphi_{m}(\bm{z}))\right)\\
			&=C_{\mathscr{C}}\varphi_{j^{*}}(\bm{z}),\quad\text{for all}\quad  \bm{z}\in [0,1]^{d-1}.
		\end{align*}
		Therefore,
		\begin{equation}
			\bm{\theta}\cdot\bm{\varphi}(\bm{z})=\bm{\theta}'\cdot\bm{\varphi}(\bm{z})\quad \text{for all}\quad \bm{z}\in [0,1]^{d-1}\setminus\suppf{{\varphi}_{j^{*}}},\label{eqoutQj}
		\end{equation}
		and
		\begin{equation}
			f_{\bm{\theta}}\mathbbm{1}_{\mathcal{R}}(\bm{x})=f_{\bm{\theta}'}\mathbbm{1}_{\mathcal{R}}(\bm{x})\quad\text{for all}\quad \bm{x}\in ([0,1]^{d-1}\setminus\suppf{{\varphi}_{j^{*}}})\times [0,1],\label{f1f2eqinR}
		\end{equation}
		since \eqref{deftubethet}, \eqref{defdenst}, \eqref{idRminusT} and \eqref{Cthetaeq} depend only on the expression $ \bm{\theta}\cdot\bm{\varphi} $ in the region $ \mathcal{R} $.
		
		By \eqref{Rinside01}, \eqref{defregR}, \eqref{defdenst} and \eqref{Cthetaeq}, we see that 
		\begin{align*}
			\int_{\mathcal{X}\setminus\mathcal{R}} \left| f_{\bm{\theta}} - f_{\bm{\theta}'}\right| d\lambda&=
			\left|C_{\bm{\theta}}-C_{\bm{\theta}'}\right|\int_{\mathcal{X}\setminus\mathcal{R}} 1 d\lambda\\
			&=
			\frac{1-\lambda(\mathcal{R})}{1-4C_{\mathscr{C}}} \left|1-\int_{\mathcal{R}} f_{\bm{\theta}} d\lambda-\left(1-\int_{\mathcal{R}} f_{\bm{\theta}'} d\lambda\right)\right|\\
			&=
			\left|\int_{\mathcal{R}} f_{\bm{\theta}'} d\lambda-\int_{\mathcal{R}} f_{\bm{\theta}} d\lambda\right| \\
			&\leq 
			\int_{\mathcal{R}}  \left| f_{\bm{\theta}} - f_{\bm{\theta}'}\right| d\lambda.
		\end{align*}
		Then, by the above inequality and \eqref{f1f2eqinR},
		\begin{align} 
			\int_{\mathcal{X} } \left| f_{\bm{\theta}}  - f_{\bm{\theta}'} \right|d\lambda 
			&= \int_{\mathcal{R}\cup(\mathcal{X}\setminus\mathcal{R})} \left| f_{\bm{\theta}}  - f_{\bm{\theta}'} \right|d\lambda\nonumber\\			
			&\leq 2\int_{\mathcal{R}}  \left| f_{\bm{\theta}} - f_{\bm{\theta}'}\right| d\lambda\nonumber\\
			&= 2\int_{\mathcal{R}\cap\mathcal{S}_{j^{*}}} \left| f_{\bm{\theta}} - f_{\bm{\theta}'}\right| d\lambda\quad \text{where}\quad \mathcal{S}_{j^{*}}:=\suppf{{\varphi}_{j^{*}}}\times[0,1].\label{lastlemH1}
		\end{align}
		To analyze the upper bound \eqref{lastlemH1}, we look closely at the set $ \mathcal{R}\cap\mathcal{S}_{j^{*}} $ as well as the behavior of $ f_{\bm{\theta}} $ and $ f_{\bm{\theta}'} $ in this region. Note that
		\begin{equation}
			\mathcal{R}\cap\mathcal{S}_{j^{*}}=\mathcal{U}_{\bm{\theta}''}^{(1)}\cup \mathcal{U}_{\bm{\theta}''}^{(2)}\quad\text{with}\quad \mathcal{U}_{\bm{\theta}''}^{(1)}\cap \mathcal{U}_{\bm{\theta}''}^{(2)}=\emptyset,\label{U1U2}
		\end{equation}
		where
		\[
		\quad\mathcal{U}_{\bm{\theta}''}^{(1)}:=((\mathcal{R}\cap\mathcal{T}_{\bm{\theta}''})\cap\mathcal{S}_{j^{*}})\quad\text{and}\quad \mathcal{U}_{\bm{\theta}''}^{(2)}:=((\mathcal{R}\setminus \mathcal{T}_{\bm{\theta}''})\cap\mathcal{S}_{j^{*}})\quad\text{for all}\quad \bm{\theta}''\in \Theta.
		\]
		Since we assumed $ \bm{\theta}-\bm{\theta}'=\bm{e}_{j^{*}} $, we have $ \theta_{j^{*}}=1 $ and $ \theta'_{j^{*}}=0 $. So, we consider the following two cases.		
		\begin{itemize}
			\item For $ \bm{\theta} $: We use that $ \theta_{j^{*}}=1 $, \eqref{aesuppsemp} and \eqref{RcapTae} to obtain
			\begin{align}
				\mathcal{U}_{\bm{\theta}}^{(1)}&=\mathcal{T}_{\bm{\theta}}\cap\mathcal{S}_{j^{*}}\nonumber\\
				&=\left\{\bm{x}\in \mathcal{S}_{j^{*}}\mid |x_d-b_{\bm{\theta}}(\bm{x}^{(d)})|\leq C_{\mathscr{C}}+\bm{\theta}\cdot\bm{\varphi}(\bm{x}^{(d)}) \right\}\nonumber\\
				&= \left\{\bm{x}\in \mathcal{S}_{j^{*}} \mid |x_d-			
				(b_{0}(\bm{x}^{(d)})+C_{\mathscr{C}}(\theta_{j^{*}}\varphi_{j^{*}}(\bm{x}^{(d)})))			
				|\leq C_{\mathscr{C}}+  C_{\mathscr{C}}(\theta_{j^{*}}\varphi_{j^{*}}(\bm{x}^{(d)})) \right\}\nonumber\\
				&=\left\{\bm{x}\in \mathcal{S}_{j^{*}} \mid |x_d-			
				b_{0}(\bm{x}^{(d)})-C_{\mathscr{C}}\varphi_{j^{*}}(\bm{x}^{(d)})	
				|\leq C_{\mathscr{C}}\left(1+  \varphi_{j^{*}}(\bm{x}^{(d)})\right) \right\}.\label{lastlemH2}
			\end{align}
			By \eqref{aesuppsemp} and \eqref{idRminusT}, we have
			\begin{equation}
				\mathcal{U}_{\bm{\theta}}^{(2)}= \left\{\bm{x}\in \mathcal{S}_{j^{*}}:  b_{0}(\bm{x}^{(d)})+C_{\mathscr{C}}+2C_{\mathscr{C}}\varphi_{j^{*}}(\bm{x}^{(d)})	 \leq x_{d}\leq b_{0}(\bm{x}^{(d)})+3C_{\mathscr{C}}  \right\}.\label{lastlemH4}
			\end{equation}	
			Moreover, \eqref{defdenst} and \eqref{aesuppsemp} imply
			\begin{align}
				f_{\bm{\theta}}\mathbbm{1}_{\mathcal{U}_{\bm{\theta}}^{(1)}}(\bm{x})&=\frac{1}{2}|x_d-b_{0}(\bm{x}^{(d)})-C_{\mathscr{C}}\varphi_{j^{*}}(\bm{x}^{(d)})	|^{\widetilde{\gamma}-1}\mathbbm{1}_{\mathcal{U}_{\bm{\theta}}^{(1)}}(\bm{x})\quad\text{and}\nonumber\\
				f_{\bm{\theta}}\mathbbm{1}_{\mathcal{U}_{\bm{\theta}}^{(2)}}(\bm{x})&=\frac{1}{2}C_{\mathscr{C}}^{\widetilde{\gamma}-1}(1-\varphi_{j^{*}}(\bm{x}^{(d)}))^{\widetilde{\gamma}-1}\mathbbm{1}_{\mathcal{U}_{\bm{\theta}}^{(2)}}(\bm{x}).\label{lastlemH401}
			\end{align}

			\item For $ \bm{\theta}' $: We know that $ \theta'_{j^{*}}=0 $ and analogously, 
			\begin{align}
				\mathcal{U}_{\bm{\theta}'}^{(1)}&=\mathcal{T}_{\bm{\theta}'}\cap\mathcal{S}_{j^{*}}
				= \left\{\bm{x}\in \mathcal{S}_{j^{*}} \mid |x_d-			
				b_{0}(\bm{x}^{(d)})			
				|\leq C_{\mathscr{C}} \right\},\nonumber\\
				\mathcal{U}_{\bm{\theta}'}^{(2)}&=\left\{\bm{x}\in \mathcal{S}_{j^{*}}:  b_{0}(\bm{x}^{(d)})+C_{\mathscr{C}} \leq x_{d}\leq b_{0}(\bm{x}^{(d)})+3C_{\mathscr{C}}  \right\},\nonumber\\
				f_{\bm{\theta}'}\mathbbm{1}_{\mathcal{U}_{\bm{\theta}'}^{(1)}}(\bm{x})&=\frac{1}{2}|x_d-b_{0}(\bm{x}^{(d)})|^{\widetilde{\gamma}-1}\mathbbm{1}_{\mathcal{U}_{\bm{\theta}'}^{(1)}}(\bm{x})\quad\text{and}\nonumber\\
				f_{\bm{\theta}'}\mathbbm{1}_{\mathcal{U}_{\bm{\theta}'}^{(2)}}(\bm{x})&=\frac{1}{2}C_{\mathscr{C}}^{\widetilde{\gamma}-1}\mathbbm{1}_{\mathcal{U}_{\bm{\theta}'}^{(2)}}(\bm{x}).\label{lastlemH5}
			\end{align}			
			\end{itemize}			
			Thus, by \eqref{lastlemH1} and  \eqref{U1U2},			
			\begin{align} 
				\frac{1}{2}\int_{\mathcal{X} } \left| f_{\bm{\theta}}  - f_{\bm{\theta}'} \right|d\lambda&\leq \int_{\mathcal{R}\cap\mathcal{S}_{j^{*}}}  f_{\bm{\theta}} d\lambda+\int_{\mathcal{R}\cap\mathcal{S}_{j^{*}}}f_{\bm{\theta}'} d\lambda,\nonumber\\
				&=\int_{\mathcal{U}_{\bm{\theta}}^{(1)}}  f_{\bm{\theta}} d\lambda+\int_{\mathcal{U}_{\bm{\theta}}^{(2)}} f_{\bm{\theta}} d\lambda+\int_{\mathcal{U}_{\bm{\theta}'}^{(1)}}  f_{\bm{\theta}'} d\lambda+\int_{\mathcal{U}_{\bm{\theta}'}^{(2)}}  f_{\bm{\theta}'} d\lambda,\label{lastlemH12}
			\end{align}
			where using \eqref{lastlemH1}, \eqref{U1U2}, \eqref{lastlemH2}, \eqref{lastlemH4},  \eqref{lastlemH401}, and \eqref{lastlemH5}, we obtain first
			\begin{flalign}
				\bullet\;\int_{\mathcal{U}_{\bm{\theta}}^{(1)}}  f_{\bm{\theta}} d\lambda&\leq \frac{1}{2}\int_{|x_d-			
					b_{0}(\bm{x}^{(d)})-C_{\mathscr{C}}\varphi_{j^{*}}(\bm{x}^{(d)})	
					|\leq C_{\mathscr{C}}\left(1+  \varphi_{j^{*}}(\bm{x}^{(d)})\right)}  |x_d-b_{0}(\bm{x}^{(d)})-C_{\mathscr{C}}\varphi_{j^{*}}(\bm{x}^{(d)})	|^{\widetilde{\gamma}-1} d\bm{x}&&\nonumber\\ 
				&= \frac{1}{\widetilde{\gamma}}\int_{[0,1]^{d-1}}
				C_{\mathscr{C}}^{\widetilde{\gamma}}		
				\left( 1+  \varphi_{j^{*}}(\bm{x}^{(d)})
				\right)^{\widetilde{\gamma}} d\bm{x}^{(d)}&&\nonumber\\
				&= \frac{1}{\widetilde{\gamma}}\int_{[0,1]^{d-1}}
				C_{\mathscr{C}}^{\widetilde{\gamma}}		
				\left(1+ 				\varphi\left(M(\bm{z}-\bm{v}_{j^{*}}/M)\right)
				\right)^{\widetilde{\gamma}} d\bm{z}&&\nonumber\\
				&= \frac{1}{\widetilde{\gamma}}C_{\mathscr{C}}^{\widetilde{\gamma}}	 M^{-(d-1)}\norm{1+ \varphi}_{L^{\widetilde{\gamma}}([0,1]^{d-1})}^{\widetilde{\gamma}}.&&\label{U1fthtalb}
			\end{flalign}
			Next,
			\begin{flalign}
				\bullet\; \int_{\mathcal{U}_{\bm{\theta}'}^{(1)}}  f_{\bm{\theta}'} d\lambda&=\frac{1}{2}\int_{\mathcal{U}_{\bm{\theta}'}^{(1)}}  |x_d-b_{0}(\bm{x}^{(d)})|^{\widetilde{\gamma}-1} d\bm{x}&&\nonumber\\
				&=\frac{1}{2}\int_{\suppf{{\varphi}_{j^{*}}}}\int_{|x_d-			
					b_{0}(\bm{x}^{(d)})			
					|\leq C_{\mathscr{C}}}  |x_d-b_{0}(\bm{x}^{(d)})|^{\widetilde{\gamma}-1} dx_d d\bm{x}^{(d)}\nonumber\\
				&=\frac{1}{\widetilde{\gamma}}C_{\mathscr{C}}^{\widetilde{\gamma}}\int_{\left\{\bm{z}\in [0,1]^{d-1} \mid M(\bm{z}-\bm{v}_{j^{*}}/M)\in\suppf{\varphi} \right\}} 1 d\bm{z}&& \nonumber\\
				&=\frac{1}{\widetilde{\gamma}}C_{\mathscr{C}}^{\widetilde{\gamma}}M^{-(d-1)}\int_{\suppf{\varphi}} 1 d\bm{z} &&\nonumber\\
				&\leq \frac{1}{\widetilde{\gamma}}C_{\mathscr{C}}^{\widetilde{\gamma}}M^{-(d-1)}.&& \label{U1fthtaprilb}
			\end{flalign}			
			Moreover,
			\begin{flalign*}
				\bullet\;\int_{\mathcal{U}_{\bm{\theta}}^{(2)}} f_{\bm{\theta}} d\lambda&=\frac{1}{2}\int_{\mathcal{U}_{\bm{\theta}}^{(2)}} C_{\mathscr{C}}^{\widetilde{\gamma}-1}(1-\varphi_{j^{*}}(\bm{x}^{(d)}))^{\widetilde{\gamma}-1} d\bm{x}&&\\
				&\leq \int_{[0,1]^{d-1}} C_{\mathscr{C}}^{\widetilde{\gamma}}(1-\varphi_{j^{*}}(\bm{x}^{(d)}))^{\widetilde{\gamma}}  d\bm{x}^{(d)}&&\\
				&= C_{\mathscr{C}}^{\widetilde{\gamma}}	 M^{-(d-1)}\norm{1- \varphi}_{L^{\widetilde{\gamma}}([0,1]^{d-1})}^{\widetilde{\gamma}}.&&
			\end{flalign*}
			Finally,
			\begin{flalign*}
				\bullet\;\int_{\mathcal{U}_{\bm{\theta}'}^{(2)}}  f_{\bm{\theta}'} d\lambda&=\frac{1}{2}\int_{\mathcal{U}_{\bm{\theta}'}^{(2)}}  C_{\mathscr{C}}^{\widetilde{\gamma}-1} d\bm{x}&&\\
				&\leq C_{\mathscr{C}}^{\widetilde{\gamma}} M^{-(d-1)}.&&
			\end{flalign*} 
			Therefore, combining the above bounds on the four integrals with \eqref{lastlemH12}, we obtain
			\begin{align*}
				\frac{1}{2}\int_{\mathcal{X} } \left| f_{\bm{\theta}}  - f_{\bm{\theta}'} \right|d\lambda&\leq C_{\mathscr{C}}^{\widetilde{\gamma}} M^{-(d-1)}\left(\frac{1}{\widetilde{\gamma}}\norm{1+ \varphi}_{L^{\widetilde{\gamma}}([0,1]^{d-1})}^{\widetilde{\gamma}}+ \norm{1- \varphi}_{L^{\widetilde{\gamma}}([0,1]^{d-1})}^{\widetilde{\gamma}}+\frac{1}{\widetilde{\gamma}} +1\right)\\
				&\leq 2^{\widetilde{\gamma}+2}C_{\mathscr{C}}^{\widetilde{\gamma}} M^{-(d-1)}
			\end{align*}
			and \eqref{boundpHam1} holds.
			
			Next, to prove \eqref{boundpHam11} we see that \eqref{bcont} implies 
			\begin{align}
				G_{\bm{\theta},\bm{\theta}'}&:=\left\{\bm{x}\in \mathcal{X}  \mid h_{\bm{\theta}}(\bm{x}) \neq h_{\bm{\theta}' }(\bm{x})\right\}\nonumber\\
				&=\left\{\bm{x}\in \mathcal{X}  \mid h_{\bm{\theta}}(\bm{x})=0 \land h_{\bm{\theta}' }(\bm{x})=1\right\}\cup \left\{\bm{x}\in \mathcal{X}  \mid h_{\bm{\theta}}(\bm{x})=1 \land h_{\bm{\theta}' }(\bm{x})=0\right\}\nonumber\\
				&=\left\{\bm{x}\in \mathcal{X}  \mid   b_{\bm{\theta}' }(\bm{x}^{(d)})\leq x_{d}< b_{\bm{\theta}}(\bm{x}^{(d)}) \right\}\cup \left\{ \bm{x}\in \mathcal{X}  \mid  b_{\bm{\theta}}(\bm{x}^{(d)})\leq x_{d}< b_{\bm{\theta}' }(\bm{x}^{(d)}) \right\}\nonumber\\
				&=\left\{ \bm{x}\in \mathcal{X}  \mid \min \left\{b_{\bm{\theta}}(\bm{x}^{(d)}), b_{\bm{\theta}' }(\bm{x}^{(d)}) \right\}\leq x_{d}< \max\left\{b_{\bm{\theta}}(\bm{x}^{(d)}), b_{\bm{\theta}' }(\bm{x}^{(d)}) \right\}\right\}.\label{G12subset}
			\end{align}			
			Using \eqref{btCphij}, we obtain $ b_{0}(\bm{z})\leq b_{\bm{\theta}''}(\bm{z})\leq b_{0}(\bm{z})+C_{\mathscr{C}} $ for all $ \bm{z}\in [0,1]^{d-1} $ and all $ \bm{\theta}''\in \Theta $. 
			Thus,
			\begin{align}
				G_{\bm{\theta},\bm{\theta}'}&\subseteq\left\{\bm{x}\in \mathcal{X}  \mid   b_{0}(\bm{x}^{(d)}) \leq x_{d}\leq b_{0}(\bm{x}^{(d)})+C_{\mathscr{C}} \right\}\nonumber\\
				&\subset \left\{\bm{x}\in \mathcal{X}  \mid   b_{0}(\bm{x}^{(d)})-C_{\mathscr{C}} \leq x_{d}\leq b_{0}(\bm{x}^{(d)})+C_{\mathscr{C}} + 2\bm{\theta}''\cdot\bm{\varphi}(\bm{x}^{(d)}) \right\}\nonumber\\
				&=\mathcal{T}_{\bm{\theta}''}\quad\text{for all}\quad \bm{\theta}''\in \Theta.\label{Gththpsub0}
			\end{align}
			Moreover,   \eqref{eqoutQj}  implies
			\[
				b_{\bm{\theta}}(\bm{z})=b_{\bm{\theta}'}(\bm{z}) \quad \text{for all}\quad \bm{z}\in [0,1]^{d-1}\setminus\suppf{{\varphi}_{j^{*}}},
			\]
			and therefore 
			\begin{equation}
				G_{\bm{\theta},\bm{\theta}'}\subseteq \mathcal{S}_{j^{*}}=\suppf{{\varphi}_{j^{*}}}\times[0,1].\label{Gththpsub1}
			\end{equation}
			Then, by \eqref{Gththpsub0} and \eqref{Gththpsub1}, we get 
			\[
				G_{\bm{\theta},\bm{\theta}'}\subseteq \mathcal{T}_{\bm{\theta}''}\cap\mathcal{S}_{j^{*}} \quad\text{for all}\quad \bm{\theta}''\in \Theta,
			\]
			and with the notation in \eqref{lastlemH2}, \eqref{lastlemH5}, we obtain
			\begin{align}
				\int_{\left\{\bm{x}\in \mathcal{X}\mid h_{\bm{\theta}}(\bm{x})\neq h_{\bm{\theta}'}(\bm{x})\right\}}  (f_{\bm{\theta}}   + f_{\bm{\theta}'})   d\lambda&= \int_{G_{\bm{\theta},\bm{\theta}'}}  f_{\bm{\theta}}    d\lambda +\int_{G_{\bm{\theta},\bm{\theta}'}}  f_{\bm{\theta}'}   d\lambda\nonumber\\
				&\leq  \int_{\mathcal{T}_{\bm{\theta}}\cap\mathcal{S}_{j^{*}}}  f_{\bm{\theta}}    d\lambda +\int_{\mathcal{T}_{\bm{\theta}'}\cap\mathcal{S}_{j^{*}}}  f_{\bm{\theta}'}   d\lambda\nonumber\\
				&=  \int_{\mathcal{U}_{\bm{\theta}}^{(1)}}  f_{\bm{\theta}}    d\lambda +\int_{\mathcal{U}_{\bm{\theta}'}^{(1)}}  f_{\bm{\theta}'}   d\lambda.\label{hthtpneq}
			\end{align} 
			In conclusion, from  \eqref{U1fthtalb}, \eqref{U1fthtaprilb} and  \eqref{hthtpneq}, we arrive at 
			\begin{align*}
				\int_{\left\{\bm{x}\in \mathcal{X}\mid h_{\bm{\theta}}(\bm{x})\neq h_{\bm{\theta}'}(\bm{x})\right\}}  (f_{\bm{\theta}}   + f_{\bm{\theta}'})   d\lambda&\leq C_{\mathscr{C}}^{\widetilde{\gamma}}M^{-(d-1)}\left(\frac{1}{\widetilde{\gamma}} \norm{1+ \varphi}_{L^{\widetilde{\gamma}}([0,1]^{d-1})}^{\widetilde{\gamma}}+\frac{1}{\widetilde{\gamma}}\right)\\&\leq 2^{\widetilde{\gamma}+1}C_{\mathscr{C}}^{\widetilde{\gamma}}M^{-(d-1)}.
			\end{align*} 
\end{proof}

In what follows, we use the Hellinger distance to compare different probability measures. For densities $f$ and $g$ with respect to a measure $ \nu $, the Hellinger distance is defined by
\begin{equation}
	\rho_{H,\nu}(f,g)=\left(\int_{\mathcal{X}} \left(\sqrt{f}-\sqrt{g} \right)^{2}d\nu\right)^{1/2}.\label{defhellin}
\end{equation}

The next lemma provides an upper bound on the Hellinger distance between product measures corresponding to parameters at Hamming distance one.

\begin{lem}\label{lemauxtoass}
	Let $ \mathcal{D} $ be defined as 
	\[
	\mathcal{D}:=\left\{\bm{f_{\bm{\theta}}}^{\otimes n} \mid  \bm{\theta}\in \Theta \right\},\quad \text{where}\quad \bm{f_{\bm{\theta}}}(\bm{x},y):=\frac{f_{\bm{\theta}}(\bm{x})\mathbbm{1}_{\left\{y=h_{\bm{\theta}}(\bm{x})\right\}}}{\eta(\{y\})},
	\]
	and $ f_{\bm{\theta}} $ is as in \eqref{defdenst}. Then, $ \mathcal{D} $ is a family of densities on $ \Lambda^{n} $ with respect to $\bm{\lambda}^{\otimes n}$ (see \eqref{bmlambda}), and  for all $ \bm{\theta},\bm{\theta}'\in \Theta $ with $ \rho_{Ham}(\bm{\theta},\bm{\theta}')=1 $, we obtain  
	\begin{equation}
		\rho_{H,\bm{\lambda}^{\otimes n}}^{2}(\bm{f_{\bm{\theta}}}^{\otimes n},\bm{f_{\bm{\theta}'}}^{\otimes n})  \leq 2^{\widetilde{\gamma}+4}nC_{\mathscr{C}}^{\widetilde{\gamma}}M^{-(d-1)}.\label{prodHup}
	\end{equation}
\end{lem}
 \begin{remark}\label{remarkdenbmf}
 	In Lemma \ref{lemauxtoass}, we denote by $ \bm{\mu_{\bm{\theta}}} $ the probability measure on $\Lambda$  having density  $ \bm{f_{\bm{\theta}}} $ with respect to $\bm{\lambda}$. Accordingly, $ \bm{\mu_{\bm{\theta}}} $ has marginal $\mu_{\bm{\theta}}$ on $\mathcal{X}$, and $ \bm{f_{\bm{\theta}}}^{\otimes n} $ is the density of the probability measure $\bm{\mu_{\bm{\theta}}}^{\otimes n} $ on $\Lambda^{n}$ with respect to $\bm{\lambda}^{\otimes n}$.
 	 
 \end{remark}
 \begin{proof} 	
 	Note that $ \bm{f_{\bm{\theta}}} $ is indeed a density function, since 
 	\begin{align*}
 		\int_{\Lambda} \bm{f_{\bm{\theta}}} d\bm{\lambda} &= \int_{\mathcal{X}\times \{0,1\}} \frac{f_{\bm{\theta}}(\bm{x})\mathbbm{1}_{\left\{y=h_{\bm{\theta}}(\bm{x})\right\}}}{\eta(\{y\})} d\lambda(\bm{x})d\eta(y)\\
 		&=\int_{\mathcal{X}\times\{0\}} \frac{f_{\bm{\theta}}(\bm{x})\mathbbm{1}_{\left\{y=h_{\bm{\theta}}(\bm{x})\right\}}}{\eta(\{y\})} d\lambda(\bm{x})d\eta(y)+\int_{\mathcal{X}\times\{1\}} \frac{f_{\bm{\theta}}(\bm{x})\mathbbm{1}_{\left\{y=h_{\bm{\theta}}(\bm{x})\right\}}}{\eta(\{y\})} d\lambda(\bm{x})d\eta(y)\\
 		&=\int_{\mathcal{X}} f_{\bm{\theta}}(\bm{x})\mathbbm{1}_{\left\{h_{\bm{\theta}}(\bm{x})=0\right\}}  d\bm{x}+  \int_{\mathcal{X}} f_{\bm{\theta}}(\bm{x})\mathbbm{1}_{\left\{h_{\bm{\theta}}(\bm{x})=1\right\}}  d\bm{x} \\
 		&=\int_{\mathcal{X}} f_{\bm{\theta}}  d\lambda\\
 		&=1.
 	\end{align*}
 	Then, $ \mathcal{D} $ is a family of densities on $ \Lambda^{n} $ with respect to $\bm{\lambda}^{\otimes n}$. 
 	In addition (see \cite[Section 2.4]{tsybankovnonpa}),
 	\begin{equation}
 		\rho_{H,\bm{\lambda}^{\otimes n}}^{2}(\bm{f_{\bm{\theta}}}^{\otimes n},\bm{f_{\bm{\theta}'}}^{\otimes n})=2\left(1-\left(1-\frac{\rho_{H,\bm{\lambda}}^{2}(\bm{f_{\bm{\theta}}},\bm{f_{\bm{\theta}'}})}{2}\right)^{n}\right)\leq n\rho_{H,\bm{\lambda}}^{2}(\bm{f_{\bm{\theta}}},\bm{f_{\bm{\theta}'}}), \label{hellot}
 	\end{equation}
 	and 
 	\begin{align*}
 		\rho_{H,\bm{\lambda}}^{2}(\bm{f_{\bm{\theta}}},\bm{f_{\bm{\theta}'}})&= \int_{\Lambda} (\sqrt{\bm{f_{\bm{\theta}}}}-\sqrt{\bm{f_{\bm{\theta}'}}})^{2} d\bm{\lambda}\\
 		&\leq \int_{\Lambda} \left|\bm{f_{\bm{\theta}}}-\bm{f_{\bm{\theta}'}}\right| d\bm{\lambda}\\
 		&=\int_{\mathcal{X}}\int_{\{0,1\}} \left|\frac{f_{\bm{\theta}}(\bm{x})\mathbbm{1}_{\left\{y=h_{\bm{\theta}}(\bm{x})\right\}}}{\eta(\{y\})}-\frac{f_{\bm{\theta}'}(\bm{x})\mathbbm{1}_{\left\{y=h_{\bm{\theta}'}(\bm{x})\right\}}}{\eta(\{y\})}\right|d\eta(y) d\lambda(\bm{x})\\
 		&=\sum_{j\in \{0,1\}}\int_{\mathcal{X}}  \left|f_{\bm{\theta}}(\bm{x})\mathbbm{1}_{\left\{h_{\bm{\theta}}(\bm{x})=j\right\}}-f_{\bm{\theta}'}(\bm{x})\mathbbm{1}_{\left\{h_{\bm{\theta}'}(\bm{x})=j\right\}}\right|  d\bm{x}\\
 		&=\sum_{j\in \{0,1\}}\int_{\left\{\bm{x}\in \mathcal{X}\mid h_{\bm{\theta}}(\bm{x})=h_{\bm{\theta}'}(\bm{x})\right\}} \left|f_{\bm{\theta}}(\bm{x})-f_{\bm{\theta}'}(\bm{x}) \right| \mathbbm{1}_{\left\{h_{\bm{\theta}}(\bm{x})=j\right\}} d\bm{x}\\
 		&\quad+\sum_{j\in \{0,1\}}\int_{\left\{\bm{x}\in \mathcal{X}\mid h_{\bm{\theta}}(\bm{x})\neq h_{\bm{\theta}'}(\bm{x})\right\}} \left|f_{\bm{\theta}}(\bm{x})\mathbbm{1}_{\left\{h_{\bm{\theta}}(\bm{x})=j\right\}}-f_{\bm{\theta}'}(\bm{x})\mathbbm{1}_{\left\{h_{\bm{\theta}'}(\bm{x})=j\right\}}\right|  d\bm{x}\\
 		&= \int_{\left\{\bm{x}\in \mathcal{X}\mid h_{\bm{\theta}}(\bm{x})=h_{\bm{\theta}'}(\bm{x})\right\}} \left|f_{\bm{\theta}} -f_{\bm{\theta}'}  \right|  d\lambda +\int_{\left\{\bm{x}\in \mathcal{X}\mid h_{\bm{\theta}}(\bm{x})\neq h_{\bm{\theta}'}(\bm{x})\right\}}  (f_{\bm{\theta}}   + f_{\bm{\theta}'})   d\lambda.
 	\end{align*}
 	Therefore, by item \ref{itemdisho5} of Lemma \ref{premainlemma} together with the above inequality, it follows that
 	\begin{align}
 		\rho_{H,\bm{\lambda}}^{2}(\bm{f_{\bm{\theta}}},\bm{f_{\bm{\theta}'}})&\leq \int_{\mathcal{X}} \left|f_{\bm{\theta}} -f_{\bm{\theta}'}  \right|  d\lambda +\int_{\left\{\bm{x}\in \mathcal{X}\mid h_{\bm{\theta}}(\bm{x})\neq h_{\bm{\theta}'}(\bm{x})\right\}}  (f_{\bm{\theta}}   + f_{\bm{\theta}'}) d\lambda\nonumber\\
 		&\leq 2^{\widetilde{\gamma}+3}C_{\mathscr{C}}^{\widetilde{\gamma}}M^{-(d-1)}+2^{\widetilde{\gamma}+1}C_{\mathscr{C}}^{\widetilde{\gamma}}M^{-(d-1)}\nonumber\\
 		&\leq 2^{\widetilde{\gamma}+4}C_{\mathscr{C}}^{\widetilde{\gamma}}M^{-(d-1)}.\label{hellot2}
 	\end{align}
 	Finally, \eqref{hellot} and \eqref{hellot2} imply 
 	\[
 		\rho_{H,\bm{\lambda}^{\otimes n}}^{2}(\bm{f_{\bm{\theta}}}^{\otimes n},\bm{f_{\bm{\theta}'}}^{\otimes n}) \leq n\rho_{H,\bm{\lambda}}^{2}(\bm{f_{\bm{\theta}}},\bm{f_{\bm{\theta}'}})\leq 2^{\widetilde{\gamma}+4}nC_{\mathscr{C}}^{\widetilde{\gamma}}M^{-(d-1)}.
 	\]

 \end{proof}
\section{Proof of Theorem \ref{maintheo}}\label{mainteopro}

\begin{figure}[htbp]
	\begin{subfigure}{0.379\linewidth}
		\centering
		\includegraphics[width=\linewidth]{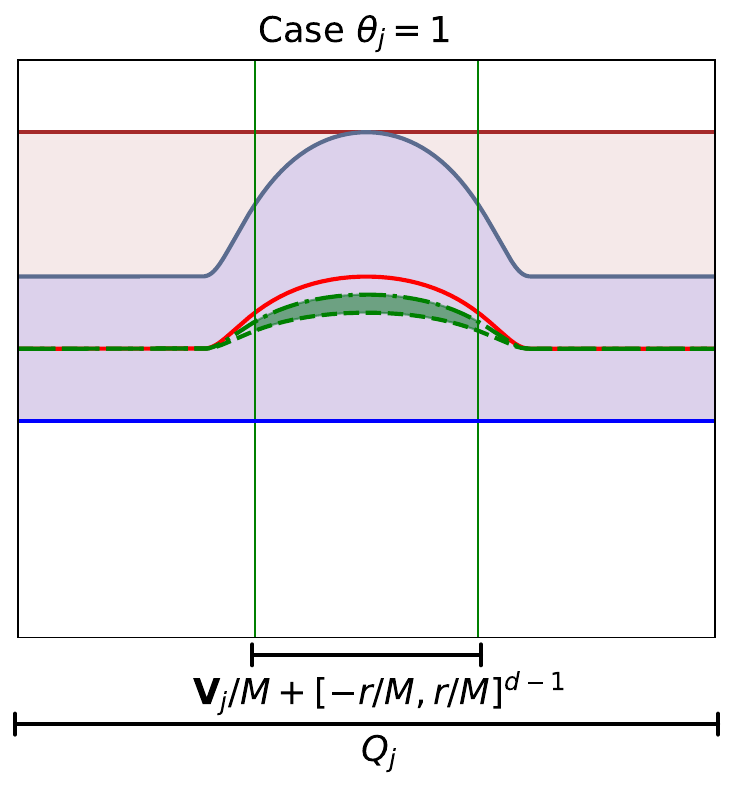}
	\end{subfigure}	
	\begin{subfigure}{0.62\linewidth}
		\centering
		\includegraphics[width=\linewidth]{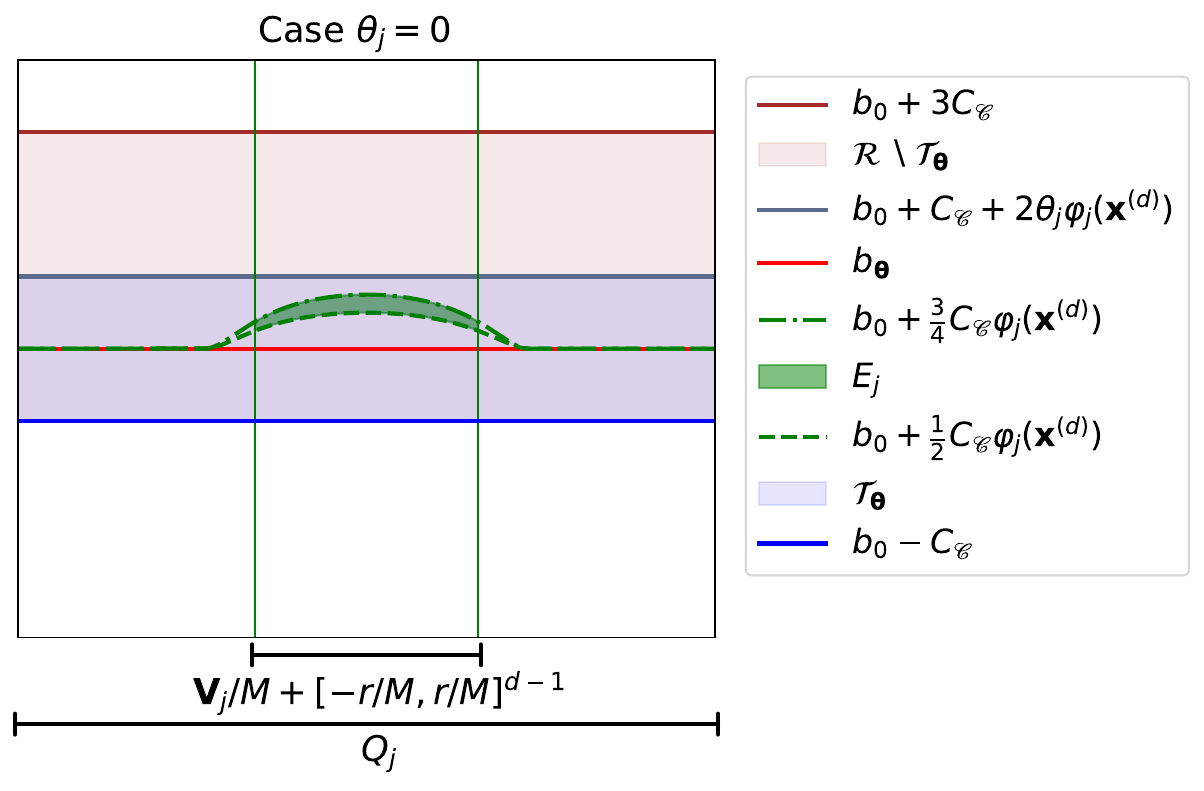}
	\end{subfigure}	
	\caption{Illustration of $E_j$ for $ d=2 $, where $\varphi$ is a bump function and  $b_0 := 1/2$.	
		In the image on the left, the bump is active and the function $ b_{\bm{\theta}} $ lies above the region $ E_j $, which implies by \eqref{bcont} that $ h_{\bm{\theta}}(\bm{x})=0 $  for all $ \bm{x}\in E_{j} $. In contrast, in the image on the right, the bump is inactive and the function $ b_{\bm{\theta}} $ lies below  $E_{j}$; therefore $ h_{\bm{\theta}}(\bm{x})=1 $, for all $ \bm{x}\in E_{j} $. In both cases the region $E_j$ is sufficiently far from the decision boundary, and therefore $f_{\bm{\theta}}$ can be bounded from below on $E_j$, as in \eqref{lbforfinEj}. Consequently, $E_j$ is a suitable region for projecting the estimator $A(\bm{S}_n)$ onto the discrete set $\Theta$, as in~\eqref{projonT}. Moreover, by taking different parameters $ d $, $ b_{0} $, and $ \varphi $, the same conclusions about $E_j$ still hold. For visualization purposes, we use a particular choice of parameters. The same idea applies for arbitrary choices of $d$, $b_0$, and $\varphi$ satisfying the hypotheses of Theorem~\ref{maintheo}.}
		\label{Ej}		
\end{figure}

We know by \eqref{varphihold0} that $ \varphi $ is Hölder continuous at $ \bm{0} $ with exponent $\alpha\in (0,1] $ and constant $ C_{\varphi} $, then there exists $ r\in(0,1]  $ such that 
\begin{equation}
	\varphi(\bm{z})\geq 1/2, \quad\text{for all}\quad 	\bm{z}\in \mathbb{R}^{d-1} \quad\text{satisfying}\quad \norm{\bm{z}}_{\infty}\leq r.\label{defr}
\end{equation}
Indeed, we fix
\[
r:= \min\{1,(2C_{\varphi})^{-1/\alpha}\}
\]
and using \eqref{varphihold0}, \eqref{suppphijdef}, we see that 
\[
	1-\varphi(\bm{z})=|\varphi(\bm{z})-\varphi(\bm{0})|\leq C_{\varphi}\norm{\bm{z}}_{\infty}^{\alpha}\quad\text{for all}\quad \bm{z} \in \mathbb{R}^{d-1},
\]
which implies 
\[
	 \varphi(\bm{z}) \geq  1- C_{\varphi}\norm{\bm{z}}_{\infty}^{\alpha}\geq1-\min\{C_{\varphi},1/2\} \geq   1/2 \quad\text{for all}\quad 	\bm{z}\in \mathbb{R}^{d-1} \quad\text{satisfying}\quad \norm{\bm{z}}_{\infty}\leq r,
\]
and \eqref{defr} holds. Thus, by \eqref{eqvarphij} and \eqref{defr}, we obtain 
\begin{equation}
	\varphi_j(\bm{z})\geq 1/2, \quad \text{for all}\quad \bm{z}\in \mathbb{R}^{d-1} \quad\text{satisfying}\quad  \norm{\bm{z}-\bm{v}_{j}/M}_{\infty}\leq r/M,\label{varphigq}
\end{equation}
and for each  $j\in \{1,\ldots,m\}$. We define the region (see Figure~\ref{Ej})
\begin{equation}
	E_{j}=\left\{\bm{x}\in \mathcal{X}\mid  \bm{x}^{(d)} \in N_{j}  \quad\text{and}\quad C_{\mathscr{C}}\varphi_{j}(\bm{x}^{(d)})/2  \leq x_d-b_{0}(\bm{x}^{(d)})\leq   3C_{\mathscr{C}}\varphi_{j}(\bm{x}^{(d)})/4 \right\}\label{defEj}
\end{equation}
where (see \eqref{defQj})
\begin{equation}
	N_{j}=\left\{\bm{z}\in [0,1]^{d-1}\mid \norm{\bm{z}-\bm{v}_{j}/M}_{\infty}\leq r/M  \right\}\subseteq Q_{j}\in P,\label{defEj2}
\end{equation} 
and \eqref{varphigq}, \eqref{defEj} and \eqref{defEj2} imply
\begin{align}
	\lambda(E_{j})&=\int_{N_{j}} \int_{C_{\mathscr{C}}\varphi_{j}(\bm{x}^{(d)})/2  \leq x_d-b_{0}(\bm{x}^{(d)})\leq   3C_{\mathscr{C}}\varphi_{j}(\bm{x}^{(d)})/4 }  1 dx_d d\bm{x}^{(d)}\nonumber\\
	&=\int_{N_{j}} C_{\mathscr{C}}\varphi_{j}(\bm{x}^{(d)})/4 d\bm{x}^{(d)}\nonumber\\
	&\geq \frac{1}{8}C_{\mathscr{C}} \int_{\norm{\bm{z}-\bm{v}_{j}/M}_{\infty}\leq r/M}1d\bm{z}\nonumber\\
	&=\frac{1}{8}(2r)^{d-1}C_{\mathscr{C}}M^{-(d-1)}. \label{measEj}
\end{align}  
For all $ \bm{\theta}\in \Theta $, from \eqref{deftubethet}, \eqref{RcapTae} and \eqref{defEj}, we see that 
\begin{equation}
	E_{j}\subset\mathcal{T}_{\bm{\theta}}=\mathcal{R}\cap\mathcal{T}_{\bm{\theta}} \quad\text{and}\quad  
	h_{\bm{\theta}}(\bm{x})=1-\theta_{j} 
	\quad\text{for all}\quad \bm{x}\in E_{j}.\label{htinEj}
\end{equation}
Using \eqref{aesuppsemp}, \eqref{defEj}, \eqref{defEj2} and \eqref{htinEj}, we get
\begin{align*}
	E_{j}&=\left\{\bm{x}\in \mathcal{X}\mid  \bm{x}^{(d)} \in N_{j}  \quad\text{and}\quad (1/2-\theta_{j})C_{\mathscr{C}}\varphi_{j}(\bm{x}^{(d)}) \leq x_d-b_{\bm{\theta}}(\bm{x}^{(d)})\leq   (3/4-\theta_{j})C_{\mathscr{C}}\varphi_{j}(\bm{x}^{(d)}) \right\}\\
	&\subseteq L_{E_{j}}:= \left\{\bm{x}\in \mathcal{X}\mid  \bm{x}^{(d)} \in N_{j}  \quad\text{and}\quad |x_d-b_{\bm{\theta}}(\bm{x}^{(d)})|\geq C_{\mathscr{C}}\varphi_{j}(\bm{x}^{(d)})/4 \right\},
\end{align*}
and 
\begin{equation}
	f_{\bm{\theta}}\mathbbm{1}_{E_{j}}(\bm{x})=f_{\bm{\theta}}\mathbbm{1}_{E_{j}\cap L_{E_{j}}}(\bm{x})=\frac{1}{2}|x_d-b_{\bm{\theta}}(\bm{x}^{(d)})|^{\widetilde{\gamma}-1}\geq \frac{1}{2}(C_{\mathscr{C}}\varphi_{j}(\bm{x}^{(d)})/4)^{\widetilde{\gamma}-1}.\label{lbforfinEj}
\end{equation}

By definition \eqref{defestim}, for all $ A\in\mathcal{A}_{n}(L^{2}(\lambda)) $ and for all $ \bm{S}_{n}\in \Lambda^{n} $, we have  $ A(\bm{S}_{n})\in L^{2}(\lambda) $. Since $ f_{\bm{\theta}} $ in \eqref{defdenst} is upper bounded, then $ A(\bm{S}_{n})\in L^{2}(\mu_{\bm{\theta}}) $. Therefore, by \eqref{varphigq}, \eqref{defEj}, \eqref{defEj2}, \eqref{htinEj} and \eqref{lbforfinEj}, we obtain
\begin{align}
	\norm{A(\bm{S}_{n})-h_{\bm{\theta}}}_{L^{2}(\mu_{\bm{\theta}})}^{2}&=\int_{\mathcal{X}} |A(\bm{S}_{n})-h_{\bm{\theta}}|^{2} f_{\bm{\theta}} d\lambda\nonumber\\
	&\geq \sum_{j=1}^{m} \int_{E_{j}} |A(\bm{S}_{n})-h_{\bm{\theta}}|^{2} f_{\bm{\theta}} d\lambda\nonumber\\
	&\geq \frac{1}{2}(C_{\mathscr{C}}/4)^{\widetilde{\gamma}-1} \sum_{j=1}^{m} \int_{E_{j}} |A(\bm{S}_{n})(\bm{x})-t(\theta_{j})|^{2} (\varphi_{j}(\bm{x}^{(d)}))^{\widetilde{\gamma}-1} d\bm{x}\nonumber\\
	&\geq \frac{1}{2}(C_{\mathscr{C}}/8)^{\widetilde{\gamma}-1} \sum_{j=1}^{m} \int_{E_{j}} |A(\bm{S}_{n})-t(\theta_{j})|^{2} d\lambda\quad\text{where}\quad  t(\theta_{j}):=1-\theta_j.\label{suptomax0} 
\end{align}

For each $ A\in\mathcal{A}_{n}(L^{2}(\lambda)) $, we define the projection $ \widehat{\bm{\theta}}_{A}: \Lambda^{n} \to \Theta $ of $ A(\bm{S}_{n}) $ onto $ \Theta $, such that 
\begin{equation}
	\widehat{\bm{\theta}}_{A}(\bm{S}_{n}):=(\widehat{\theta_{1}}_{A}(\bm{S}_{n}),\ldots,\widehat{\theta_{m}}_{A}(\bm{S}_{n}))\quad\text{where}\quad\widehat{\theta_{j}}_{A}(\bm{S}_{n})\in \underset{\theta\in \{0,1\}}{\operatorname{arg\,min}} \norm{A(\bm{S}_{n})-t(\theta)}_{L^{2}(\lambda\mid_{E_{j}})}.\label{projonT}
\end{equation}
Here we used the notation $ \norm{\cdot}_{L^{2}(\lambda\mid_{E_{j}})}^{2}:=\int_{E_{j}} |\cdot|^2 d\lambda $  and  $ t(\theta):=1-\theta $. If both values $\theta=0$ and $\theta=1$ attain the minimum in~\eqref{projonT}, we fix $ \widehat{\theta_{j}}_{A}(\bm{S}_{n}):=0$. Then,
\begin{align}
	|\widehat{\theta_{j}}_{A}(\bm{S}_{n}) - \theta_{j}|\sqrt{\lambda(E_{j})}
	&=\norm{\widehat{\theta_{j}}_{A}(\bm{S}_{n}) - \theta_{j}}_{L^{2}(\lambda\mid_{E_{j}})}\nonumber\\
	&=\norm{t(\widehat{\theta_{j}}_{A}(\bm{S}_{n}))-t(\theta_{j})}_{L^{2}(\lambda\mid_{E_{j}})}\nonumber\\
	&\leq \norm{A(\bm{S}_{n})-t(\widehat{\theta_{j}}_{A}(\bm{S}_{n}))}_{L^{2}(\lambda\mid_{E_{j}})}+\norm{A(\bm{S}_{n})-t(\theta_{j})}_{L^{2}(\lambda\mid_{E_{j}})}\nonumber\\
	&\leq 2\norm{A(\bm{S}_{n})-t(\theta_{j})}_{L^{2}(\lambda\mid_{E_{j}})}.\label{argm2ub}
\end{align}
So, we use \eqref{suptomax0} and \eqref{argm2ub}, to obtain
\begin{align*}
	\norm{A(\bm{S}_{n})-h_{\bm{\theta}}}_{L^{2}(\mu_{\bm{\theta}})}^{2}&\geq\frac{1}{2}(C_{\mathscr{C}}/8)^{\widetilde{\gamma}-1} \sum_{j=1}^{m}  \norm{A(\bm{S}_{n})-t(\theta_{j})}_{L^{2}(\lambda\mid_{E_{j}})}^{2}\\
	&\geq\frac{1}{8}(C_{\mathscr{C}}/8)^{\widetilde{\gamma}-1} \sum_{j=1}^{m} |\widehat{\theta_{j}}_{A}(\bm{S}_{n}) - \theta_{j}|\lambda(E_{j}),
\end{align*}
and by \eqref{indexsetham}, \eqref{measEj}, it follows that
\begin{align*}
	\norm{A(\bm{S}_{n})-h_{\bm{\theta}}}_{L^{2}(\mu_{\bm{\theta}})}^{2}&\geq \frac{1}{8}(C_{\mathscr{C}}/8)^{\widetilde{\gamma}-1} \sum_{j\in J\left(\widehat{\bm{\theta}}_{A}(\bm{S}_{n}),\bm{\theta}\right)} \lambda(E_{j})\\
	&\geq \frac{(2r)^{d-1}}{8^{\widetilde{\gamma}+1}} C_{\mathscr{C}}^{\widetilde{\gamma}}M^{-(d-1)} \sum_{j\in J\left(\widehat{\bm{\theta}}_{A}(\bm{S}_{n}),\bm{\theta}\right)} 1\\
	&=\left(\frac{(2r)^{d-1}}{8^{\widetilde{\gamma}+1}}\right) C_{\mathscr{C}}^{\widetilde{\gamma}}M^{-(d-1)}\rho_{Ham}(\widehat{\bm{\theta}}_{A}(\bm{S}_{n}),\bm{\theta}).
\end{align*}
Taking the expectation $ \mathbb{E}_{\bm{\theta}}:=\mathbb{E}_{\{\bm{x}_{i}\}_{i=1}^{n} \widesim{iid} \mu_{\bm{\theta}}} $ on both sides of the last inequality, we arrive at
\begin{equation}
	\mathbb{E}_{ \bm{\theta}}\norm{A(\bm{S}_{n})-h_{\bm{\theta}}}_{L^{2}(\mu_{\bm{\theta}})}^{2}\geq\left(\frac{(2r)^{d-1}}{8^{\widetilde{\gamma}+1}}\right) C_{\mathscr{C}}^{\widetilde{\gamma}}M^{-(d-1)}\mathbb{E}_{ \bm{\theta}}\rho_{Ham}(\widehat{\bm{\theta}}_{A}(\bm{S}_{n}),\bm{\theta}).\label{suptomax1}
\end{equation}
In conclusion, \eqref{suptomax1} together with the notation in \eqref{eqinfsup}, imply
\begin{align}
	\sup_{(h,\mu)\in \mathcal{P}_{\mathscr{C}}(\mathcal{L}_{\lambda})} \mathbb{E}_{\{\bm{x}_{i}\}_{i=1}^{n} \widesim{iid} \mu}\norm{A(\bm{S}_{n})-h}_{L^{2}(\mu)}^{2}&\geq \max_{(h,\mu)\in \mathcal{P}_{\mathscr{C}_{\Theta}}(\mathcal{M}_{\Theta})} \mathbb{E}_{\{\bm{x}_{i}\}_{i=1}^{n} \widesim{iid} \mu}\norm{A(\bm{S}_{n})-h}_{L^{2}(\mu)}^{2}\label{eqimpsup00}\\
	&\geq \max_{\bm{\theta}\in \Theta} \mathbb{E}_{ \bm{\theta}}\norm{A(\bm{S}_{n})-h_{\bm{\theta}}}_{L^{2}(\mu_{\bm{\theta}})}^{2}\nonumber\\
	&\geq \left(\frac{(2r)^{d-1}}{8^{\widetilde{\gamma}+1}}\right) C_{\mathscr{C}}^{\widetilde{\gamma}}M^{-(d-1)}\max_{\bm{\theta}\in \Theta}\mathbb{E}_{ \bm{\theta}}\rho_{Ham}(\widehat{\bm{\theta}}_{A}(\bm{S}_{n}),\bm{\theta}),\label{eqimpsup}
\end{align}
since $\mathscr{C}_{\Theta} \subseteq \mathscr{C}$ (by \eqref{eqmaint0}), and
\[
\mathcal{P}_{\mathscr{C}}(\mathcal{L}_{\lambda})\supseteq\mathcal{P}_{\mathscr{C}_{\Theta}}(\mathcal{M}_{\Theta})\supseteq  \left\{(h,\mu)\in H_{\mathscr{C}_{\Theta}}\times\mathcal{M}_{\Theta}\mid (h,\mu)=(h_{\bm{\theta}},\mu_{\bm{\theta}})\right\}.
\]
 
We state Assouad’s lemma as in \cite[Theorem 2.12 (Hellinger version)]{tsybankovnonpa} to relate it to the lower bound \eqref{eqimpsup}.

\begin{lem}\label{AssouadLemma}  Let $ \mathcal{D}:=\left\{\bm{P}_{\bm{\theta}}\mid \bm{\theta}\in \Theta\right\} $ be a set of $ 2^{m} $ probability densities on $ \Lambda^{n} $ with respect to $\bm{\lambda}^{\otimes n}$, where $ \bm{\lambda} $ is defined in \eqref{bmlambda}. If 
	\[
		\rho_{H,\bm{\lambda}^{\otimes n}}^{2}(\bm{P}_{\bm{\theta}},\bm{P}_{\bm{\theta}'})\leq \vartheta<2\quad\text{for all}\quad \rho_{Ham}(\bm{\theta},\bm{\theta}')=1,
	\]
	where $ \rho_{H,\cdot}^{2} $ denotes the Hellinger distance defined in \eqref{defhellin}. Then
	\begin{equation}
		\inf_{\widehat{\bm{\theta}}: \Lambda^{n} \to \Theta}\max_{\bm{\theta}\in \Theta} \mathbb{E}_{\bm{\theta}} \rho_{Ham}(\widehat{\bm{\theta}}(\bm{S}_{n}),\bm{\theta})\geq \frac{m}{2}\left(1-\sqrt{\vartheta(1-\vartheta/4)}\right),\label{eqassouad0}
	\end{equation}
	where $ \mathbb{E}_{\bm{\theta}}:=\mathbb{E}_{\bm{S}_{n}\widesim{~~} \bm{\nu}_{\bm{P}_{\bm{\theta}}}} $ and $  \bm{\nu}_{\bm{P}_{\bm{\theta}}} $ is the probability measure induced by $ \bm{P}_{\bm{\theta}} $. 
	
\end{lem}
 
 To apply Lemma \ref{AssouadLemma}, we use the family of densities
 \[
 	\mathcal{D}:=\left\{\bm{f_{\bm{\theta}}}^{\otimes n} \mid  \bm{\theta}\in \Theta \right\}
 \] 
 defined in Lemma \ref{lemauxtoass}. By \eqref{eqmaint0} and \eqref{prodHup}, for all $ \bm{\theta},\bm{\theta}'\in \Theta $ with $ \rho_{Ham}(\bm{\theta},\bm{\theta}')=1 $, we have
 \[
 	\rho_{H,\bm{\lambda}^{\otimes n}}^{2}(\bm{f_{\bm{\theta}}}^{\otimes n},\bm{f_{\bm{\theta}'}}^{\otimes n})  \leq 2^{\widetilde{\gamma}+4}nC_{\mathscr{C}}^{\widetilde{\gamma}}M^{-(d-1)}\leq 1/4 <2.
 \] 
Therefore, the hypotheses of Lemma \ref{AssouadLemma} are satisfied with  $ \vartheta:=2^{\widetilde{\gamma}+4}nC_{\mathscr{C}}^{\widetilde{\gamma}}M^{-(d-1)} $ and $ m=(M/2)^{d-1} $ (see Construction \ref{mainconst}). In conclusion, since $ \vartheta\leq 1/4 $,  \eqref{eqassouad0} implies
\begin{align}
	\inf_{\widehat{\bm{\theta}}: \Lambda^{n} \to \Theta}\max_{\bm{\theta}\in \Theta} \mathbb{E}_{\bm{\theta}} \rho_{Ham}(\widehat{\bm{\theta}}(\bm{S}_{n}),\bm{\theta})&\geq 2^{-d} M^{d-1}\left(1-\sqrt{\vartheta(1-\vartheta/4)}\right)\nonumber\\
	&\geq 2^{-d-1} M^{d-1},\label{concluAss}
\end{align}
 where from the notation in \eqref{bmmutens} and Remark~\ref{remarkdenbmf}, we know that 
 \[
 	 \mathbb{E}_{\bm{\theta}}:=\mathbb{E}_{\bm{S}_{n}\widesim{~~} \bm{\mu_{\bm{\theta}}}^{\otimes n}}:=\mathbb{E}_{\{\bm{x}_{i}\}_{i=1}^{n} \widesim{iid}  \mu_{\bm{\theta}}}.
 \]  
 Since $ A $ in \eqref{eqimpsup} is an arbitrary element of $ \mathcal{A}_{n}(L^{2}(\lambda)) $, then \eqref{eqimpsup} combined with \eqref{concluAss} imply 
\begin{align*}
	\sup_{(h,\mu)\in \mathcal{P}_{\mathscr{C}}(\mathcal{L}_{\lambda})} \mathbb{E}_{\{\bm{x}_{i}\}_{i=1}^{n} \widesim{iid} \mu}\norm{A(\bm{S}_{n})-h}_{L^{2}(\mu)}^{2}&\geq \left(\frac{(2r)^{d-1}}{8^{\widetilde{\gamma}+1}}\right) C_{\mathscr{C}}^{\widetilde{\gamma}}M^{-(d-1)}\max_{\bm{\theta}\in \Theta}\mathbb{E}_{ \bm{\theta}}\rho_{Ham}(\widehat{\bm{\theta}}_{A}(\bm{S}_{n}),\bm{\theta})\nonumber\\
	&\geq \left(\frac{(2r)^{d-1}}{8^{\widetilde{\gamma}+1}}\right) C_{\mathscr{C}}^{\widetilde{\gamma}}M^{-(d-1)} \inf_{\widehat{\bm{\theta}}: \Lambda^{n} \to \Theta}\max_{\bm{\theta}\in \Theta} \mathbb{E}_{\bm{\theta}} \rho_{Ham}(\widehat{\bm{\theta}}(\bm{S}_{n}),\bm{\theta})\nonumber\\ 
	&\geq   \left(\frac{r^{d-1}}{8^{\widetilde{\gamma}+2}}\right) C_{\mathscr{C}}^{\widetilde{\gamma}}.
\end{align*}
Finally, by taking infimums in the previous inequality over all $ A\in\mathcal{A}_{n}(L^{2}(\lambda)) $, we conclude 
\[
\mathcal{I}_{n}(\mathscr{C})\geq \left(\frac{r^{d-1}}{8^{\widetilde{\gamma}+2}}\right) C_{\mathscr{C}}^{\widetilde{\gamma}},
\]
and \eqref{eqmaint1} is fulfilled. \qed 


\section{Proof of Corollary \ref{maincoro}}\label{seccoroproff}

\subsection{Hölder space}\label{subsecHolder}

Let $ \psi\in C_{c}^{\infty}(\mathbb{R};\mathbb{R})  $ be any $ 1 $-dimensional bump function, with $ \psi(x)=0 $ for all $ |x|\geq1 $,  $ 0\leq \psi(x)\leq 1  $ for all $ x\in \mathbb{R} $, and  $ \psi(0)=1 $. For Construction~\ref{mainconst}, we set $ \varphi:\mathbb{R}^{d-1}\to \mathbb{R} $ as 
\begin{equation}
	\varphi(\bm{z}):=\psi(2\norm{\bm{z}}_{\infty}).\label{vphiCorHol}
\end{equation}
By definition above, $ \varphi(\bm{0})=1$,  $0\leq \varphi(\bm{z})\leq 1$ for all $\bm{z}\in \mathbb{R}^{d-1}$ and
\begin{align}
		\suppf{\varphi}&=\left\{\bm{z}\in \mathbb{R}^{d-1}\mid 2\norm{\bm{z}}_{\infty}\in \suppf{\psi} \subset (-1,1)\right\}\nonumber\\
		&\subseteq\left\{\bm{z}\in \mathbb{R}^{d-1}\mid \norm{\bm{z}}_{\infty}\leq1/2  \right\}\nonumber\\
		&\subset (-1,1)^{d-1}, \label{vphiCorHol2}
\end{align}
therefore \eqref{suppphijdef} holds. By the mean value theorem, since $ \psi\in C_{c}^{\infty}(\mathbb{R};\mathbb{R}) $, we have 
\begin{align*}
	|\varphi(\bm{z})-\varphi(\bm{w})|&=|\psi(2\norm{\bm{z}}_{\infty})-\psi(2\norm{\bm{w}}_{\infty})|\\
	&\leq 2\norm{\psi'}_{\infty}|\norm{\bm{z}}_{\infty}-\norm{\bm{w}}_{\infty}|\\
	&\leq 2\norm{\psi'}_{\infty} \norm{\bm{z}-\bm{w}}_{\infty} \quad \text{for all}\quad \bm{z},\bm{w}\in \mathbb{R}^{d-1},
\end{align*}
where $ \norm{\psi'}_{\infty}=\max_{x\in \mathbb{R}}|\psi'(x)| $ and $ \psi' $ denotes the first derivative of $\psi$. Thus, for $ \alpha\in (0,1] $, and for all $ \bm{z},\bm{w}\in \mathbb{R}^{d-1} $, we see that
\begin{itemize}
	\item If $ \norm{\bm{z}-\bm{w}}_{\infty}\leq 1 $, 
	\begin{equation}
		|\varphi(\bm{z})-\varphi(\bm{w})| \leq 2\norm{\psi'}_{\infty} \norm{\bm{z}-\bm{w}}_{\infty}  \leq  2\norm{\psi'}_{\infty} \norm{\bm{z}-\bm{w}}_{\infty}^{\alpha}.\label{holpsiphi1}
	\end{equation}
	\item If $ \norm{\bm{z}-\bm{w}}_{\infty}> 1 $, since $ 0\leq \varphi(\bm{v})\leq 1 $ for all $\bm{v}\in \mathbb{R}^{d-1}   $, then 
	\begin{equation}
			|\varphi(\bm{z})-\varphi(\bm{w})|\leq 1\leq \norm{\bm{z}-\bm{w}}_{\infty}^{\alpha}.\label{holpsiphi2}
	\end{equation}
\end{itemize}
Therefore, \eqref{holpsiphi1} and \eqref{holpsiphi2} imply that 
\begin{align}
	|\varphi(\bm{z})-\varphi(\bm{w})| &\leq K_{\varphi} \norm{\bm{z}-\bm{w}}_{\infty}^{\alpha}\label{holpsiphi3}\\
	&\leq K_{\varphi} \norm{\bm{z}-\bm{w}}_{2}^{\alpha} \quad \text{for all}\quad \bm{z},\bm{w}\in \mathbb{R}^{d-1},\label{holpsiphi31}
\end{align}
where $ K_{\varphi}:=\max\{1,2\norm{\psi'}_{\infty}\} $. So, $ \varphi $ is a Hölder continuous function with exponent $\alpha\in (0,1] $ and constant $ K_{\varphi} $ with respect to the Euclidean norm. Moreover, \eqref{holpsiphi3} implies that $ \varphi $ is Hölder continuous at $ \bm{0} $ with exponent $\alpha $ and constant $ C_{\varphi}:=K_{\varphi} $ with respect to the $ \ell_{\infty} $-norm (see \eqref{varphihold0}).

Up to this point, all the properties required for $ \varphi $ in Construction~\ref{mainconst} have been verified. It remains to choose $ C_{\mathscr{C}} $ in \eqref{condtildeC} such that $ \mathscr{C}_{\Theta}\subseteq \mathscr{C} $, as required by Theorem~\ref{maintheo} in \eqref{eqmaint0}. The remaining conditions on $ C_{\mathscr{C}} $ will then be checked, while all other elements of the construction (the parameters $ M $, $ m $, the partition $ P $, etc) are taken exactly as in Construction~\ref{mainconst}.

In Section~\ref{somespaces}, we defined the class of H\"older continuous functions $ \mathcal{H}_\alpha $ as the class $ \mathscr{C} $ satisfying Condition~\ref{conticon}. Then, by item~\ref{itemdishol21} in Lemma~\ref{premainlemma}, we know that 
\[
	\mathscr{C}:=\mathcal{H}_\alpha\supseteq \mathscr{C}_{\Theta}.
\]
Furthermore, the only important assumption in the proof of item~\ref{itemdishol21} in Lemma~\ref{premainlemma}, apart from the hypotheses on  $ \varphi $ and the fact that $ \mathscr{C} $ satisfies Condition~\ref{conticon}, is that $ C_{\mathscr{C}}\leq M^{-\alpha}/4 $ in \eqref{Rinside01}. Hence, by choosing
\begin{equation}
	C_{\mathscr{C}}:=M^{-\alpha}/4,\label{choCcH}
\end{equation}
we conclude that $ \mathscr{C}_{\Theta}\subseteq \mathscr{C} $.

	Next, we still need to ensure that the baseline function $ b_{0}\in \mathscr{C} $  satisfies
	\[
		b_{0}(\bm{z})\in [C_{\mathscr{C}},1-3C_{\mathscr{C}}] \quad\text{for all}\quad \bm{z}\in[0,1]^{d-1}
	\]
	in \eqref{Rinside01}. This can be achieved either by assuming that $ b_{0}\in \mathcal{H}_\alpha $ satisfies this condition, or by fixing a specific choice, for instance $ b_{0}\equiv1/2 $ or $ b_{0} $ as in Figure~\ref{ExamElBCH}, which clearly belongs to $ \mathcal{H}_\alpha $ and satisfies the above condition.

Finally, by \eqref{choCcH} and in order to satisfy the right-hand side in \eqref{eqmaint0}, it suffices to choose $ M $  such that
\begin{align*}
	1&\geq 2^{\widetilde{\gamma}+6}nC_{\mathscr{C}}^{\widetilde{\gamma}}M^{-(d-1)}\\
	&= 2^{\widetilde{\gamma}+6}n(M^{-\alpha}/4)^{\widetilde{\gamma}}M^{-(d-1)}\\
	&=2^{-\widetilde{\gamma}+6}n M^{-(d-1)-\gamma}.
\end{align*}  
Therefore, we choose $ M $ as the smallest even integer such that 
\begin{equation}
	M\geq 2^{\frac{-\widetilde{\gamma}+6}{\gamma+(d-1)}}n^{\frac{1}{\gamma+(d-1)}}:=M_{*}, \quad\text{that is}\quad M:=2\lceil M_{*}/2 \rceil,	\label{choM}
\end{equation}
where $ \lceil . \rceil $ denotes the ceiling function for integers. 
In conclusion, we have shown that all the conditions of Theorem~\ref{maintheo} are satisfied. So, \eqref{eqmaint1}, \eqref{choCcH} and \eqref{choM} imply 
\begin{equation}
	\mathcal{I}_{n}(\mathscr{C})\geq \left(\frac{r^{d-1}}{8^{\widetilde{\gamma}+2}}\right) C_{\mathscr{C}}^{\widetilde{\gamma}}= 
	\left(\frac{r^{d-1}}{8^{\widetilde{\gamma}+2}}\right) (M^{-\alpha}/4)^{\widetilde{\gamma}} =\left(\frac{4^{-\widetilde{\gamma}}r^{d-1}}{8^{\widetilde{\gamma}+2}}\right) \left(2\lceil M_{*}/2 \rceil\right)^{-\gamma}  \quad \text{for all}\quad n\in \mathbb{N},\label{almHolCol}
\end{equation}
where $ r= \min\{1,(2C_{\varphi})^{-1/\alpha}\} $ and $ C_{\varphi}=K_{\varphi}=\max\{1,2\norm{\psi'}_{\infty}\}  $. Note that 
\[
	\lceil M_{*}/2 \rceil\leq  M_{*}/2+1\quad \text{and}\quad M_{*}\geq 2^{-\frac{\widetilde{\gamma}}{\gamma+(d-1)}}=\frac{1}{C_{*}-1/2},\quad\text{with}\quad C_{*}:=2^{\frac{\widetilde{\gamma}}{\gamma+(d-1)}}+1/2.
\]
Thus, $ \lceil M_{*}/2 \rceil\leq  M_{*}/2+1\leq C_{*}M_{*} $ and 
\begin{equation}
	\left(2\lceil M_{*}/2 \rceil\right)^{-\gamma}    \geq (2C_{*}M_{*})^{-\gamma}.\label{almHolCol2}
\end{equation}
In addition, since $ C_{\varphi}=\max\{1,2\norm{\psi'}_{\infty}\} \geq 1 $, we have
\begin{equation}
	r= \min\{1,(2C_{\varphi})^{-1/\alpha}\}=(2C_{\varphi})^{-1/\alpha}. \label{almHolCo3}
\end{equation}
Then, from \eqref{almHolCol}, \eqref{almHolCol2} and \eqref{almHolCo3}, it follows that
\begin{equation}		
\mathcal{I}_{n}(\mathscr{C})\geq\left(\frac{4^{-\widetilde{\gamma}}r^{d-1}}{8^{\widetilde{\gamma}+2}}\right) (2C_{*}M_{*})^{-\gamma}= C^{*} n^{-\frac{\gamma}{\gamma+(d-1)}}  \quad \text{for all}\quad n\in \mathbb{N},\label{almHolCo4}
\end{equation}
where
\begin{align}
	C^{*}&:=\left(\frac{4^{-\widetilde{\gamma}}\left((2C_{\varphi})^{-1/\alpha}\right)^{d-1}}{8^{\widetilde{\gamma}+2}}\right)\left(\left(2^{\frac{\widetilde{\gamma}}{\gamma+(d-1)}+1}+1\right)\left(2^{\frac{-\widetilde{\gamma}+6}{\gamma+(d-1)}} \right)\right)^{-\gamma}\nonumber\\
	&\geq C_{\varphi}^{-\frac{d-1}{\alpha}}\left(2^{-2\widetilde{\gamma}}\left(2^{-1/\alpha}\right)^{d-1}2^{-3\widetilde{\gamma}-6}\right) \left(\left(2+2^{\frac{-\widetilde{\gamma}}{\gamma+(d-1)}}\right)\left(2^{\frac{6}{\gamma+(d-1)}} \right)\right)^{-\gamma}  \nonumber\\
	&\geq C_{\varphi}^{-\frac{d-1}{\alpha}}\left(  2^{-\frac{d-1}{\alpha}} 2^{-5\widetilde{\gamma}-6}\right) \left(2^{-2\gamma }\right) 2^{-6\gamma}\nonumber\\
	&\geq  2^{-(\frac{d-1}{\alpha}+13\widetilde{\gamma}+6)}  C_{\varphi}^{-\frac{d-1}{\alpha}}.\label{almHolCo5}
\end{align}
Here, the dependence on $ C_{\varphi} $ can be removed by fixing a specific bump function $\psi$; for instance, the standard bump
\begin{equation}
	\psi(x)=\begin{cases}
		e^{1-\frac{1}{1-x^{2}}} &\text{ if }~|x|<1\\
		0&\text{ if }~|x|\geq 1.
	\end{cases}\label{partiHpsi}
\end{equation}
Indeed, 
\[
	|\psi'(x)|=\begin{cases}
		2|x|(1-x^{2})^{-2}\psi(x)&\text{ if }~|x|<1\\		
		0&\text{ if }~|x|\geq 1,
	\end{cases} 
\]
and
 \[
 	\psi(x)=e^{-\frac{1}{1-x^{2}}}e\leq (1-x^{2})^{2}e\quad\text{for all}\quad x\in (-1,1).
 \]
 Therefore,
 \[
 |\psi'(x)|\leq \begin{cases}
 	2|x|e&\text{ if }~|x|<1\\		
 	0&\text{ if }~|x|\geq 1,
 \end{cases} \leq 2e\quad\text{for all}\quad x\in \mathbb{R},\quad \text{which implies} \quad \norm{\psi'}_{\infty}\leq 2e.
 \]
 Thus,
 \begin{equation} 	
 	C_{\varphi}=\max\{1,2\norm{\psi'}_{\infty}\}\leq\max\{1,4e\}=4e<2^4.\label{almHolCo6}
 \end{equation}
 
 Finally, from \eqref{almHolCo4}, \eqref{almHolCo5} and \eqref{almHolCo6}, we arrive at 
 \[
 	\mathcal{I}_{n}(\mathscr{C})\geq  2^{-\left(\frac{5(d-1)}{\alpha}+13\widetilde{\gamma}+6\right)} n^{-\frac{\gamma}{\gamma+(d-1)}}  \quad \text{for all}\quad n\in \mathbb{N},
 \]
 and therefore \eqref{HolCor} holds.\qed
 
\subsection{Barron space}\label{subsecbarron}
  Let $ C>0 $ be a constant. By Remark~\ref{Rlipbarron}, every Barron function $b \in \mathscr{C}:=\mathcal{B}_C$ is Lipschitz on $[0,1]^{d-1}$. So, in Construction~\ref{mainconst}, the parameter corresponding to Condition \ref{conticon} is $ \alpha=1 $, which in turn implies that $ \widetilde{\gamma}=\gamma $. 
  
  Similarly to the Hölder case in the proof of item \ref{holditem}, we consider a baseline $ 1 $-dimensional bump function $ \psi\in C_{c}^{\infty}(\mathbb{R};\mathbb{R})  $ satisfying  $ 0\leq \psi(x)\leq 1  $ for all $ x\in \mathbb{R} $, $\suppf{\psi} \subset (-1,1) $, and  $ \psi(x)=1 $ for all $ x\in [-1/2,1/2] $. We define $ \varphi:\mathbb{R}^{d-1}\to \mathbb{R} $ as 
  \begin{equation}
  	\varphi(\bm{z})=\prod_{j=1}^{d-1}\psi(z_{j})\quad \text{for all}\quad \bm{z}=(z_{1},\ldots,z_{d-1})\in \mathbb{R}^{d-1}.\label{barrvarphi}
  \end{equation}
  Therefore, $0\leq \varphi(\bm{z})\leq 1$ for all $\bm{z}\in \mathbb{R}^{d-1}$, $ \varphi(\bm{z})=1 $ for all $ \bm{z}\in [-1/2,1/2]^{d-1} $, and
  \begin{align*}
  	\suppf{\varphi}&=\left\{\bm{z}\in \mathbb{R}^{d-1}\mid  z_{j}\in \suppf{\psi} ~\text{ for all }~j\in\{1,\ldots,d-1\}  \right\}\nonumber\\
  	&=(\suppf{\psi})^{d-1}\nonumber\\
  	&\subset (-1,1)^{d-1}.
  \end{align*}
  Thus, assumption \eqref{suppphijdef} is satisfied. 
  
  Since $ \psi\in C_{c}^{\infty}(\mathbb{R};\mathbb{R}) $, $ \suppf{\varphi}=(\suppf{\psi})^{d-1} $ and $ \psi(z_{j}) $ is the projection of $ \varphi(\bm{z}) $ on the $ j $-component, we have $ \varphi\in C_{c}^{\infty}(\mathbb{R}^{d-1};\mathbb{R}) $. Then, $ \norm{\nabla\varphi}_{2}< \infty $, and the mean value theorem implies 
  \[
  	|\varphi(\bm{z})-\varphi(\bm{w})|\leq K_{\varphi}\norm{\bm{z}-\bm{w}}_{2} \quad\text{for all}\quad \bm{z},\bm{w}\in [0,1]^{d-1},\quad\text{where}\quad K_{\varphi}:=\sup_{\bm{v}\in \mathbb{R}^{d-1}}\norm{\nabla\varphi(\bm{v})}_{2},
  \]
  i.e., $ \varphi $ is a Hölder (Lipschitz) continuous function  with exponent $\alpha =1 $ and constant $ K_{\varphi}>0 $ with respect to the Euclidean norm. Moreover, 
  \begin{itemize}
  	\item If $ \bm{z}\in [-1/2,1/2]^{d-1} $, we know that 
  	\[
  		|\varphi(\bm{z})-\varphi(\bm{0})|=|1-1|=0\leq C_{\varphi}\norm{\bm{z}}_{\infty}\quad\text{for all}\quad C_{\varphi}>0.
  	\]
  	\item If $ \bm{z}\in \mathbb{R}^{d-1}\setminus[-1/2,1/2]^{d-1} $, we have $ \norm{\bm{z}}_{\infty}>1/2 $ and
  	\[
  		|\varphi(\bm{z})-\varphi(\bm{0})|=|\varphi(\bm{z})-1|\leq 1<2\norm{\bm{z}}_{\infty}.  
  	\]
  	Here, we used that $0\leq \varphi(\bm{z})\leq 1$ for all $\bm{z}\in \mathbb{R}^{d-1}$.
  \end{itemize}
  Then, from the previous two items we get
  \begin{equation}
  	|\varphi(\bm{z})-\varphi(\bm{0})|\leq C_{\varphi}\norm{\bm{z}}_{\infty}\quad\text{with}\quad C_{\varphi}:=2, \quad\text{for all}\quad \bm{z} \in \mathbb{R}^{d-1},\label{Cvaphibarr}
  \end{equation}
  which implies \eqref{varphihold0} with $ \alpha=1 $. This completes the proof of the conditions on $ \varphi $ in Construction~\ref{mainconst}.
  
  In what follows, we justify the choice of $ C_{\mathscr{C}} $ so as to satisfy the hypotheses in \eqref{eqmaint0}. Let $ \bm{\theta}=(\theta_{1},\ldots,\theta_{m}) $ be an arbitrary element of $\Theta $. By \eqref{eqvarphij} and \eqref{condtildeC}, we have
  \begin{equation}
  	b_{\bm{\theta}}(\bm{z})=b_{0}(\bm{z})+C_{\mathscr{C}}\sum_{j=1}^{m}\theta_{j}\varphi\left(M(\bm{z}-\bm{v}_{j}/M)\right)\quad\text{for all}\quad \bm{z}\in [0,1]^{d-1}.\label{btin}
  \end{equation}
  By Barron definition \eqref{barroncondition0} and since $ b_{0}\in \mathscr{C} $, there exist $ C_{0}>0 $, $ c_{0}\in [-C_{0},C_{0}] $ and a measurable function $ F_{0}:\mathbb{R}^{d-1}\to \mathbb{C} $ satisfying
  \begin{equation}
  	b_{0}(\bm{z})=c_{0}+\int_{\mathbb{R}^{d-1}} (e^{i\bm{z}\cdot\bm{\xi}}-1) F_{0}(\bm{\xi})d\bm{\xi}\quad\text{and}\quad \int_{\mathbb{R}^{d-1}}\norm{\bm{\xi}}_{1}|F_{0}(\bm{\xi})|d\bm{\xi} \leq C_{0}\leq C.\label{btin1}
  \end{equation}
    Therefore, we choose $ b_{0} $ so that the corresponding constant $ C_{0} $ satisfies $ C_{0}\leq C/2 $ (for example, we may take $ b_{0}:= 1/2 $; or $ b_{0} $ can be chosen depending on $ C $ so that this condition holds).

  Let $ g:\mathbb{R}^{d-1}\to\mathbb{R} $ be defined by\footnote{Although $ b_{\bm{\theta}}:[0,1]^{d-1}\to[0,1] $, we define $ g $ on $ \mathbb{R}^{d-1} $ with value in $\mathbb{R}$, in order to use Fourier analysis on the whole space. This causes no ambiguity, since $ \varphi $ is already defined on $ \mathbb{R}^{d-1} $, satisfies $ \suppf{\varphi}\subset (-1,1)^{d-1} $, and $ 0\leq \varphi(\bm{z})\leq 1 $ for all $ \bm{z}\in \mathbb{R}^{d-1} $.}
  \begin{equation}
  	g(\bm{z}):=\sum_{j=1}^{m}\theta_{j}\varphi\left(M(\bm{z}-\bm{v}_{j}/M)\right).\label{btin2}
  \end{equation} 
  To prove that $ b_{\bm{\theta}}\in\mathscr{C} $, it suffices to show that  
  \begin{equation}
  	g(\bm{z})=c_{1}+\int_{\mathbb{R}^{d-1}} (e^{i\bm{z}\cdot\bm{\xi}}-1) F_{1}(\bm{\xi})d\bm{\xi}\quad\text{and}\quad \int_{\mathbb{R}^{d-1}}\norm{\bm{\xi}}_{1}|F_{1}(\bm{\xi})|d\bm{\xi} \leq C_{1}\leq C_{\mathscr{C}}^{-1}C/2,\label{btin3}
  \end{equation} 
  for some constant $ C_1>0 $, $ c_{1}\in [-C_{1},C_{1}] $, and some measurable function $ F_{1}:\mathbb{R}^{d-1}\to \mathbb{C} $.  Indeed, \eqref{btin}, \eqref{btin1}, \eqref{btin2} and \eqref{btin3}, imply that
  \begin{equation}
  	b_{\bm{\theta}}(\bm{z})=c+\int_{\mathbb{R}^{d-1}} (e^{i \bm{z}\cdot\bm{\xi}}-1) F(\bm{\xi}) d\bm{\xi}, \quad\text{with}\quad c:=c_{0}+C_{\mathscr{C}}c_{1}\quad\text{and}\quad F(\bm{\xi}):=F_{0}(\bm{\xi})+C_{\mathscr{C}}F_{1}(\bm{\xi}),\label{btin4}
  \end{equation} 
  where  
  \begin{align}
  	\int_{\mathbb{R}^{d-1}}\norm{\bm{\xi}}_{1}\left|F(\bm{\xi})\right|d\bm{\xi} &\leq \int_{\mathbb{R}^{d-1}}\norm{\bm{\xi}}_{1}\left|F_{0}(\bm{\xi})\right|d\bm{\xi}+\int_{\mathbb{R}^{d-1}}\norm{\bm{\xi}}_{1}\left|C_{\mathscr{C}}F_{1}(\bm{\xi})\right|d\bm{\xi}\nonumber\\
  	&\leq C_{0}+C_{\mathscr{C}}C_{1}\nonumber\\
  	&\leq C/2+C_{\mathscr{C}}C_{\mathscr{C}}^{-1}C/2\nonumber\\
  	&= C,\label{btin5}
  \end{align}
  showing that $ b_{\bm{\theta}}\in \mathscr{C} $.
  
  In order to prove \eqref{btin3}, we define
  \[
  \widetilde{F}(\bm{\xi}):=\mathscr{F}[g](\bm{\xi})=\int_{\mathbb{R}^{d-1}} g(\bm{z}) e^{-2\pi i \bm{z}\cdot\bm{\xi}} d\bm{z}\quad \text{and}\quad F_{1}(\bm{\xi}):=(2\pi)^{-(d-1)}\widetilde{F}(\bm{\xi}/2\pi),
  \]
  where $ \mathscr{F} $ denotes the Fourier transform (see \cite[Definition 2.2.8]{fourier}). Once the right-hand side of \eqref{btin3} is established, the identity on the left-hand side of \eqref{btin3} follows from the Fourier inversion formula (see \cite[Definition 2.2.13]{fourier}). Indeed, since $ \widetilde{F}=\mathscr{F}[g] $, the Fourier inversion formula yields
  \begin{align*}
  	g(\bm{z})&=\int_{\mathbb{R}^{d-1}} e^{2\pi i \bm{z}\cdot\bm{\xi}} \widetilde{F}(\bm{\xi})d\bm{\xi}\\
  	&= \int_{\mathbb{R}^{d-1}} e^{i \bm{z}\cdot\bm{\xi}} (2\pi)^{-(d-1)}\widetilde{F}(\bm{\xi}/2\pi) d\bm{\xi}\\
  	&= \int_{\mathbb{R}^{d-1}} e^{i \bm{z}\cdot\bm{\xi}} F_{1}(\bm{\xi}) d\bm{\xi}\\
  	&= c_{1}+\int_{\mathbb{R}^{d-1}} (e^{i \bm{z}\cdot\bm{\xi}}-1) F_{1}(\bm{\xi}) d\bm{\xi},\quad\text{with}\quad c_{1}:=\int_{\mathbb{R}^{d-1}}  F_{1}(\bm{\xi}) d\bm{\xi},
  \end{align*}
  where $ c_{1}\in [-C_{1},C_{1}] $ is well defined with $ C_{1} $ as in \eqref{btin3} (see Remark~\ref{remarkC1bar}).

  Then, we proceed to prove that the right-hand side of \eqref{btin3} is satisfied. Note that 
  \begin{align*}
 	\int_{\mathbb{R}^{d-1}}\norm{\bm{\xi}}_{1}|F_{1}(\bm{\xi})|d\bm{\xi}&=\int_{\mathbb{R}^{d-1}}\norm{\bm{\xi}}_{1}\left|(2\pi)^{-(d-1)}\widetilde{F}(\bm{\xi}/2\pi)\right|d\bm{\xi}\\
 	&=2\pi\int_{\mathbb{R}^{d-1}}\norm{\bm{\xi}}_{1} |\widetilde{F}(\bm{\xi}) |d\bm{\xi}\\
 	&=2\pi\int_{\mathbb{R}^{d-1}}\norm{\bm{\xi}}_{1} |\mathscr{F}[g](\bm{\xi}) |d\bm{\xi}
  \end{align*}
  and 
  \begin{align*}
  	\mathscr{F}[g](\bm{\xi}) &=\int_{\mathbb{R}^{d-1}} g(\bm{z}) e^{-2\pi i \bm{z}\cdot\bm{\xi}} d\bm{z}\\
  	&= \sum_{j=1}^{m}\theta_{j}\int_{\mathbb{R}^{d-1}}\varphi\left(M(\bm{z}-\bm{v}_{j}/M)\right)e^{-2\pi i \bm{z}\cdot\bm{\xi}} d\bm{z}\\ 
  	&=M^{-(d-1)} \sum_{j=1}^{m}\theta_{j}\int_{\mathbb{R}^{d-1}}\varphi\left(\bm{\omega}\right)e^{-2\pi i ((\bm{\omega}+\bm{v}_{j})/M)\cdot\bm{\xi}} d\bm{\omega}\\ 
  	&=M^{-(d-1)} s_{\bm{\theta}}(\bm{\xi})\int_{\mathbb{R}^{d-1}}\varphi\left(\bm{\omega}\right)e^{-2\pi i  \bm{\omega}\cdot(\bm{\xi}/M)} d\bm{\omega}\\
  	&=M^{-(d-1)}\mathscr{F}[\varphi](\bm{\xi}/M) s_{\bm{\theta}}(\bm{\xi}),\quad\text{where}\quad s_{\bm{\theta}}(\bm{\xi}):=\sum_{j=1}^{m}\theta_{j} e^{-2\pi i (\bm{v}_{j}/M)\cdot\bm{\xi}}.
  \end{align*} 
  Therefore,
  \[
  	\int_{\mathbb{R}^{d-1}}\norm{\bm{\xi}}_{1}|F_{1}(\bm{\xi})|d\bm{\xi}=2\pi M^{-(d-1)}\int_{\mathbb{R}^{d-1}}\norm{\bm{\xi}}_{1} \left|\mathscr{F}[\varphi](\bm{\xi}/M) s_{\bm{\theta}}(\bm{\xi})\right|d\bm{\xi}
  \]
  and considering the partition $\{U_{\bm{k}}\}_{\bm{k}\in \mathbb{Z}^{d-1}}$ of $ \mathbb{R}^{d-1} $, defined by 
  \[
  U_{\bm{k}}:=[0,M)^{d-1}+M\bm{k}=\left\{\bm{x}+M\bm{k}\mid \bm{x}\in[0,M)^{d-1} \right\}\quad\text{where}\quad \mathbb{R}^{d-1}=\bigsqcup_{\bm{k}\in \mathbb{Z}^{d-1}}U_{\bm{k}},
  \]
  we get 
  \begin{equation}
  	\int_{\mathbb{R}^{d-1}}\norm{\bm{\xi}}_{1}|F_{1}(\bm{\xi})|d\bm{\xi}=2\pi M^{-(d-1)}\sum_{\bm{k}\in \mathbb{Z}^{d-1}}\int_{U_{\bm{k}}}\norm{\bm{\xi}}_{1} \left|\mathscr{F}[\varphi](\bm{\xi}/M) s_{\bm{\theta}}(\bm{\xi})\right|d\bm{\xi}.\label{upbarF1}
  \end{equation}
  We make the following observations.
  \begin{itemize}
  	\item Note that    	
  	\begin{align}
  		\int_{U_{\bm{k}}}|s_{\bm{\theta}}(\bm{\xi})|^{2} d\bm{\xi}&=\int_{U_{\bm{k}}}s_{\bm{\theta}}(\bm{\xi})\,\overline{s_{\bm{\theta}}(\bm{\xi})} d\bm{\xi}\nonumber\\ 
  		&=\int_{U_{\bm{k}}}\left(\sum_{j=1}^{m}\theta_{j} e^{-2\pi i (\bm{v}_{j}/M)\cdot\bm{\xi}}\right)\left(\sum_{l=1}^{m}\theta_{l} e^{2\pi i(\bm{v}_{l}/M)\cdot\bm{\xi}}\right) d\bm{\xi}\nonumber\\	
  		&= \int_{U_{\bm{k}}}\sum_{j=1}^{m}  \sum_{l=1}^{m} \theta_{j}\theta_{l} e^{\frac{2\pi i}{M}  (\bm{v}_{l}-\bm{v}_{j})\cdot\bm{\xi}}  d\bm{\xi} \nonumber \\
  		&\leq \sum_{j,l=1}^{m} \int_{U_{\bm{k}}} e^{\frac{2\pi i}{M}  (\bm{v}_{l}-\bm{v}_{j})\cdot\bm{\xi}}  d\bm{\xi}\nonumber\\
  		&= \sum_{j,l=1}^{m} \int_{[0,M)^{d-1}}   e^{\frac{2\pi i}{M}  (\bm{v}_{l}-\bm{v}_{j})\cdot\bm{\xi}}  d\bm{\xi},\label{ubst0}
  	\end{align}
  	where   	
  	\begin{equation}
  		\int_{[0,M)^{d-1}}   e^{\frac{2\pi i}{M} (\bm{v}_{l}-\bm{v}_{j})\cdot\bm{\xi}}  d\bm{\xi}=\prod_{r=1}^{d-1}\int_{0}^{M}e^{\frac{2\pi i}{M}(v_{lr}-v_{jr})\xi_{r}}  d\xi_{r} \label{ubst1}
  	\end{equation}
  	and
  	\begin{equation}
  		\int_{0}^{M}e^{\frac{2\pi i}{M}(v_{lr}-v_{jr})\xi_{r}}  d\xi_{r}=\begin{cases}
  			M &\text{ if }~ v_{lr}=v_{jr}\\
  			0 &\text{ if }~ v_{lr}\neq v_{jr}.\\
  		\end{cases}\label{ubst2}
  	\end{equation}
  	Then, from \eqref{ubst0}, \eqref{ubst1} and \eqref{ubst2}, we obtain 	
  	\begin{align}
  		\int_{U_{\bm{k}}}\left|s_{\bm{\theta}}(\bm{\xi})\right|^{2}d\bm{\xi} &\leq \sum_{j,l=1}^{m} \int_{[0,M)^{d-1}}   e^{\frac{2\pi i}{M} (\bm{v}_{l}-\bm{v}_{j})\cdot\bm{\xi}}  d\bm{\xi}\nonumber\\
  		&=  M^{d-1}\sum_{j=1}^{m} 1\nonumber\\
  		&=  mM^{d-1}.\label{ubst3}
  	\end{align} 
  	
  	\item For $ \bm{\xi}\in U_{\bm{k}} $, we have  $ Mk_{j}\leq \xi_{j}< M(k_{j}+1)$, and this implies $ |\xi_{j}|\leq M(|k_{j}|+1) $. Therefore,
  	\begin{equation}
  		\norm{\bm{\xi}}_{1}\leq \sum_{j=1}^{d-1} |\xi_{j}|\leq M \sum_{j=1}^{d-1}(|k_{j}|+1)\leq (d-1)M(1+\norm{\bm{k}}_{\infty}).\label{upxi}
  	\end{equation}
  	In addition, let $ j^{*}\in \{1,\ldots,d-1\} $ satisfy $ \norm{\bm{k}}_{\infty}=|k_{j^{*}}|=t\geq 1 $. Then,
  	\[
  	0<t  \leq \xi_{j^{*}}/M<t+1\quad\text{whenever}\quad k_{j^{*}}=t
  	\]
  	and 
  	\[
  	0\leq t-1 \leq -\xi_{j^{*}}/M< t \quad\text{whenever}\quad k_{j^{*}}=-t,
  	\] 
  	which imply 
  	\[
  	\norm{\bm{\xi}}_{2}/M \geq |\xi_{j^{*}}|/M \geq t-1=\norm{\bm{k}}_{\infty}-1 \geq 2^{-1}(\norm{\bm{k}}_{\infty}+1)-1
  	\]
  	and therefore
  	\begin{equation}
  		(1+\norm{\bm{\xi}}_{2}/M)^{-(d+1)} \leq 2^{d+1}(\norm{\bm{k}}_{\infty}+1)^{-(d+1)} \quad\text{for all}\quad   \bm{k}\in \mathbb{Z}^{d-1}.\label{upxi2}
  	\end{equation}
  	
  	\item Since $ \varphi\in C_{c}^{\infty}(\mathbb{R}^{d-1};\mathbb{R}) $, we know  by \cite[Definition 2.2.1 and Example 2.2.2]{fourier} that $ \varphi $ is in the class of Schwartz functions.  Therefore, by \cite[Remark 2.2.4]{fourier}, there exists a constant $ C_{d}>0 $, such that 
  	\begin{equation}
  		|\mathscr{F}[\varphi](\bm{\xi}/M)|\leq C_{d}(1+\norm{\bm{\xi}/M}_{2})^{-(d+1)}.\label{upfu0}
  	\end{equation}
  	Then, from \eqref{upxi}, \eqref{upxi2} and \eqref{upfu0}, we have
  	\begin{align}
  		\int_{U_{\bm{k}}}\norm{\bm{\xi}}_{1}^{2} \left|\mathscr{F}[\varphi](\bm{\xi}/M) \right|^{2}d\bm{\xi}&\leq \left((d-1)M(1+\norm{\bm{k}}_{\infty})\right)^{2}\int_{U_{\bm{k}}}  \left|\mathscr{F}[\varphi](\bm{\xi}/M) \right|^{2}d\bm{\xi}\nonumber\\
  		&\leq \left((d-1)M(1+\norm{\bm{k}}_{\infty})C_{d}(1+\norm{\bm{\xi}/M}_{2})^{-(d+1)}\right)^{2}\int_{U_{\bm{k}}}  1d\bm{\xi}\nonumber\\
  		&\leq \left(2^{d+1}(d-1)MC_{d}(1+\norm{\bm{k}}_{\infty})^{-d}\right)^{2} M^{d-1}.\label{ubxifou}
  	\end{align} 
  	
  \end{itemize} 
  In conclusion, by \eqref{ubst3} and \eqref{ubxifou}, we may apply Hölder's inequality in $ L^{2} $ to \eqref{upbarF1} as follows,
  \begin{align}
  	\int_{\mathbb{R}^{d-1}}\norm{\bm{\xi}}_{1}|F_{1}(\bm{\xi})|d\bm{\xi}&=2\pi M^{-(d-1)}\sum_{\bm{k}\in \mathbb{Z}^{d-1}}\int_{U_{\bm{k}}}\norm{\bm{\xi}}_{1} \left|\mathscr{F}[\varphi](\bm{\xi}/M) s_{\bm{\theta}}(\bm{\xi})\right|d\bm{\xi}\nonumber\\
  	&\leq 2\pi M^{-(d-1)}\sum_{\bm{k}\in \mathbb{Z}^{d-1}}\left(\int_{U_{\bm{k}}}\norm{\bm{\xi}}_{1}^{2} \left|\mathscr{F}[\varphi](\bm{\xi}/M)\right|^{2}d\bm{\xi}\right)^{1/2}\left(\int_{U_{\bm{k}}}  \left|s_{\bm{\theta}}(\bm{\xi})\right|^{2}d\bm{\xi}\right)^{1/2}\nonumber\\
  	&\leq 2\pi \sum_{\bm{k}\in \mathbb{Z}^{d-1}}\left(\left(2^{d+1}(d-1)MC_{d}(1+\norm{\bm{k}}_{\infty})^{-d}\right)M^{-(d-1)/2}\right) (mM^{d-1})^{1/2}\nonumber\\
  	&= 2^{d+2}\pi C_{d} (d-1) M\sqrt{m} \sum_{\bm{k}\in \mathbb{Z}^{d-1}} (1+\norm{\bm{k}}_{\infty})^{-d}\nonumber\\
  	&= 2^{(d+5)/2} \pi C_{d} (d-1) M^{(d+1)/2} \sum_{\bm{k}\in \mathbb{Z}^{d-1}} (1+\norm{\bm{k}}_{\infty})^{-d}.\label{upbarlast0}
  \end{align}
  Moreover, let $\{T_{t}\}_{t=0}^{\infty}$ be the partition of $ \mathbb{Z}^{d-1} $, defined by
  \[
  \mathbbm{Z}^{d-1}=\bigsqcup_{t=0}^{\infty} T_{t}	\quad\text{where}\quad T_{t}=\left\{\bm{k}\in \mathbb{Z}^{d-1}\mid \norm{\bm{k}}_{\infty}=t\right\}.
  \]
  Then, 
  \begin{align*}
  	\sum_{\bm{k}\in \mathbb{Z}^{d-1}} \left(1+\norm{\bm{k}}_{\infty}\right)^{-d} 
  	&=\sum_{t=0}^{\infty}\sum_{\bm{k}\in T_{t}} \left(1+\norm{\bm{k}}_{\infty}\right)^{-d}\\
  	&=\sum_{t=0}^{\infty}\left(1+t\right)^{-d}\sum_{\bm{k}\in T_{t}} 1\\
  	&= \sum_{t=0}^{\infty} \left(1+t\right)^{-d}\# T_{t}\\
  	&=1+ \sum_{t=1}^{\infty} \left(1+t\right)^{-d}\# T_{t},
  \end{align*}
  where $ \# T_{0}=1 $ and for all $ t\geq 1 $,  
  \begin{align*}
  	\# T_{t}&=\# \left(\left\{\bm{k}\in \mathbb{Z}^{d-1}\mid \norm{\bm{k}}_{\infty}\leq t\right\}\setminus \left\{\bm{k}\in \mathbb{Z}^{d-1}\mid \norm{\bm{k}}_{\infty}\leq t-1\right\}\right)\nonumber\\
  	&=(2t+1)^{d-1}-(2(t-1)+1)^{d-1}\nonumber\\
  	&\leq 2\sum_{r=0}^{d-2} (2t+1)^{d-2-r}(2t-1)^{r}\nonumber\\
  	&\leq 2\sum_{r=0}^{d-2} (2t+1)^{d-2}\nonumber\\
  	&= 2 (d-1) (2t+1)^{d-2} \nonumber\\
  	&\leq  3^{d-1}(d-1)t^{d-2}.
  \end{align*}
  Therefore, 
  \begin{align}
  	\sum_{\bm{k}\in \mathbb{Z}^{d-1}} \left(1+\norm{\bm{k}}_{\infty}\right)^{-d} 
  	&=1+ \sum_{t=1}^{\infty} \left(1+t\right)^{-d}\# T_{t}\nonumber\\
  	&\leq 1+ 3^{d-1}(d-1)\sum_{t=1}^{\infty} \left(1+t\right)^{-d} t^{d-2}\nonumber\\
  	&\leq 1+ 3^{d-1}(d-1)\sum_{t=1}^{\infty} t^{-2}\nonumber\\
  	&= 1+ 3^{d-1}(d-1)(\pi^2/6).\label{sumZ0}
  \end{align}
  Combining \eqref{upbarlast0} and \eqref{sumZ0}, we obtain 
  \begin{align*}
  	\int_{\mathbb{R}^{d-1}}\norm{\bm{\xi}}_{1}|F_{1}(\bm{\xi})|d\bm{\xi}&\leq 2^{(d+5)/2} \pi C_{d} (d-1) M^{(d+1)/2} \sum_{\bm{k}\in \mathbb{Z}^{d-1}} (1+\norm{\bm{k}}_{\infty})^{-d}\\
  	&=\widetilde{C}_{d} M^{(d+1)/2},
  \end{align*}
  where 
  \[
  	\widetilde{C}_{d}:=2^{(d+5)/2} \pi C_{d} (d-1) (1+ 3^{d-1}(d-1)(\pi^2/6)).
  \]
  Then, in the right-hand side of \eqref{btin3},    
  \begin{equation}
  	\text{we set}\quad C_{1}:=\widetilde{C}_{d} M^{(d+1)/2}\quad\text{and choose}\quad  C_{\mathscr{C}}:= (C\widetilde{C}_{d}^{-1}/2) M^{-(d+1)/2},\label{CCfirsBar}
  \end{equation}
   so that \eqref{btin3} is satisfied.


\begin{remark}\label{remarkC1bar}
   	Note that $ C_{1}=\widetilde{C}_{d} M^{(d+1)/2} $ as in \eqref{CCfirsBar}, also satisfies
   	\[
   		\int_{\mathbb{R}^{d-1}} |F_{1}(\bm{\xi})|d\bm{\xi}<   C_{1}.
   	\]
   	Indeed, using the same argument applied to obtain \eqref{upbarF1} but now without the factor $ \norm{\bm{\xi}}_{1} $, we get
   	\[
   		\int_{\mathbb{R}^{d-1}} |F_{1}(\bm{\xi})|d\bm{\xi}=2\pi M^{-(d-1)}\sum_{\bm{k}\in \mathbb{Z}^{d-1}}\int_{U_{\bm{k}}}  \left|\mathscr{F}[\varphi](\bm{\xi}/M) s_{\bm{\theta}}(\bm{\xi})\right|d\bm{\xi}.
   	\]
   	By \eqref{ubst3}, \eqref{upxi2}, and \eqref{upfu0}, we apply Hölder's inequality in $ L^{2} $ to the terms in the sum on the right-hand side of the previous equation and obtain   	
   	\begin{align*}
   		\int_{\mathbb{R}^{d-1}} |F_{1}(\bm{\xi})|d\bm{\xi}&\leq 2\pi M^{-(d-1)}\sum_{\bm{k}\in \mathbb{Z}^{d-1}} \left(\int_{U_{\bm{k}}}  \left|\mathscr{F}[\varphi](\bm{\xi}/M)\right|^{2}d\bm{\xi}\right)^{1/2} \left(\int_{U_{\bm{k}}}  \left| s_{\bm{\theta}}(\bm{\xi})\right|^{2}d\bm{\xi}\right)^{1/2}\\
   		&\leq 2\pi M^{-(d-1)}\sqrt{m}M^{(d-1)/2}\sum_{\bm{k}\in \mathbb{Z}^{d-1}} \left(\int_{U_{\bm{k}}}  \left|\mathscr{F}[\varphi](\bm{\xi}/M)\right|^{2}d\bm{\xi}\right)^{1/2}\\
   		&\leq 2^{d+2}\pi  \sqrt{m}M^{-(d-1)/2}  C_{d} \sum_{\bm{k}\in \mathbb{Z}^{d-1}}  (\norm{\bm{k}}_{\infty}+1)^{-(d+1)} \left(\int_{U_{\bm{k}}}  1d\bm{\xi}\right)^{1/2}\\
   		&< 2^{d+2}\pi  \sqrt{m}  C_{d} \sum_{\bm{k}\in \mathbb{Z}^{d-1}}  (\norm{\bm{k}}_{\infty}+1)^{-d}.
   	\end{align*}
   	Then, the above inequality and \eqref{sumZ0} imply 
   	\[
   		\int_{\mathbb{R}^{d-1}} |F_{1}(\bm{\xi})|d\bm{\xi}<2^{(d+5)/2}\pi   C_{d}  \left(1+ 3^{d-1}(d-1)(\pi^2/6)\right)M^{(d-1)/2} < \widetilde{C}_{d} M^{(d+1)/2}.
   	\] 
   \end{remark}


   In summary, we have proved that $ \mathscr{C}_{\Theta} \subseteq \mathscr{C} $, that is, the left-hand side of \eqref{eqmaint0} holds. We now choose $ M $ so that the right-hand side of \eqref{eqmaint0} is satisfied. Thus, by \eqref{eqmaint0} and \eqref{CCfirsBar}, we have
   \[
   	2^{\widetilde{\gamma}+6}nC_{\mathscr{C}}^{\widetilde{\gamma}}M^{-(d-1)}=2^{\gamma+6} (C\widetilde{C}_{d}^{-1}/2)^{\gamma}  n M^{-(d+1)\gamma/2-(d-1)}  \leq 1,
   \]
   and therefore we choose $M$ to be 
   \begin{equation}
   	M:=2\lceil M_{*}/2 \rceil \quad\text{with}\quad M_{*}:=\left(2^{\gamma+6} (C\widetilde{C}_{d}^{-1}/2)^{\gamma}  n \right)^{\frac{1}{(d+1)\gamma/2+(d-1)}}.\label{chobarrM}
   \end{equation}
   Using the same argument as in \eqref{almHolCol2}, we get
   \begin{equation}
   	\left(2\lceil M_{*}/2 \rceil\right)^{-\gamma}    \geq (2C_{*}M_{*})^{-\gamma}\quad\text{with}\quad C_{*}:=\left( C\widetilde{C}_{d}^{-1}\right)^{-\frac{\gamma}{(d+1)\gamma/2+(d-1)}} +1/2.\label{barrMast}
   \end{equation}
   
   With the choices of $ M $ and $ C_{\mathscr{C}} $ in \eqref{CCfirsBar} and \eqref{chobarrM}, we know that the left-hand side condition in \eqref{Rinside01} is satisfied. However, for the right-hand side of \eqref{Rinside01} to hold, it is enough to choose a $ b_{0} $ that satisfies it, for instance  $ b_0\equiv1/2 $, $ b_{0} $ as in Figure~\ref{ExamElBCH}, or some more general choice. Therefore, since we have already verified all the hypotheses of Theorem~\ref{maintheo}, applying it and using \eqref{eqmaint1}, \eqref{CCfirsBar}, \eqref{chobarrM} and \eqref{barrMast}, we obtain that  
    \begin{align*}
    	\mathcal{I}_{n}(\mathscr{C})&\geq \left(\frac{r^{d-1}}{8^{\widetilde{\gamma}+2}}\right) C_{\mathscr{C}}^{\widetilde{\gamma}}\\
    	&=\left(\frac{r^{d-1}}{8^{\gamma+2}}\right)  (C\widetilde{C}_{d}^{-1}/2)^{\gamma} M^{-(d+1)\gamma/2}\\
    	&=\left(\frac{r^{d-1}}{8^{\gamma+2}}\right)  (C\widetilde{C}_{d}^{-1}/2)^{\gamma} (2\lceil M_{*}/2 \rceil)^{-(d+1)\gamma/2}\\
    	&\geq \left(\frac{r^{d-1}}{8^{\gamma+2}}\right)  (C\widetilde{C}_{d}^{-1}/2)^{\gamma} (2C_{*}M_{*})^{-(d+1)\gamma/2}\\
    	&= \left(\frac{r^{d-1}}{8^{\gamma+2}}\right)  (C\widetilde{C}_{d}^{-1}/2)^{\gamma}(2C_{*})^{-(d+1)\gamma/2} \left(2^{\gamma+6} (C\widetilde{C}_{d}^{-1}/2)^{\gamma}  n \right)^{-\frac{(d+1)\gamma/2}{(d+1)\gamma/2+(d-1)}},   \quad \text{for all}\quad n\in \mathbb{N}, 
    \end{align*}
   where $ r= \min\{1,(2C_{\varphi})^{-1/\alpha}\}=\min\{1,4^{-1}\}=4^{-1} $  (see \eqref{Cvaphibarr}). In other words,
   \[
   		\mathcal{I}_{n}(\mathscr{C}) \geq c_{d,C,\gamma}\, n^{-\frac{ \gamma}{\gamma+\left(\frac{2(d-1)}{d+1}\right)}},  \quad \text{for all}\quad n\in \mathbb{N}, 
   \]
   where 
   \[
   		c_{d,C,\gamma}:=\left(\frac{4^{-(d-1)}}{8^{\gamma+2}}\right)  (C\widetilde{C}_{d}^{-1}/2)^{\gamma}(2C_{*})^{-(d+1)\gamma/2} \left(2^{\gamma+6} (C\widetilde{C}_{d}^{-1}/2)^{\gamma}\right)^{-\frac{(d+1)\gamma/2}{(d+1)\gamma/2+(d-1)}}
   \] 
   is a constant depending only on $ d $, $ C $ and $\gamma$.\qed
   
 \subsection{Convex-Lipschitz class of functions}\label{subsecconvexlips}
 
  In this case, we take $ b_{0}\in \mathscr{C}:=\mathcal{C} $ for Construction~\ref{mainconst}, where $ \mathcal{C} $  is defined in Section \ref{somespaces}. It is well known that a convex function admits subgradients at every point in the interior of its domain (see \cite[Proposition 1.1]{subgrad}). Since $ \bm{v}_{j}/M \in (0,1)^{d-1} $, we may choose a subgradient $\bm{g}_{j} \in \mathbb{R}^{d-1}$ of $ b_{0} $ at $\bm{v}_{j}/M$, that is,
 \[
 b_{0}(\bm{z})\geq  \tau_{j}(\bm{z}):=b_{0}(\bm{v}_{j}/M)+\bm{g}_j\cdot(\bm{z}-\bm{v}_{j}/M)\quad \text{for all}\quad \bm{z}\in [0,1]^{d-1}.
 \] 
 Note that $ \tau_{j} $ defines a supporting hyperplane to the epigraph of $ b_{0} $ at $ (\bm{v}_{j}/M,b_{0}(\bm{v}_{j}/M)) $ (see \cite{subgrad}). If we choose $ b_{0} $ such that 
 \begin{equation}
 	b_{0}(\bm{z})>  \tau_{j}(\bm{z}) \quad\text{for all}\quad \bm{z}\in \mathbb{R}^{d-1}\setminus\{\bm{v}_{j}/M\},\label{cb0conv}
 \end{equation}
 then we can apply a sufficiently small vertical shift to $ \tau_{j} $ by adding some constant $ \delta>0 $ satisfying
\begin{equation}
	\left\{\bm{z}\in[0,1]^{d-1} \mid  \tau_{j}(\bm{z})+\delta>b_{0}(\bm{z}) \right\}\subseteq \left\{\bm{z}\in[0,1]^{d-1}\mid \norm{\bm{z}-\bm{v}_{j}/M}_{\infty}< (2M)^{-1}\right\}.\label{deltadef}
\end{equation}
 Hence $ \tau_{j}+\delta $ lies above $ b_{0} $ only on a subset of $ \operatorname{int}Q_{j}  $ around $ \bm{v}_{j}/M $, and with the notation $ (\cdot)_{+}:=\max\{\cdot,0\} $, we define
 \begin{equation}
 	\psi_{j}(\bm{z}):=\left(\tau_{j}(\bm{z})+\delta-b_{0}(\bm{z})\right)_{+},\label{defconsi}
 \end{equation}
 where by \eqref{deltadef}, we get 
 \begin{equation}
 	\suppf{\psi_{j}}\subseteq \left\{\bm{z}\in[0,1]^{d-1}\mid \norm{\bm{z}-\bm{v}_{j}/M}_{\infty}\leq (2M)^{-1}\right\}\subset \operatorname{int}Q_{j}\quad\text{for all}\quad j\in \{1,\ldots,m\}.\label{deltadef1}
 \end{equation}
 With the above definitions, our construction of $ b_{\bm{\theta}} $ with $\bm{\theta}  \in \Theta $ will be similar to the bump-type  used in the Hölder and Barron cases. If the $ j $-th entry of $ \bm{\theta} $ is $ 1 $, then the local affine function $ \tau_{j}(\bm{z})+\delta $ is activated on $ \suppf{\psi}_{j} $, replacing $b_{0}$; whereas if $ \theta_{j}=0 $, the function $ b_{\bm{\theta}} $ remains equal to $ b_{0} $ on $ \suppf{\psi_{j}} $ (see Figure~\ref{ExamElBCH}).
 
 In order to satisfy \eqref{cb0conv}, obtain uniform localized perturbations, and a function $ \varphi $ as required in Construction~\ref{mainconst}, we now fix a specific choice of $ b_{0}\in \mathscr{C} $, namely\footnote{This choice of $ b_{0} $ is convex, as it is a translation and a positive multiple of the convex function $\bm{z}\mapsto\norm{\bm{z}}_{2}^{2}$, and it is Lipschitz on $ [0,1]^{d-1} $ since its gradient is bounded on $ [0,1]^{d-1} $. Here, $ \bm{1/2}:=(1/2,\ldots,1/2)\in \mathbb{R}^{d-1} $.}
\begin{equation}
	b_{0}(\bm{z}):=1/4+(d-1)^{-1}\norm{\bm{z}-\bm{1/2}}_{2}^{2}.\label{b0convd}
\end{equation}
 Since $ b_{0} $ is differentiable on $ (0,1)^{d-1} $, the subgradient $ \bm{g}_{j} $ at $ \bm{v}_{j}/M $ is uniquely determined and coincides with the gradient of $ b_{0} $ at the same point, that is, 
 \[
 	\bm{g}_{j}=\nabla b_{0}(\bm{v}_{j}/M)=2(d-1)^{-1}(\bm{v}_{j}/M-\bm{1/2}).
 \]
Therefore, 
 \begin{small}
 	\begin{align*}
 		\psi_{j}(\bm{z})&=\left(\tau_{j}(\bm{z})+\delta-b_{0}(\bm{z})\right)_{+}\\
 		&=\left((d-1)^{-1}\norm{\bm{v}_{j}/M-\bm{1/2}}_{2}^{2} +\nabla b_{0}(\bm{v}_{j}/M)\cdot(\bm{z}-\bm{v}_{j}/M)+\delta-(d-1)^{-1}\norm{\bm{z}-\bm{1/2}}_{2}^{2}\right)_{+}\\
 		&=\left(-(d-1)^{-1}\left(\norm{\bm{z}-\bm{1/2}}_{2}^{2}-\norm{\bm{v}_{j}/M-\bm{1/2}}_{2}^{2}\right) +\nabla b_{0}(\bm{v}_{j}/M)\cdot(\bm{z}-\bm{v}_{j}/M)+\delta\right)_{+}\\ 
 		&=\left(-(d-1)^{-1}\left(\norm{\bm{z}-\bm{v}_{j}/M}_{2}^{2}	+2(\bm{v}_{j}/M-\bm{1/2})(\bm{z}-\bm{v}_{j}/M) \right) +\nabla b_{0}(\bm{v}_{j}/M)\cdot(\bm{z}-\bm{v}_{j}/M)+\delta\right)_{+}\\
 		&=\left(-(d-1)^{-1}\norm{\bm{z}-\bm{v}_{j}/M}_{2}^{2} +\delta\right)_{+}\\
 		&=\delta\left(1-(\delta(d-1))^{-1}M^{-2} \norm{M(\bm{z}-\bm{v}_{j}/M)}_{2}^{2} \right)_{+},
 	\end{align*}
 \end{small}
 where we need to choose $ \delta $ satisfying \eqref{deltadef}. In particular, we know from the previous identity that
 \begin{align*}
 	\left\{\bm{z}\in[0,1]^{d-1} \mid  \tau_{j}(\bm{z})+\delta>b_{0}(\bm{z}) \right\}&=\left\{\bm{z}\in[0,1]^{d-1}\mid \norm{\bm{z}-\bm{v}_{j}/M}_{2}< \sqrt{\delta(d-1)}  \right\}\\
 	&\subseteq\left\{\bm{z}\in[0,1]^{d-1}\mid \norm{\bm{z}-\bm{v}_{j}/M}_{\infty}< \sqrt{\delta(d-1)}  \right\},
 \end{align*} 
 so we can set  $\delta$ such that $ \sqrt{\delta(d-1)} = (2M)^{-1} $, i.e., $ \delta = (2M)^{-2}(d-1)^{-1} $. Then
 \begin{equation}
 	\psi_{j}(\bm{z}) =(2M)^{-2}(d-1)^{-1}\left(1-4\norm{M(\bm{z}-\bm{v}_{j}/M)}_{2}^{2} \right)_{+}, \label{psijpart}
 \end{equation}
 and to ensure that the hypotheses in Construction~\ref{mainconst} hold, we define $ \bm{\varphi} $ in \eqref{condtildeC} by
 \begin{align}
 	\bm{\varphi}(\bm{z})&:=\left(\psi_{1}(\bm{z}),\ldots,\psi_{m}(\bm{z})\right)\label{bmvarppsi}\\
 	&=(2M)^{-2}(d-1)^{-1}\left(\left(1-4\norm{M(\bm{z}-\bm{v}_{1}/M)}_{2}^{2} \right)_{+},\ldots,\left(1-4\norm{M(\bm{z}-\bm{v}_{m}/M)}_{2}^{2} \right)_{+}\right).\nonumber
 \end{align}         
 Accordingly, we take
 \begin{equation}
 	C_{\mathscr{C}}:=(2M)^{-2}(d-1)^{-1},\quad \varphi(\bm{z}):=\left(1-4\norm{\bm{z}}_{2}^{2} \right)_{+},\quad \varphi_{j}(\bm{z}):=\varphi(M(\bm{z}-\bm{v}_{j}/M)),\label{chconvarp}
 \end{equation}
 and verify the hypotheses of Theorem~\ref{maintheo} with these assumptions. 

By \eqref{chconvarp}, it follows that $ \varphi(\bm{0})=1 $, $ 0\leq \varphi(\bm{z})\leq1  $ for all $ \bm{z}\in \mathbb{R}^{d-1} $, and 
\begin{align*}
	\suppf{\varphi}&=\overline{\left\{\bm{z}\in\mathbb{R}^{d-1}\mid \varphi(\bm{z})>0 \right\}}\\
	&=\left\{\bm{z}\in\mathbb{R}^{d-1}\mid \norm{\bm{z}}_{2}\leq 1/2   \right\}\\
	&\subseteq \left\{\bm{z}\in\mathbb{R}^{d-1}\mid \norm{\bm{z}}_{\infty}\leq 1/2   \right\}\\
	&\subset (-1,1)^{d-1},
\end{align*}
 then \eqref{suppphijdef} is satisfied. In addition, 
 \begin{itemize}
 	\item If $ \varphi(\bm{z})=\varphi(\bm{w})=0 $, we obtain $ |\varphi(\bm{z})-\varphi(\bm{w})|=0 $.
 	
 	\item If  $ \varphi(\bm{z}),\varphi(\bm{w})\neq 0 $, we have $ \norm{\bm{z}}_{2},\norm{\bm{w}}_{2}<1/2 $ and
 	\begin{align*}
 			|\varphi(\bm{z})-\varphi(\bm{w})|&\leq ||1-4\norm{\bm{z}}_{2}^{2}|-|1-4\norm{\bm{w}}_{2}^{2}||\\
 			&\leq 4|\norm{\bm{z}}_{2}^{2}-\norm{\bm{w}}_{2}^{2}|\\
 			&\leq 4(|\norm{\bm{z}}_{2}+\norm{\bm{w}}_{2}|)\norm{\bm{z}-\bm{w}}_{2}\\
 			&\leq 4\norm{\bm{z}-\bm{w}}_{2}.
 	\end{align*}
 	
 	\item If $ \varphi(\bm{z})\neq 0 $ and $ \varphi(\bm{w})=0 $ (similarly, if  $ \varphi(\bm{z})=0 $ and $ \varphi(\bm{w})\neq 0 $), we get $ \norm{\bm{z}}_{2}<1/2\leq \norm{\bm{w}}_{2} $ and 
 	\begin{align*}
 		|\varphi(\bm{z})-\varphi(\bm{w})|&=1-4\norm{\bm{z}}_{2}^{2}\\
 		&=(1+2\norm{\bm{z}}_{2})(1-2\norm{\bm{z}}_{2})\\
 		&< 4(1/2-\norm{\bm{z}}_{2})\\
 		&=4(\norm{\bm{w}}_{2}-\norm{\bm{z}}_{2})\\
 		&\leq 4\norm{\bm{z}-\bm{w}}_{2}.
 	\end{align*}
 \end{itemize}
 Therefore, 
 \[
 	|\varphi(\bm{z})-\varphi(\bm{w})| \leq 4\norm{\bm{z}-\bm{w}}_{2}\quad \text{for all}\quad \bm{z},\bm{w}\in \mathbb{R}^{d-1},
 \] 
which shows that $ \varphi $ is a Hölder (Lipschitz) continuous function with exponent $\alpha=1 $ and constant $ K_{\varphi}=4 $, with respect to the Euclidean norm. Moreover, $ \varphi $ is Hölder (Lipschitz) continuous at $ \bm{0} $ with exponent  $ \alpha=1  $ and constant $ C_{\varphi}:=4\sqrt{d-1} $ with respect to the $ \ell_{\infty} $-norm, since
\[
|\varphi(\bm{z})-\varphi(\bm{0})| \leq 4\norm{\bm{z}}_{2}\leq 4\sqrt{d-1}\norm{\bm{z}}_{\infty}\quad \text{for all}\quad \bm{z}\in \mathbb{R}^{d-1}.
\] 
This completes the verification of the assumptions on $ \varphi $ required in Construction~\ref{mainconst}. 

Now, by \eqref{b0convd}, we see that 
\begin{align*}
	1/4\leq b_{0}(\bm{z})&=1/4+(d-1)^{-1}\norm{\bm{z}-\bm{1/2}}_{2}^{2} \\
	& \leq 1/4+ \norm{\bm{z}-\bm{1/2}}_{\infty}^{2}\\
	&\leq 1/2\quad \text{for all}\quad \bm{z} \in [0,1]^{d-1}.
\end{align*}
With $C_{\mathscr{C}}$ defined as in \eqref{chconvarp}, we obtain
\[
	C_{\mathscr{C}}=(2M)^{-2}(d-1)^{-1}<M^{-1}/4,
\]
and in particular $ C_{\mathscr{C}}<1/8 $, since $ M\geq 2 $.
Then, 
\[
C_{\mathscr{C}}<M^{-1}/4 \quad\text{and}\quad	b_{0}(\bm{z})\in[1/4,1/2]\subset [C_{\mathscr{C}},1-3C_{\mathscr{C}}] \quad\text{for all}\quad \bm{z}\in[0,1]^{d-1},
\]
which imply \eqref{Rinside01}.

Next, we only need to test hypothesis \eqref{eqmaint0} of Theorem \ref{maintheo}.  We start with $ \mathscr{C}_{\Theta} \subseteq \mathscr{C}  $. From \eqref{deltadef1}, \eqref{psijpart} and \eqref{chconvarp}, it follows that $ 	\suppf{\psi_{j}}=\suppf{\varphi_{j}} $. Then, using \eqref{defconsi}, \eqref{deltadef1}, \eqref{bmvarppsi} and item \ref{itemdishol2} of Lemma \ref{premainlemma}, with the notation 
\[
	J_{1}(\bm{\theta}):=\left\{j\in \{1,\ldots,m\}\mid \theta_{j}=1\right\},
\]
we obtain by cases that:
\begin{itemize}
	\item If $ \bm{z}\in \suppf{\psi_{k}} $ for some $ k\in \{1,\ldots,m\} $, 
	\begin{align*}
		b_{\bm{\theta}}(\bm{z})&=b_{0}(\bm{z})+\bm{\theta}\cdot\bm{\varphi}(\bm{z})\\
		&=b_{0}(\bm{z})+ \sum_{j\in J_{1}(\bm{\theta})}\psi_{j}(\bm{z})\\
		&=\begin{cases}
			\max\{b_{0}(\bm{z}),\tau_{k}(\bm{z})+\delta\} &~\text{ if }~ k\in J_{1}(\bm{\theta})\\
			b_{0}(\bm{z}) &~\text{ otherwise.}
		\end{cases}
	\end{align*}
		
	\item  If  $ \bm{z}\notin \suppf{\psi_{j}} $ for all $ j\in \{1,\ldots,m\} $,  
	\begin{align*}
		b_{\bm{\theta}}(\bm{z})&=b_{0}(\bm{z})+\bm{\theta}\cdot\bm{\varphi}(\bm{z})\\
		&=b_{0}(\bm{z})+ \sum_{j=1}^{m}\theta_{j}\psi_{j}(\bm{z})\\
		&=b_{0}(\bm{z}). 
	\end{align*} 
\end{itemize}
We also know (see \eqref{deltadef1}) that 
\[
	\suppf{\psi_{j}}=\overline{\left\{\bm{z}\in[0,1]^{d-1} \mid  \tau_{j}(\bm{z})+\delta>b_{0}(\bm{z}) \right\}} \subset \operatorname{int}Q_{j},
\]
and the sets $ Q_{j} $ form a partition of $ [0,1]^{d-1} $ up to boundaries (see Construction~\ref{mainconst}). Thus,
\begin{align*}
	b_{\bm{\theta}}(\bm{z}) &=\begin{cases}
		\max\{b_{0}(\bm{z}),\tau_{k}(\bm{z})+\delta\} &~\text{ if }~  \bm{z}\in \suppf{\psi_{k}} ~\text{ for some }~   k\in J_{1}(\bm{\theta}) 	\\
		b_{0}(\bm{z}) &~\text{ otherwise}
	\end{cases}\\
	&=\begin{cases}
		\max\{b_{0}(\bm{z}),\tau_{k}(\bm{z})+\delta\} &~\text{ if }~  \bm{z}\in \suppf{\psi_{k}} ~\text{ for some }~   k\in J_{1}(\bm{\theta}) 	\\
		\max\{b_{0}(\bm{z}),\tau_{k}(\bm{z})+\delta\} &~\text{ if }~\bm{z}\notin \suppf{\psi_{k}} ~\text{ for all }~   k\in J_{1}(\bm{\theta})
	\end{cases}\\
	&=\max\left\{b_{0}(\bm{z}),\left\{\tau_{k}(\bm{z})+\delta\right\}_{k\in J_{1}(\bm{\theta})}\right\},
\end{align*} 
where we used that if $\bm{z}\notin \suppf{\psi_{k}}$ for all $k\in J_{1}(\bm{\theta})$, then $\tau_{k}(\bm{z})+\delta\leq b_{0}(\bm{z})$ for all $k\in J_{1}(\bm{\theta})$. Therefore, for all $ \bm{z}\in [0,1]^{d-1} $, the function $ b_{\bm{\theta}}   $ is the maximum of $ b_{0}  $ and the functions $  \tau_{k} +\delta $ with $ k\in J_{1}(\bm{\theta}) $. Since $ b_{0} $ is convex and Lipschitz, and each $ \tau_{k} +\delta $ is affine (hence also convex and Lipschitz), it follows that $  b_{\bm{\theta}} $, as the maximum of convex and Lipschitz functions, is itself convex and Lipschitz. In conclusion,
\[
	\mathscr{C}_{\Theta} \subseteq \mathscr{C}.
\] 
Finally, to choose $ M $ that satisfies the right-hand side of \eqref{eqmaint0}, that is, 
\[
	2^{\widetilde{\gamma}+6}nC_{\mathscr{C}}^{\widetilde{\gamma}}M^{-(d-1)}=2^{\gamma+6}n((2M)^{-2}(d-1)^{-1})^{\gamma}M^{-(d-1)} \leq 1,
\]
we take 
\begin{equation}
	M:=2\lceil M_{*}/2 \rceil \quad\text{where}\quad M_{*}:= \left(2^{-\gamma+6}(d-1)^{-\gamma}  n \right)^{\frac{1}{(d-1)+2\gamma}}.\label{choMconv}
\end{equation}
 With an argument similar to that used in \eqref{almHolCol2}, we obtain
 \begin{equation*}
 	\left(2\lceil M_{*}/2 \rceil\right)^{-\gamma}    \geq (2C_{*}M_{*})^{-\gamma}\quad\text{where}\quad C_{*}:=\left(2(d-1)\right)^{\frac{\gamma}{(d-1)+2\gamma}} +1/2.
 \end{equation*}
 Therefore, since the hypotheses of Theorem~\ref{maintheo} are already satisfied, using \eqref{eqmaint1}, \eqref{chconvarp} and \eqref{choMconv}, we obtain that
 \begin{align*}
 	\mathcal{I}_{n}(\mathscr{C})&\geq \left(\frac{r^{d-1}}{8^{\widetilde{\gamma}+2}}\right) C_{\mathscr{C}}^{\widetilde{\gamma}}\\
 	&=\left(\frac{r^{d-1}}{8^{\gamma+2}}\right)  (2M)^{-2\gamma}(d-1)^{-\gamma}\\
 	&=\left(\frac{r^{d-1}(d-1)^{-\gamma}}{2^{5\gamma+6}}\right)  (2\lceil M_{*}/2 \rceil )^{-2\gamma} \\
 	&\geq\left(\frac{r^{d-1}(d-1)^{-\gamma}}{2^{5\gamma+6}}\right)  (2C_{*}M_{*})^{-2\gamma}\\
 	&= c_{d,\gamma}\,n^{-\frac{2\gamma}{(d-1)+2\gamma}}
 	  \quad \text{for all}\quad n\in \mathbb{N}, 
 \end{align*}
 where 
 \[
 	c_{d,\gamma}:=\left[\left(\frac{r^{d-1}(d-1)^{-\gamma}}{2^{5\gamma+6}}\right)  \left(2C_{*} \right)^{-2\gamma}  \left(2^{-\gamma+6}(d-1)^{-\gamma} \right)^{-\frac{2\gamma}{(d-1)+2\gamma}}\right],
 \]
 $ r:= \min\{1,(2C_{\varphi})^{-1}\} $ and $ C_{\varphi}:=4\sqrt{d-1} $. Lastly, 
 \[
 	\mathcal{I}_{n}(\mathscr{C})\geq c_{d,\gamma} \,n^{-\frac{2\gamma}{(d-1)+2\gamma}}
 	\quad \text{for all}\quad n\in \mathbb{N},
 \]
 where $ c_{d,\gamma} $ is a constant that depends only on $ d $ and $ \gamma $. \qed

\section*{Acknowledgements}

The authors thank Felix Voigtlaender and Mario Ullrich for helpful advice on covering numbers. J.G. and P.C.P. were supported by the Austrian Science Fund (FWF) Project P-37010.

\bibliographystyle{plainnat}
\bibliography{NewTLBref}

@article{tsysmoothdis,
	author = {Enno Mammen and Alexandre B. Tsybakov},
	title = {{Smooth discrimination analysis}},
	volume = {27},
	journal = {The Annals of Statistics},
	number = {6}, 
	pages = {1808 -- 1829}, 
	year = {1999},
	doi = {10.1214/aos/1017939240}
}

@article{subgrad,
	author = {Bubeck, Sébastien},
	title = {Convex Optimization: Algorithms and Complexity},
	journal = {Foundations and Trends in Machine Learning},
	volume = {8},
	number = {3-4},
	pages = {231-357},
	year = {2015}, 
	doi = {10.1561/2200000050} 
}

@book{fourier,
	author = {Grafakos, Loukas},
	year = {2014}, 
	title = {Classical Fourier Analysis},
	publisher = {Springer New York, NY},
	doi = {10.1007/978-1-4939-1194-3}
}

@book{tsybankovnonpa,
	author = {Tsybakov, Alexandre},
	year = {2009}, 
	title = {Introduction to Nonparametric Estimation},
	publisher = {Springer New York, NY},
	doi = {10.1007/b13794}
}

@article{noiseless1,
	author = {Krieg, David and Sonnleitner, Mathias},
	title = {Random points are optimal for the approximation of Sobolev functions},
	journal = {IMA Journal of Numerical Analysis},
	volume = {44},
	number = {3},
	pages = {1346-1371},
	year = {2023},
	doi = {10.1093/imanum/drad014}
}

@article{FashionMNISTref,
	title={Fashion-MNIST: a Novel Image Dataset for Benchmarking Machine Learning Algorithms}, 
	author={Han Xiao and Kashif Rasul and Roland Vollgraf},
	year={2017},
	journal={arXiv:1708.07747},
	doi={10.48550/arXiv.1708.07747}, 
}

@article{CIFAR10ref,
	title={Learning multiple layers of features from tiny images},
	author={Krizhevsky, Alex},
	year={2009},
	journal={University of Toronto},
	url={https://www.cs.toronto.edu/~kriz/cifar.html}
}

@article{noise1,
	author = {Charles J. Stone},
	title = {{Optimal Global Rates of Convergence for Nonparametric Regression}},
	volume = {10},
	journal = {The Annals of Statistics},
	number = {4},
	pages = {1040 -- 1053},
	year = {1982},
	doi = {10.1214/aos/1176345969}
}

@article{MNISTref,
	author={Lecun, Yann and Bottou, Léon and Bengio, Yoshua and Haffner, Patrick},
	journal={Proceedings of the IEEE}, 
	title={Gradient-based learning applied to document recognition}, 
	year={1998},
	volume={86},
	number={11},
	pages={2278-2324},
	doi={10.1109/5.726791}}

@article{Clements1963,
	author = {George F.  Clements},
	doi = {10.2140/pjm.1963.13.1085},
	issue = {4},
	journal = {Pacific Journal of Mathematics},
	title = {Entropies of several sets of real valued functions},
	volume = {13},
	year = {1963}
}

@article{barronmarginup,
	title = {High-dimensional classification problems with Barron regular boundaries under margin conditions},
	journal = {Neural Networks},
	volume = {192},
	pages = {107898},
	year = {2025},
	doi = {10.1016/j.neunet.2025.107898},
	author = {Jonathan García and Philipp Petersen}
}

@article{Convexf,
	author={Guntuboyina, Adityanand and Sen, Bodhisattva},
	journal={IEEE Transactions on Information Theory}, 
	title={Covering Numbers for Convex Functions}, 
	year={2013},
	volume={59},
	number={4},
	pages={1957-1965},
	doi={10.1109/TIT.2012.2235172}}

@ARTICLE{Barron1994,
	author={Andrew R. Barron{}{}},
	journal={IEEE Transactions on Information Theory}, 
	title={Universal approximation bounds for superpositions of a sigmoidal function}, 
	year={1993},
	volume={39},
	number={3},
	pages={930-945},
	doi={10.1109/18.256500}
}

@article{YangBarron,
	author = {Yuhong Yang and Andrew Barron},
	title = {Information-theoretic determination of minimax rates of convergence},
	volume = {27},
	journal = {The Annals of Statistics},
	number = {5},
	publisher = {Institute of Mathematical Statistics},
	pages = {1564 -- 1599},
	year = {1999},
	doi={10.1214/aos/1017939142}
	
}

@article{1993_Tikhomirov,
	author = {Andrey N. Kolmogorov and Vladimir M. Tihomirov},
	year = {1993},
	title = {$\varepsilon$-Entropy and $\varepsilon$-Capacity of Sets In Functional Spaces},
	journal = {Springer Nature},
	pages ={86--170},
	doi = {10.1007/978-94-017-2973-4_7}
}

@article{petersen2021optimal,
	title= {Optimal learning of high-dimensional classification problems using deep neural networks}, 
	author= {Philipp Petersen and Felix Voigtlaender},
	year= {2021},
	journal = {arXiv:2112.12555},
	doi = {10.48550/arXiv.2112.12555}
}

@article{NNbarronclass,
	author = {Andrei Caragea and Philipp Petersen and Felix Voigtlaender},
	title = {Neural network approximation and estimation of classifiers with classification boundary in a {Barron} class},
	volume = {33},
	journal = {The Annals of Applied Probability},
	number = {4},
	pages = {3039 -- 3079},
	year = {2023},
	doi = {10.1214/22-AAP1884}
}

@book{SVM,
	title= {Support Vector Machines}, 
	author= {Andreas Christmann and Ingo Steinwart},
	year= {2008},
	publisher = {Springer-Verlag New York},
	doi={10.1007/978-0-387-77242-4}
}

@article{fastc,
	title = {Fast convergence rates of deep neural networks for classification},
	journal = {Neural Networks},
	volume = {138},
	pages = {179-197},
	year = {2021},
	author = {Yongdai Kim and Ilsang Ohn and Dongha Kim},
	doi={10.1016/j.neunet.2021.02.012}
}

\end{document}